\documentclass[lettersize,journal]{IEEEtran}
\usepackage{amsmath,amsfonts}
\usepackage{algorithmic}
\usepackage{algorithm}
\usepackage{array}
\usepackage[caption=false,font=scriptsize,labelfont=sf,textfont=sf]{subfig}
\usepackage{textcomp}
\usepackage{stfloats}
\usepackage{url}
\usepackage{verbatim}
\usepackage{graphicx}
\usepackage{cite}

\usepackage{multirow}
\usepackage{booktabs}
\usepackage{tabularx}  
\usepackage{array}
\usepackage{diagbox}
\usepackage{hyperref}
\usepackage[table]{xcolor}
\usepackage{tabularx}

\begin{document}

\title{Revisiting Adversarial Training under Hyperspectral Image}

\author{Weihua Zhang, Chengze Jiang, Minjing Dong, 
Jie Gui,~\IEEEmembership{Senior Member,~IEEE}, 
Lu Dong,~\IEEEmembership{Senior Member,~IEEE}, 
Zhipeng Gui,~\IEEEmembership{Member,~IEEE}, 
Yuan Yan Tang,~\IEEEmembership{Life Fellow,~IEEE}, 
and James Tin-Yau Kwok,~\IEEEmembership{Fellow,~IEEE}
\thanks{W. Zhang, C. Jiang and L. Dong are with the School of Cyber Science
and Engineering, Southeast University, Nanjing 210000, China (e-mail:
zwhseu@seu.edu.cn; czjiang@seu.edu.cn; ldong90@seu.edu.cn).}
\thanks{M. Dong is with the Department of Computer Science, City University of
Hong Kong. (e-mail: minjdong@cityu.edu.hk).}
\thanks{Jie Gui is with the School of Cyber Science and Engineering and the
Engineering Research Center of Blockchain Application, Supervision and
Management, Ministry of Education, Southeast University, Nanjing 210000,
China, and also with Purple Mountain Laboratories, Nanjing 210000, China
(e-mail: guijie@seu.edu.cn).}
\thanks{Z. Gui is with the School of Remote Sensing and Information Engineering and the Collaborative Innovation Center of Geospatial Technology,
Wuhan University, Wuhan 430079, China (e-mail: zhipeng.gui@whu.edu.cn).}
\thanks{Y. Tang is with the Department of Computer and Information Science,
University of Macau, Macau 999078, China (e-mail: yytang@um.edu.mo).}
\thanks{J. T. -Y. Kwok is with the Department of Computer Science and Engineering, The Hong Kong University of Science and Technology, Hong Kong,
China (e-mail: jamesk@cse.ust.hk).}
}

\markboth{MANUSCRIPT FOR IEEE TRANSACTIONS ON INFORMATION FORENSICS AND SECURITY}%
{Shell \MakeLowercase{\textit{et al.}}: A Sample Article Using IEEEtran.cls for IEEE Journals}

\maketitle

\begin{abstract}
Recent studies have shown that deep learning-based hyperspectral image (HSI) 
classification models are highly vulnerable to adversarial attacks, posing significant security risks. Although most approaches attempt to enhance robustness by optimizing network architectures, these methods often rely on customized designs with limited scalability and struggle to defend against strong attacks. To address this issue, we introduce adversarial training (AT), one of the most effective defense strategies, into the hyperspectral domain. However, unlike conventional RGB image classification, directly applying AT to HSI classification introduces unique challenges due to the high-dimensional spectral signatures and strong inter-band correlations of hyperspectral data, where discriminative information relies on subtle spectral semantics and spectral–spatial consistency that are highly sensitive to adversarial perturbations. Through extensive empirical analyses, we observe that adversarial perturbations and the non-smooth nature of adversarial examples can distort or even eliminate important spectral semantic information. To mitigate this issue, we propose two hyperspectral-specific AT methods, termed AT-HARL and AT-RA. Specifically, AT-HARL exploits spectral characteristic differences and class distribution ratios to design a novel loss function that alleviates semantic distortion caused by adversarial perturbations. Meanwhile, AT-RA introduces spectral data augmentation to enhance spectral diversity while preserving spatial smoothness. Experiments on four benchmark HSI datasets demonstrate that the proposed methods achieve competitive performance compared with state-of-the-art approaches under adversarial attacks.
\end{abstract}


\section{Introduction}
\IEEEPARstart{D}{riven} by advances in deep learning, HSI classification has achieved remarkable progress \cite{7,8,9,6}. However, the adversarial robustness of HSI models has received little attention \cite{10,11,12,13}. Recent studies show that, owing to the spectral characteristics of hyperspectral data, HSI classifiers are also vulnerable to adversarial perturbations \cite{20,19,14}, underscoring the urgency of developing effective defense strategies.
\par
Current research on adversarial defenses in remote sensing mainly emphasizes robust architectural designs. Xu $et~al.$ \cite{20} introduced SACNet to enforce spectral–spatial attention consistency, while Zhang $et~al.$ \cite{27} proposed the Robust Class Context-Aware Network with supervised affinity loss to enhance contextual discrimination. Liu $et~al.$ \cite{28} combined random input masking with graph-based self-supervised learning to strengthen feature resilience, and Xu $et~al.$ \cite{29} further advanced this line with $\text{S}^{3}$ANet, which exploits pyramid-based spatial and global spectral attention. Although these approaches improve robustness, they often rely on customized architectures or specialized inputs, limiting scalability and offering only incremental gains. 
\par

In contrast, AT has demonstrated strong effectiveness in improving the robustness of deep models against perturbations in SAR and optical images \cite{16,LIAN1}. By explicitly optimizing models under adversarially perturbed inputs, AT enhances the stability of learned feature manifolds and mitigates the impact of sensor noise, illumination variation, and geometric distortion—factors that are pervasive in remote sensing scenarios. 
Although it is natural to involve AT in the field of hyperspectral classification, the direct adoption of AT to HSI data could be rather challenging given the nature of HSI data. Specifically,
HSI captures continuous spectral responses across hundreds of narrow bands, resulting in high-dimensional, strongly correlated spectral–spatial structures. Such characteristics make perturbation propagation and feature coupling in HSI fundamentally different, as adversarial perturbations can propagate across highly correlated spectral bands and distort the underlying spectral semantics, thereby complicating feature discrimination and decision boundary learning. Despite its potential to improve model robustness, AT in HSI has not been systematically explored, leaving an open problem on how to design spectrally aware adversarial strategies tailored for hyperspectral data.
\par
Therefore, we revisit AT under the hyperspectral setting to investigate how to enhance model robustness. Specifically, we take remote sensing adversarial training (RSAT) \cite{16} as the baseline and adapt it to HSI, enabling a systematic evaluation of hyperspectral classification robustness. Our experiments in Section \ref{RethinkAT} reveal two intrinsic challenges introduced by the spectral characteristics of HSI. First, the high-dimensional and highly correlated spectral structure makes spectral semantic representations highly sensitive to adversarial perturbations, resulting in severe spectral semantic distortion that is rarely observed in RGB images. Second, adversarial perturbations in hyperspectral data tend to propagate across the spectral dimension rather than remain localized, due to strong inter-band correlations. This propagation effect amplifies spectral inconsistencies during adversarial training and biases the learned decision boundaries toward a few spectrally dominant classes. As a consequence, misclassifications tend to concentrate on specific classes, forming a phenomenon we term the \textbf{misclassification concentration phenomenon}. This phenomenon reflects a structural limitation of standard AT when applied to HSI.
\par
\begin{figure*}[htbp]
    \centering
    \includegraphics[width=1.0\linewidth]{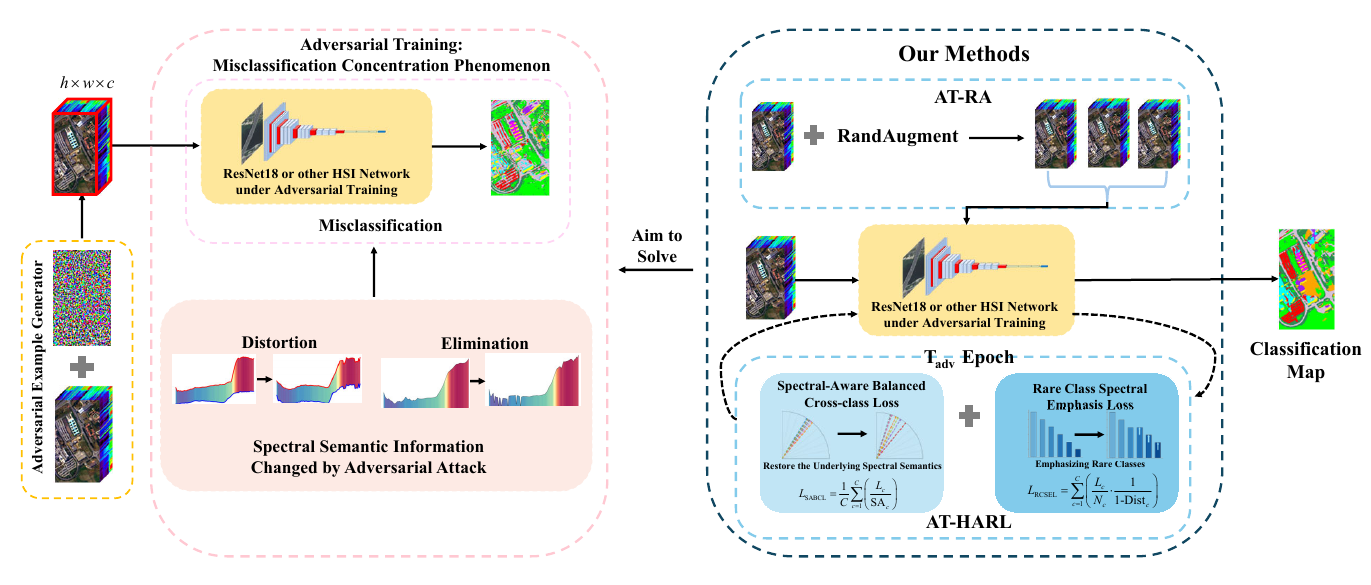}
    \caption{Overall framework of the proposed adversarial training strategy for HSI. The framework addresses the misclassification concentration phenomenon caused by adversarial perturbations (distortion and elimination of spectral semantic information) through two complementary approaches: (1) RandAugment–based adversarial training (AT-RA), and (2) Hyperspectral-Aware Rare-class Learning (AT-HARL), which incorporates spectral-aware balanced cross-class loss (SABCL) and rare class spectral emphasis loss (RCSEL) to restore spectral semantics and emphasize rare classes.}
    \label{fig0-1}
\end{figure*}

To address these challenges, we propose two alternative AT strategies tailored for HSI: AT-HARL and AT-RA. These methods provide complementary perspectives for mitigating the structural limitations of standard AT in HSI classification. The first method, AT-HARL, focuses on preserving spectral semantic consistency during adversarial training. It incorporates hyperspectral domain priors into the loss design through two regularization components: the Spectral-Aware Balanced Cross-Class Loss (SABCL), which dynamically reweights class-wise losses according to spectral angular differences to preserve discriminative spectral semantics, and the Rare Class Spectral Emphasis Loss (RCSEL), which emphasizes learning on underrepresented classes using spectral statistics to mitigate the bias toward spectrally dominant categories. In contrast, AT-RA adopts an augmentation-driven strategy that aims to improve robustness by enhancing spectral diversity and spatial smoothness. Through randomized spectral–spatial perturbations, AT-RA reduces the accumulation of spectral distortions during AT and alleviates the resulting classification bias. These two strategies offer different yet effective approaches to mitigating the misclassification concentration phenomenon in hyperspectral AT. The main contributions of this work are summarized as follows.
\begin{itemize}  
    \item We identify a \emph{misclassification concentration phenomenon} under AT in HSI, where errors collapse into dominant classes due to disrupted spectral semantics.
    \item We propose two hyperspectral-aware adversarial training strategies, namely AT-HARL and AT-RA, to alleviate misclassification concentration. The overall framework is shown in Fig. \ref{fig0-1}.
    \item Extensive experiments on four standard HSI datasets under diverse adversarial attacks demonstrate that our methods outperform state-of-the-art solutions. 
\end{itemize} 

The remainder of this paper is organized as follows. Section~\ref{ReWo} reviews related works. Section~\ref{RethinkAT} rethinks adversarial training for HSI and identifies the misclassification concentration phenomenon. Section~\ref{Methods} introduces the proposed methodology, including the AT-HARL and AT-RA frameworks. Section~\ref{experiments} presents the qualitative and quantitative experimental results along with comprehensive analyses. Finally, Section~\ref{Conclusion} concludes the paper.

\section{Related Works}\label{ReWo}
\subsection{Adversarial Attack}
Adversarial attacks deliberately craft imperceptible perturbations on input data to mislead models, posing serious threats to their reliability and security \cite{21, 22, LIAN11}. A seminal work, FGSM \cite{23}, generates adversarial examples by adding perturbations along the gradient sign direction, formulated as
\begin{equation}
    x' = x + \xi \cdot {\rm{sign}}(\nabla_x \mathcal{L}(f(x;\theta),y)),
\end{equation}
where \(x\) is the input, \(y\) is the label, \(f(.;\theta)\) is the model, and \(\xi\) controls the perturbation magnitude. After that, PGD \cite{24} extends FGSM by iteratively updating with step size \(\epsilon\)and projecting perturbations onto the \(L_\infty\)-norm ball as
\begin{equation}
    x_{t+1}' = \mathrm{clip}_{\xi}\Big(x_t' + \epsilon \cdot {\rm{sign}}(\nabla_{x_t'} \mathcal{L}(f(x_t';\theta),y))\Big).
\end{equation}
\par
Recently, several adversarial attack strategies have been explored in the hyperspectral domain. Among the classical approaches, the Carlini--Wagner (CW) attack \cite{43} and AutoAttack (AA) \cite{44} are widely regarded as strong baselines, with CW optimizing perturbations via constrained optimization and AA providing a standardized ensemble framework for reliable robustness evaluation. Beyond these, more advanced attacks tailored to hyperspectral data have emerged. Shi $et~al.$ \cite{VMI-FGSM} developed a universal object-level attack using VMI-FGSM, shifting from pixel-level perturbations to object-level universal perturbations and thereby improving transferability across models and datasets. Building on spectral unmixing, Li $et~al.$ \cite{SUGAA} proposed a sparse-unmixing guided attack (SUGAA) that enforces abundance sparsity to improve the stealthiness and interpretability of adversarial perturbations. Furthermore, Li \textit{et al.} \cite{CoEMTO} introduced CoEMTO, a collaborative evolutionary multitasking attack that perturbs abundance for physical plausibility while improving Pareto convergence across multiple objectives. Nevertheless, in contrast to the growing diversity of attack methods, research on effective adversarial defense for HSI remains limited, motivating our focus on rethinking adversarial training in this work.

\begin{figure}[h]
    \centering
    \subfloat[AT PaviaU (Clean)]{\includegraphics[width=0.5\linewidth]{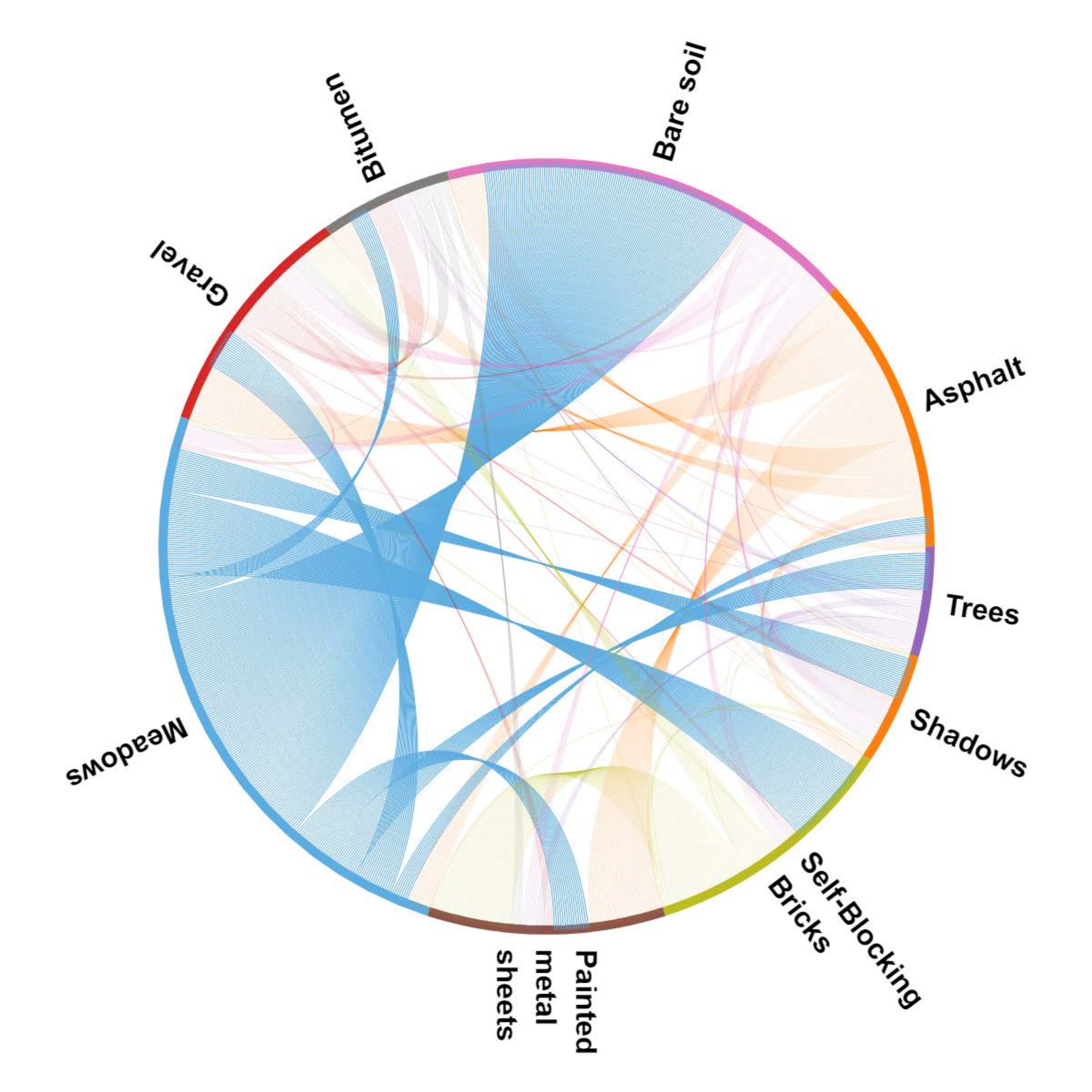}}
    \subfloat[AT PaviaU (Adversarial)]{\includegraphics[width=0.5\linewidth]{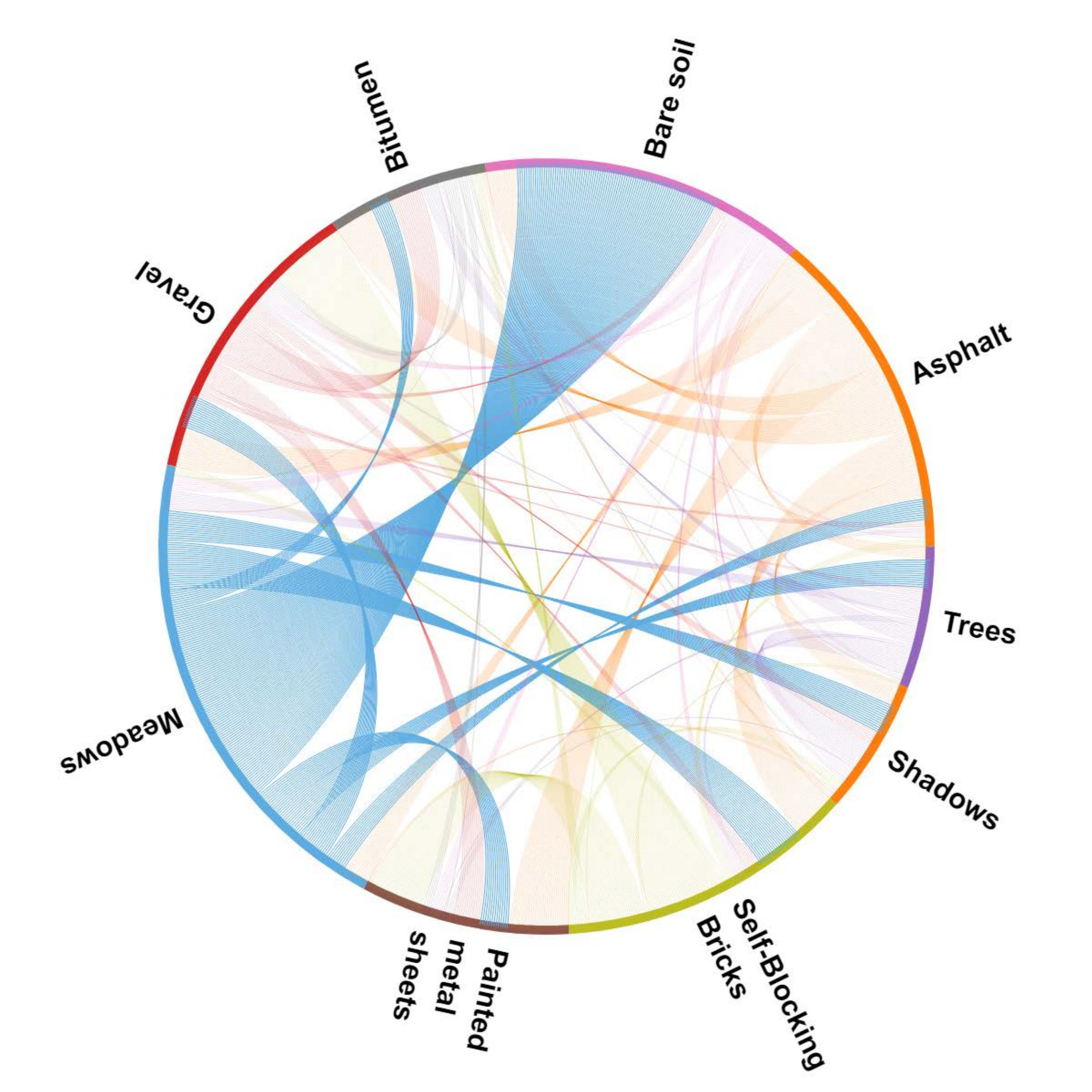}}\\
    \subfloat[AT Salinas (Clean)]{\includegraphics[width=0.5\linewidth]{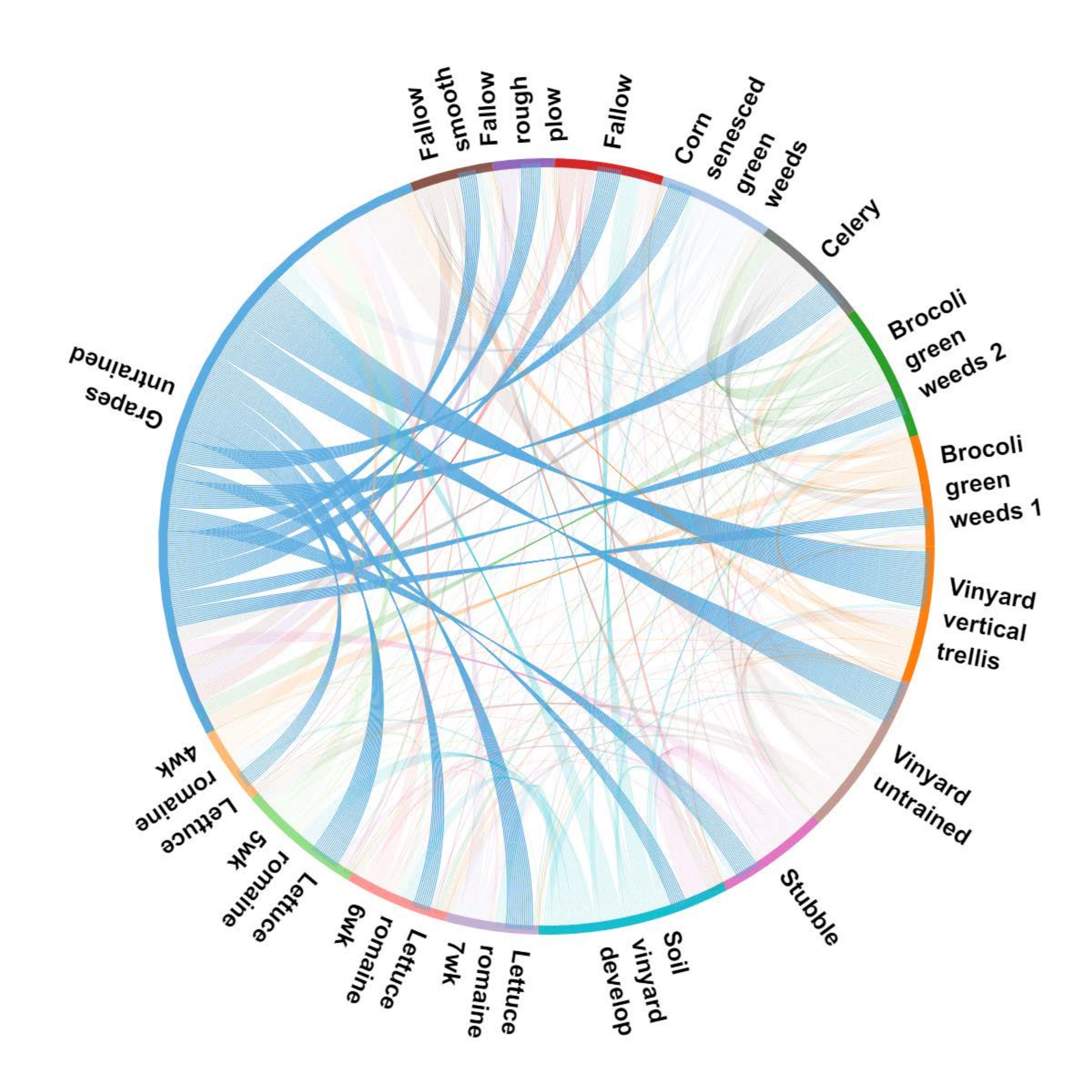}}
    \subfloat[AT Salinas (Adversarial)]{\includegraphics[width=0.5\linewidth]{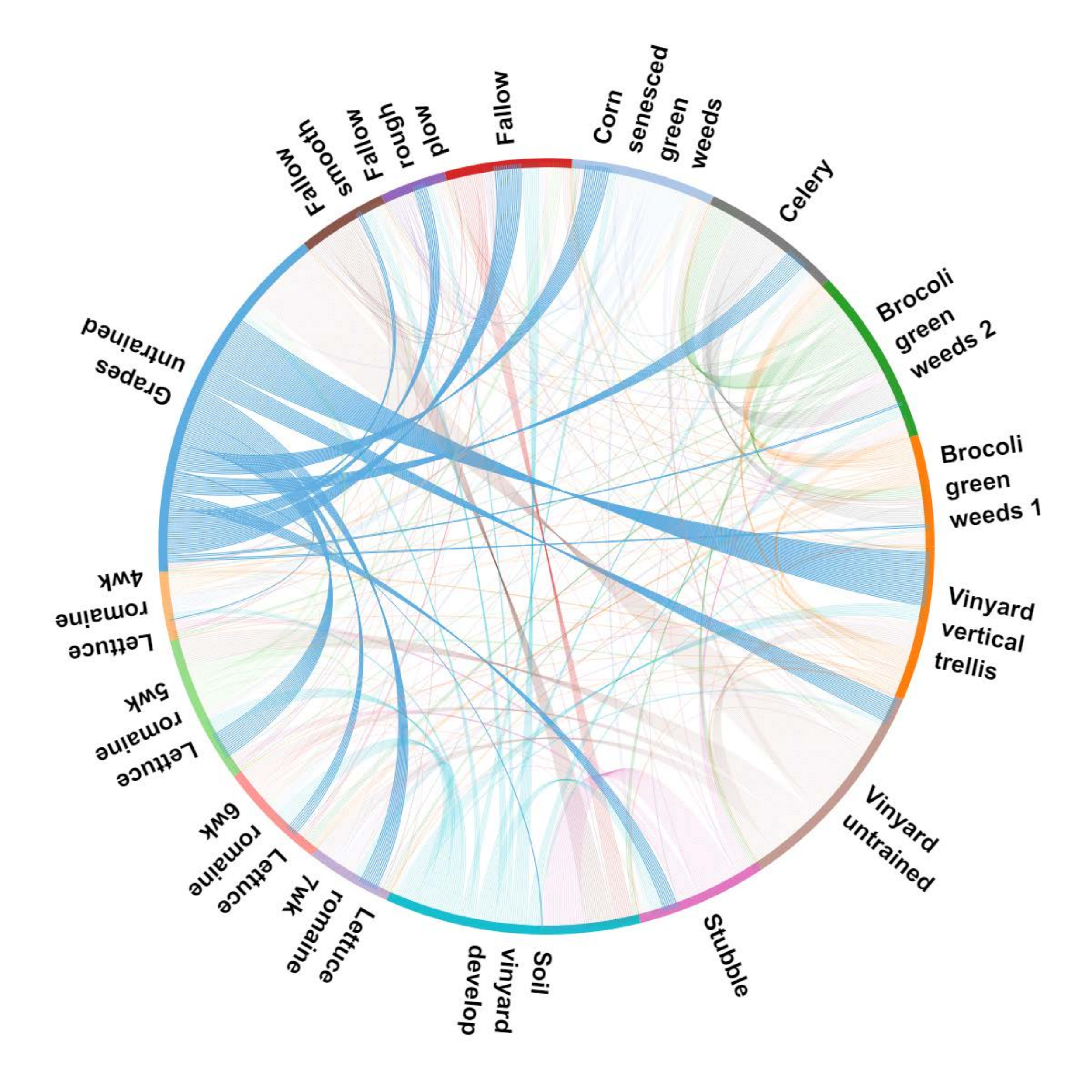}}
    \caption{Undirected chord diagrams of AT on the PaviaU and Salinas datasets. Across all settings, standard AT exhibits the \textbf{misclassification concentration phenomenon}, where errors from multiple classes collapse into a dominant class (\textit{Meadows} in PaviaU and \textit{Grapes untrained} in Salinas), with the effect amplified under adversarial perturbations.}
    \label{fig1-0}
\end{figure}

\subsection{Adversarial Defense in the Hyperspectral Domain}
Adversarial attacks expose security vulnerabilities in remote sensing deep learning models \cite{LIAN7, LIAN12, LIAN8}, drawing increasing attention to the development of effective defense mechanisms against such threats \cite{LIAN9, LIAN10, LIAN5}. 
\subsubsection{Robust Architectures Defense Methods}
In the hyperspectral domain, adversarial defenses primarily aim to design robust network architectures that capture global spatial relationships and improve robustness. Xu $et~al.$ \cite{20} proposed SACNet, which enforces spectral–spatial attention consistency to enhance feature alignment and mitigate adversarial perturbations. Similarly, Zhang $et~al.$ \cite{27} present the Robust Class Context-Aware Network, which employs a supervised affinity loss to discriminate intra- and inter-class contextual information. In addition, Liu $et~al.$ \cite{28} integrate random input masks with graph-based self-supervised learning to build a more resilient framework. Recently, Xu $et~al.$ \cite{29} propose the Spatial-Spectral Self-Attention Network ($\text{S}^{3}$ANet), which exploits pyramid-based spatial and global spectral relationships for improved adversarial defense. Nonetheless, these model customization methods require architectural modifications and impose restrictions on input data formats, which limit their scalability and practicality.

\subsubsection{Adversarial Training}
Adversarial training (AT) is widely regarded as one of the most principled defenses against adversarial attacks \cite{30}, as it augments training data with adversarial examples to encourage consistent responses on both clean and adversarial inputs. Madry $et~al.$ \cite{24} formalized adversarial training as a min-max optimization as follows:
\begin{equation}\label{eq:3}
    \operatorname*{min}_{\theta}\mathbb{E}_{(x,y)\sim \mathcal{D}}\left[\operatorname*{max}_{\delta\in[-\xi,\xi]}\mathcal{L}\left(f\left(x^{^{\prime}};\theta\right),y\right)\right],
\end{equation}
where \(\mathcal{D}\) denotes the underlying data distribution, and \(\mathbb{E}\) represents the expectation taken over examples \((x,y)\) drawn from \(\mathcal{D}\). Under the paradigm of equation \eqref{eq:3},  MART emphasizes misclassified examples to improve robustness against worst-case perturbations  \cite{MART}. Zhang $et~al.$ \cite{CFA} proposed CFA, which enforces consistency between clean and adversarial features to enhance generalization. Xu $et~al.$ \cite{RAT} developed RAT by incorporating regularization to mitigate overfitting on adversarial data. More recently, Liu $et~al.$ \cite{PIAT} presented PIAT, which progressively increases perturbation strength to enable smooth model adaptation and improve adversarial robustness.
\par
As a representative work on enhancing robustness in remote sensing, remote sensing adversarial training (RSAT) \cite{16} incorporates a benign example pretraining module (BEPM) and an adversarial–benign loss (ABL). The BEPM is applied before adversarial training to enable the model to learn features from benign examples, which is formulated as
\begin{equation}
    \begin{aligned}
        & \underset{\theta_{pt}}{\arg \min}\;\mathcal{L}_{\text{adv}}\left(\theta_{pt};x^{\prime},y\right), \\
        & \mathrm{where~}\theta_{pt}=\underset{\theta}{\arg \min}\;\mathcal{L}_{\text{ben}}\left(\theta;x,y\right), \\
        & \;\;\;\;\;\;\;\;\;\;\;x^{\prime}=\underset{\left\|x^{\prime}-x\right\|_{\text{p}}\leq\epsilon}{{{\arg \max}}}\;\mathcal{L}_{\text{ben}}(\theta_{pt};x^{\prime},y),
    \end{aligned}
\end{equation}
where \(\theta\) denotes the initial model parameters, and \(\theta_{pt}\) is the parameters after pre-training on benign examples. \(x^{\prime}\) denotes the adversarial example with $y$ as its label. ${\cal L}_{\text{ben}}$ is the classification loss on benign data, while ${\cal L}_{\text{adv}}$ is the adversarial loss. After that, ABL enforces the model to produce consistent outputs for a benign example $x$ and its adversarial counterpart $x'$, which is formulated as follows:
\begin{equation}
    {{\cal L}_{\text{adv}}}\left( {\theta ;x,y} \right) = {{\cal L}_{\text{ben}}}\left( {\theta ;x',y} \right) + {{\cal L}_{\text{ben}}}\left( {\theta ;x,y} \right).
\end{equation}

RSAT \cite{16} employs the cross-entropy loss \( {\cal L}_{\text{ben}} \) to formulate both ${\cal L}_{\text{ben}}$ and ${\cal L}_{\text{adv}}$ during training. While effective for enhancing robustness in conventional remote sensing images such as SAR and optical data, its design does not explicitly incorporate the spectral characteristics of hyperspectral data, thereby limiting its applicability to HSI.
\par
On the other hand, existing hyperspectral defenses mainly rely on customized architectures that embed spectral–spatial priors \cite{20,27,28,29}. Although these approaches can improve robustness, they often require tailored network designs or input constraints, which hinder scalability and practical deployment. 
These limitations collectively motivate a systematic re-examination of AT in the hyperspectral domain, 
with the goal of developing methods that achieve robustness while preserving generality and practicality.

\begin{table}[t]
\caption{Benign Accuracy and Adversarial Robustness of Different Components on Four Datasets Using ResNet-18 \protect\cite{36} When Remote Sensing Adversarial Training is Used for Adversarial Training, Where the Best Results Have Been Bolded.}
\label{tab1}
\centering
\setlength{\tabcolsep}{3pt}
\begin{tabular}{c|c|cc|cccc}
\toprule[2pt]
Dataset & Method & BEPM & ABL & Benign & PGD-50 & CW & AA \\
\midrule
\multirow{4}{*}{Pavia}
& AT        &       &       & 97.09 & \textbf{75.33} & \textbf{90.53} & \textbf{72.85} \\
& AT-BEPM   & \checkmark &       & 88.00 & 37.03 & 45.99 & 30.59 \\
& AT-ABL    &       & \checkmark & \textbf{97.23} & 63.21 & 74.70 & 61.01 \\
& RSAT      & \checkmark & \checkmark & 86.82 & 28.83 & 39.98 & 25.22 \\
\midrule
\multirow{4}{*}{Salinas}
& AT        &       &       & 87.19 & 86.40 & 86.73 & -- \\
& AT-BEPM   & \checkmark &       & 85.82 & 73.07 & 74.02 & -- \\
& AT-ABL    &       & \checkmark & 88.92 & \textbf{86.85} & \textbf{87.19} & -- \\
& RSAT      & \checkmark & \checkmark & \textbf{99.64} & 62.90 & 70.21 & -- \\
\midrule
\multirow{4}{*}{Houston}
& AT        &       &       & 92.50 & \textbf{34.20} & \textbf{83.85} & \textbf{27.52} \\
& AT-BEPM   & \checkmark &       & 41.33 & 16.40 & 15.74 & 10.11 \\
& AT-ABL    &       & \checkmark & 91.92 & 29.21 & 44.70 & 22.18 \\
& RSAT      & \checkmark & \checkmark & \textbf{95.84} & 6.16 & 14.63 & 3.58 \\
\midrule
\multirow{4}{*}{Washington}
& AT        &       &       & 88.14 & \textbf{77.92} & \textbf{80.32} & \textbf{77.77} \\
& AT-BEPM   & \checkmark &       & 93.36 & 40.76 & 52.34 & 35.15 \\
& AT-ABL    &       & \checkmark & 66.70 & 39.87 & 45.98 & 36.53 \\
& RSAT      & \checkmark & \checkmark & \textbf{94.25} & 40.04 & 54.87 & 36.91 \\
\toprule[2pt]
\end{tabular}
\end{table}

\section{Rethinking Adversarial Training for Hyperspectral Classification}\label{RethinkAT}
\subsection{Effects of Different Components}
We first revisit AT in the HSI setting following the protocol in~\cite{16}. 
The main results are summarized in Tables~\ref{tab1}, 
while the complete evaluation under additional attacks is provided in \textbf{Table~II of Appendix}. 
We observe that standard AT consistently outperforms RSAT~\cite{16} in terms of adversarial robustness under strong white-box attacks, including PGD-50, CW~\cite{43}, and AutoAttack (AA)~\cite{44}. Although RSAT improves benign accuracy in some cases, it exhibits robustness degradation under optimization-based attacks, indicating that its robustness gains do not transfer well to HSI. 
\par
This observation can be interpreted in light of \cite{46}, which pointed out that continuous learning under high-confidence regions may reduce rather than enhance generalization. In RSAT, the BEPM and ABL components reinforce the model’s confidence on benign examples through repeated margin maximization, but this mechanism suppresses the model’s robustness in handling perturbed inputs, thereby weakening adversarial generalization. This limitation is particularly pronounced in HSI, where the high dimensionality and strong inter-band correlations amplify the negative impact of overconfident learning. As a result, RSAT fails to improve adversarial robustness in the hyperspectral domain. We therefore argue that AT offers a more efficient and generalizable solution, and our subsequent analysis is conducted based on standard AT. 
\par
To confirm this opinion, we conduct standard AT on six advanced robust deep neural networks for HSI. As summarized in Table \ref{tab3}, AT can indeed enhance the robustness of advanced models in HSI. For completeness, the corresponding full evaluation results under additional attack settings 
are reported in \textbf{Tables~III in the Appendix}. Nevertheless, we also observe that the robustness improvements are not uniform across all classes. In particular, while some classes benefit significantly from AT, others remain disproportionately vulnerable to adversarial perturbations. This uneven distribution of robustness raises the concern that AT may introduce systematic biases in class-level performance. Motivated by this finding, we conduct a more detailed investigation in the following section.

\begin{table}[t]
\centering
\caption{Benign accuracy and adversarial robustness of six robust deep neural networks on four datasets under adversarial training. The best results are highlighted in bold.}
\label{tab3}
\setlength{\tabcolsep}{3pt}
\begin{tabular}{c|c|cc|cc|cc}
\toprule[2pt]
\multirow{2}{*}{Dataset}    & \multirow{2}{*}{Model}  
& \multicolumn{2}{c|}{Benign} 
& \multicolumn{2}{c|}{PGD-50} 
& \multicolumn{2}{c}{CW} \\
\cmidrule(r){3-4} \cmidrule(r){5-6} \cmidrule(r){7-8}
 & & Clean & AT & Clean & AT & Clean & AT \\
\midrule
\multirow{6}{*}{Pavia}
 & DilatedFCN & 94.48 & \textbf{95.92} & 51.43 & \textbf{69.59} & 66.43 & \textbf{73.17} \\
 & SSFCN & \textbf{95.72} & 94.57 & 13.52 & \textbf{23.38} & 14.81 & \textbf{35.93} \\
 & SpaFCN & \textbf{96.32} & 94.96 & 50.08 & \textbf{52.94} & 57.48 & \textbf{64.76} \\
 & SACNet & 96.01 & \textbf{97.14} & 65.01 & \textbf{71.48} & 72.55 & \textbf{79.62} \\
 & RCCA & 96.87 & \textbf{95.21} & 67.32 & \textbf{73.98} & 77.81 & \textbf{88.32} \\
 & S\textsuperscript{3}ANet & \textbf{99.08} & 97.49 & 78.65 & \textbf{86.37} & 80.00 & \textbf{89.12} \\
\midrule
\multirow{6}{*}{Salinas}
 & DilatedFCN & \textbf{96.70} & 96.20 & 44.93 & \textbf{50.72} & 48.30 & \textbf{57.00} \\
 & SSFCN & 94.47 & \textbf{94.62} & 9.54 & \textbf{15.14} & 12.78 & \textbf{31.80} \\
 & SpaFCN & 97.27 & \textbf{97.54} & 58.15 & \textbf{69.29} & 65.73 & \textbf{71.50} \\
 & SACNet & 95.21 & \textbf{96.91} & 63.54 & \textbf{69.82} & 71.35 & \textbf{77.23} \\
 & RCCA & 99.32 & \textbf{99.52} & 46.84 & \textbf{65.01} & 62.01 & \textbf{68.67} \\
 & S\textsuperscript{3}ANet & \textbf{99.94} & 99.92 & 76.31 & \textbf{89.16} & 65.49 & \textbf{95.88} \\
\midrule
\multirow{6}{*}{Houston}
 & DilatedFCN & 92.51 & \textbf{94.06} & 42.05 & \textbf{47.98} & 48.75 & \textbf{58.65} \\
 & SSFCN & 96.89 & \textbf{97.30} & 12.66 & \textbf{31.68} & 16.92 & \textbf{36.33} \\
 & SpaFCN & 96.42 & \textbf{97.12} & 14.43 & \textbf{34.92} & 16.87 & \textbf{38.45} \\
 & SACNet & 96.65 & \textbf{96.65} & 76.84 & \textbf{82.49} & 83.41 & \textbf{89.88} \\
 & RCCA & 98.26 & \textbf{98.32} & 59.71 & \textbf{65.42} & 68.03 & \textbf{75.18} \\
 & S\textsuperscript{3}ANet & 99.49 & \textbf{99.51} & 71.28 & \textbf{79.43} & 81.04 & \textbf{86.75} \\
\midrule
\multirow{6}{*}{Washington}
 & DilatedFCN & 94.85 & \textbf{96.18} & 63.50 & \textbf{74.45} & 70.95 & \textbf{78.63} \\
 & SSFCN & 99.16 & \textbf{99.19} & 73.52 & \textbf{75.11} & 77.93 & \textbf{86.78} \\
 & SpaFCN & 98.40 & \textbf{97.52} & 74.03 & \textbf{79.67} & 81.19 & \textbf{83.47} \\
 & SACNet & 97.55 & \textbf{98.15} & 85.11 & \textbf{87.10} & 87.61 & \textbf{89.94} \\
 & RCCA & \textbf{96.08} & 95.39 & 26.30 & \textbf{41.31} & 32.89 & \textbf{49.80} \\
 & S\textsuperscript{3}ANet & \textbf{98.74} & 98.48 & 86.32 & \textbf{87.60} & 93.13 & \textbf{96.88} \\
\toprule[2pt]
\end{tabular}
\end{table}

\begin{figure}[htbp]
    \centering
    \subfloat[AT EuroSAT (Clean)]{\includegraphics[width=0.5\linewidth]{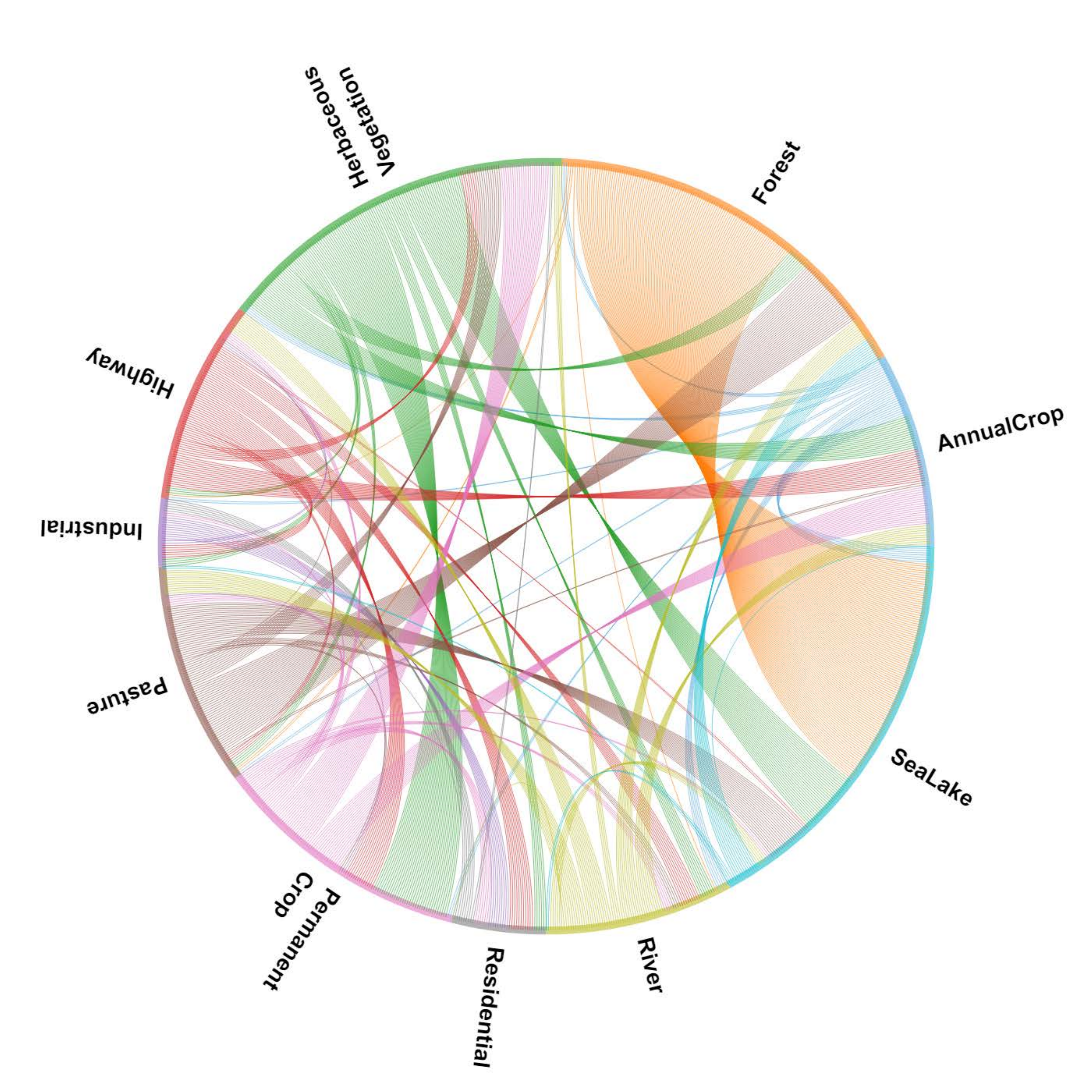}}
    \subfloat[AT EuroSAT (Adversarial)]{\includegraphics[width=0.5\linewidth]{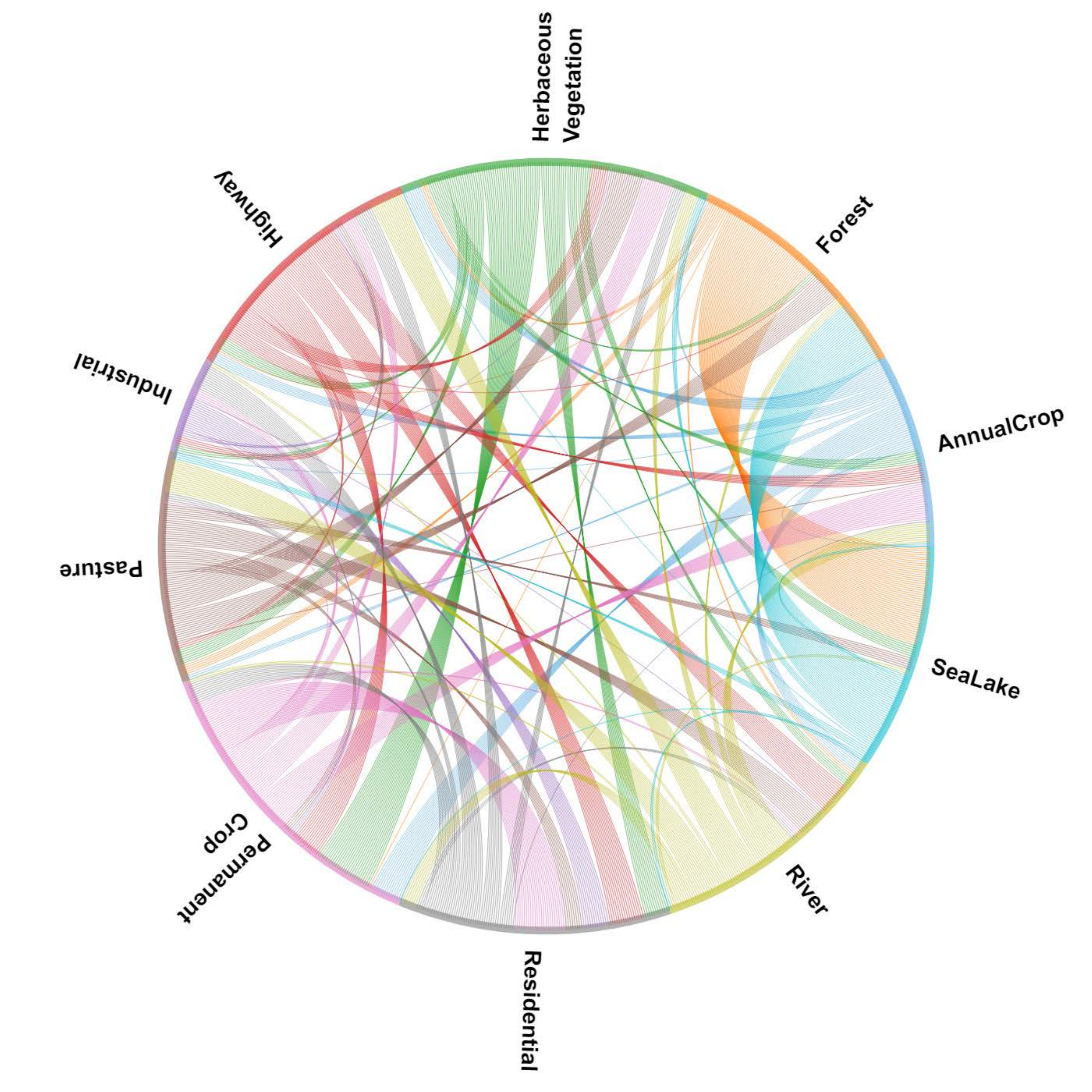}}\\
    \subfloat[AT SAT-4 (Clean)]{\includegraphics[width=0.5\linewidth]{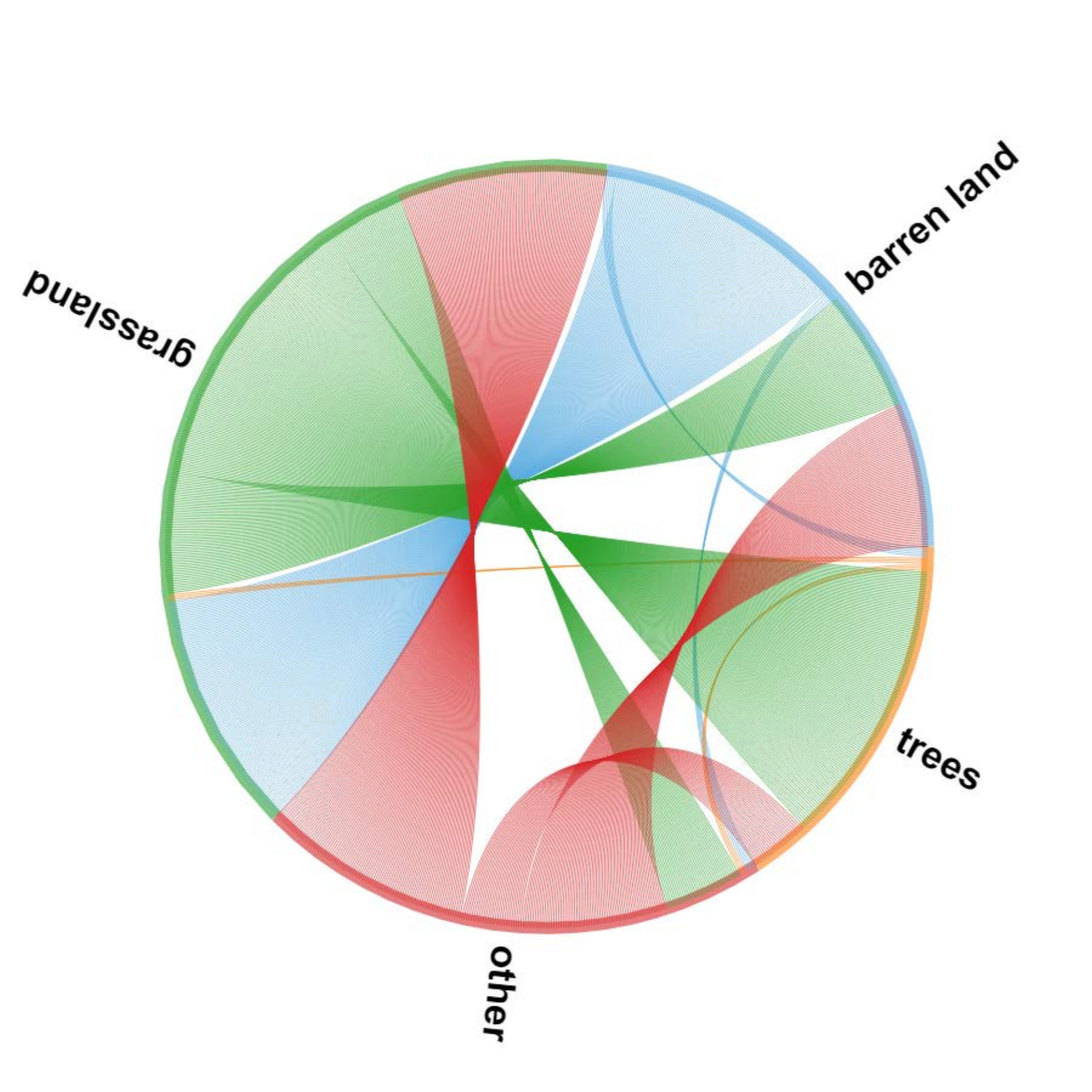}}
    \subfloat[AT SAT-4 (Adversarial)]{\includegraphics[width=0.5\linewidth]{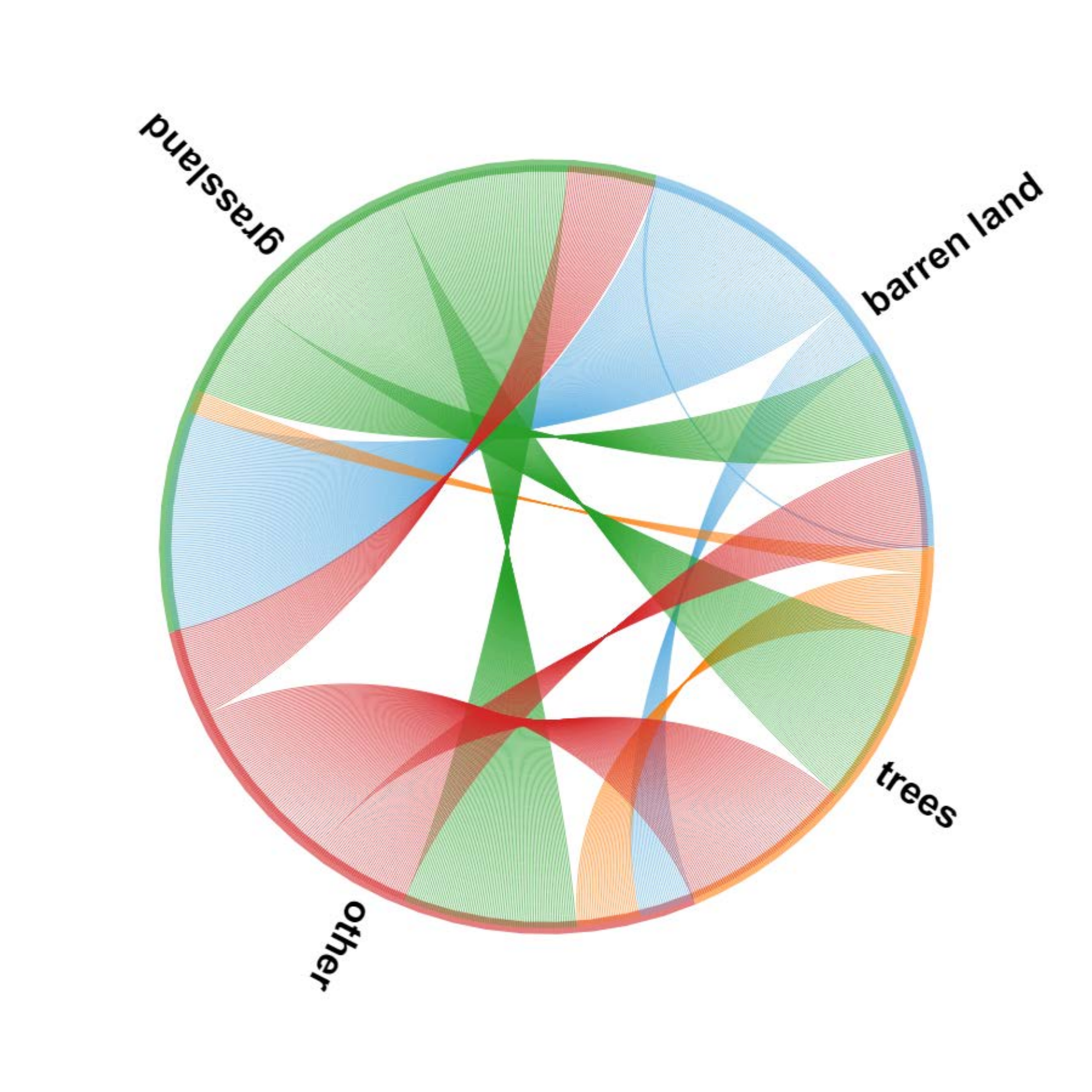}}
    \caption{Undirected chord diagrams of AT on optical remote sensing datasets EuroSAT and SAT-4. Unlike hyperspectral datasets, misclassification errors are more evenly distributed across classes, and no dominant error absorption behavior is observed, indicating the absence of the misclassification concentration phenomenon in optical images.}
    \label{fig:optical_chord}
\end{figure}

\subsection{Misclassification Concentration in Adversarial Training}

Under the AT framework, we investigate class-wise misclassification patterns and identify a critical limitation, termed the \textbf{misclassification concentration phenomenon}. As shown in the undirected chord diagrams in Fig.~\ref{fig1-0}, misclassification errors from multiple classes collapse into a single dominant class on both the PaviaU and Salinas datasets, converging toward \textit{Meadows} and \textit{Grapes untrained}, respectively. Further class-level analysis indicates that this phenomenon is unevenly distributed across categories. For instance, in the PaviaU dataset, samples from the \textit{Bare soil} class contribute a disproportionately large fraction of misclassifications into the dominant \textit{Meadows} class. The detailed class-wise confusion statistics supporting this observation are provided in \textbf{Tables~I in the Appendix}. Such concentrated misclassifications bias the learned decision boundaries toward dominant classes, thereby degrading benign accuracy and increasing vulnerability to adversarial perturbations for specific categories.

To examine whether this behavior is specific to hyperspectral images, we conduct comparative experiments on two optical remote sensing datasets, EuroSAT and SAT-4. As illustrated in Fig.~\ref{fig:optical_chord}, misclassification errors on optical datasets are more evenly distributed under AT, with no dominant error aggregation observed.

These results indicate that while class-wise imbalance may occur in general classification tasks, the pronounced misclassification concentration in Fig.~\ref{fig1-0} is strongly associated with the high-dimensional and highly correlated spectral structure of hyperspectral data, which amplifies the impact of adversarial perturbations on class decision boundaries. Motivated by this observation, we next analyze the underlying causes of this phenomenon from a spectral perspective.

\subsection{Study of the Misclassification Concentration Phenomenon}\label{sec:spec_analysis}

The misclassification concentration phenomenon observed above implies that AT can distort decision boundaries, causing predictions from multiple classes to collapse into a dominant class. This raises an important question: what underlying factors drive such boundary shifts in hyperspectral classification? Since HSI inherently rely on fine-grained spectral information for class separability, we hypothesize that adversarial perturbations may affect spectral properties in ways that compromise robustness. To explore this possibility, we perform a spectral analysis to investigate how adversarial perturbations alter the spectral characteristics of different classes.

\begin{figure*}[htbp]\centering
    \subfloat[PaviaU–Meadows (Benign)]{\includegraphics[width=0.25\linewidth]{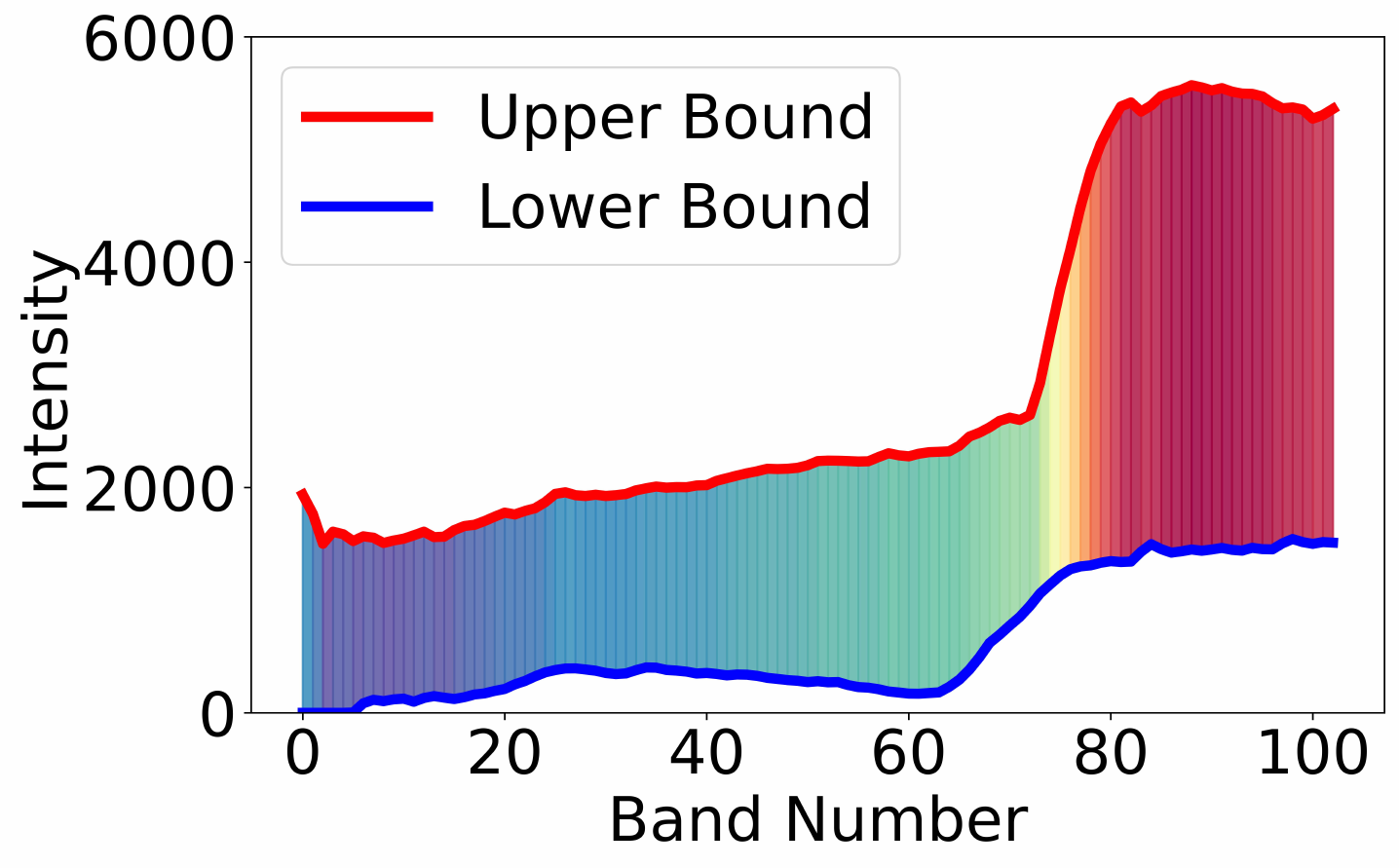}}
    \subfloat[PaviaU–Meadows (Adv)]{\includegraphics[width=0.25\linewidth]{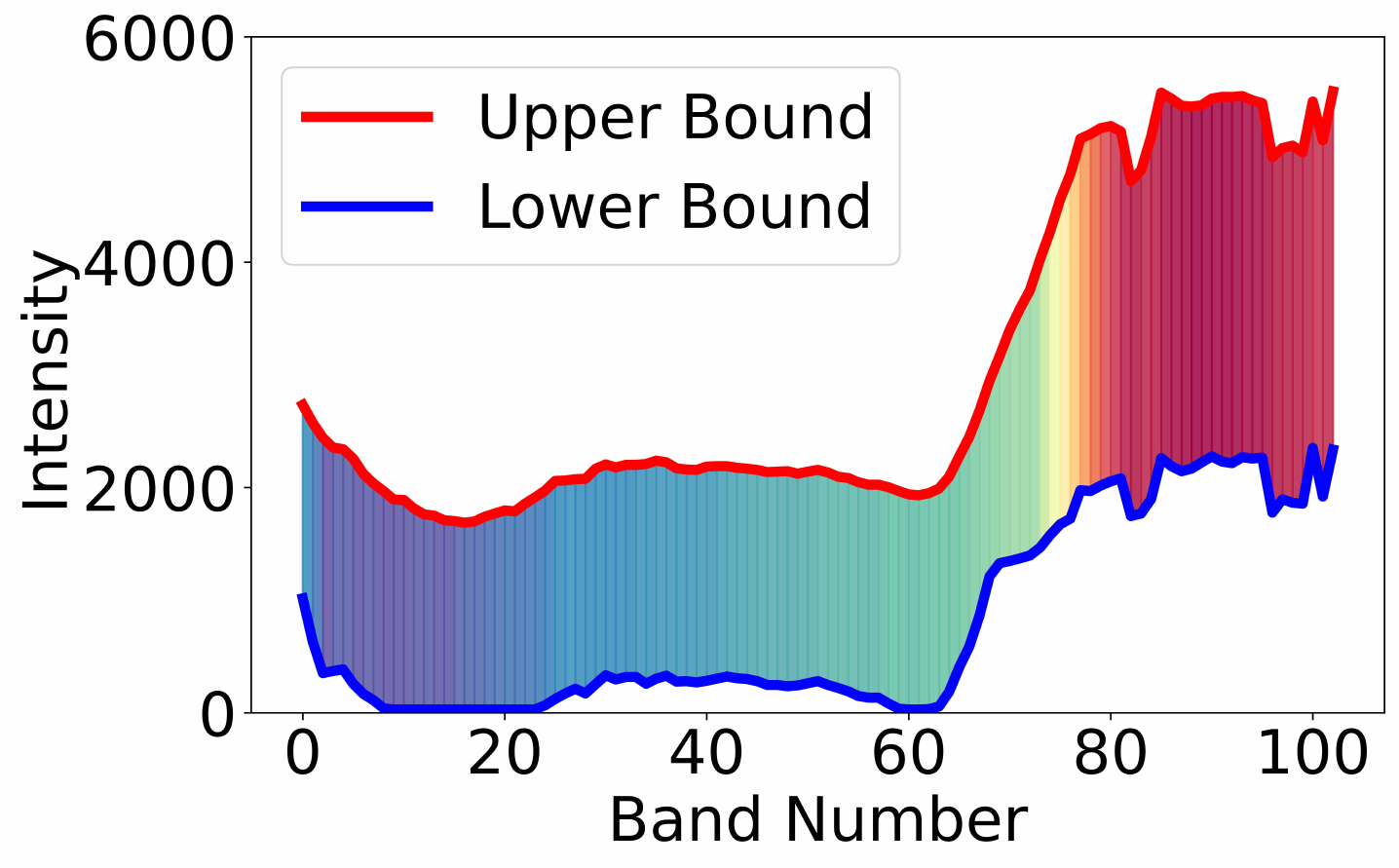}}   
    \subfloat[PaviaU–Bare soil (Benign)]{\includegraphics[width=0.25\linewidth]{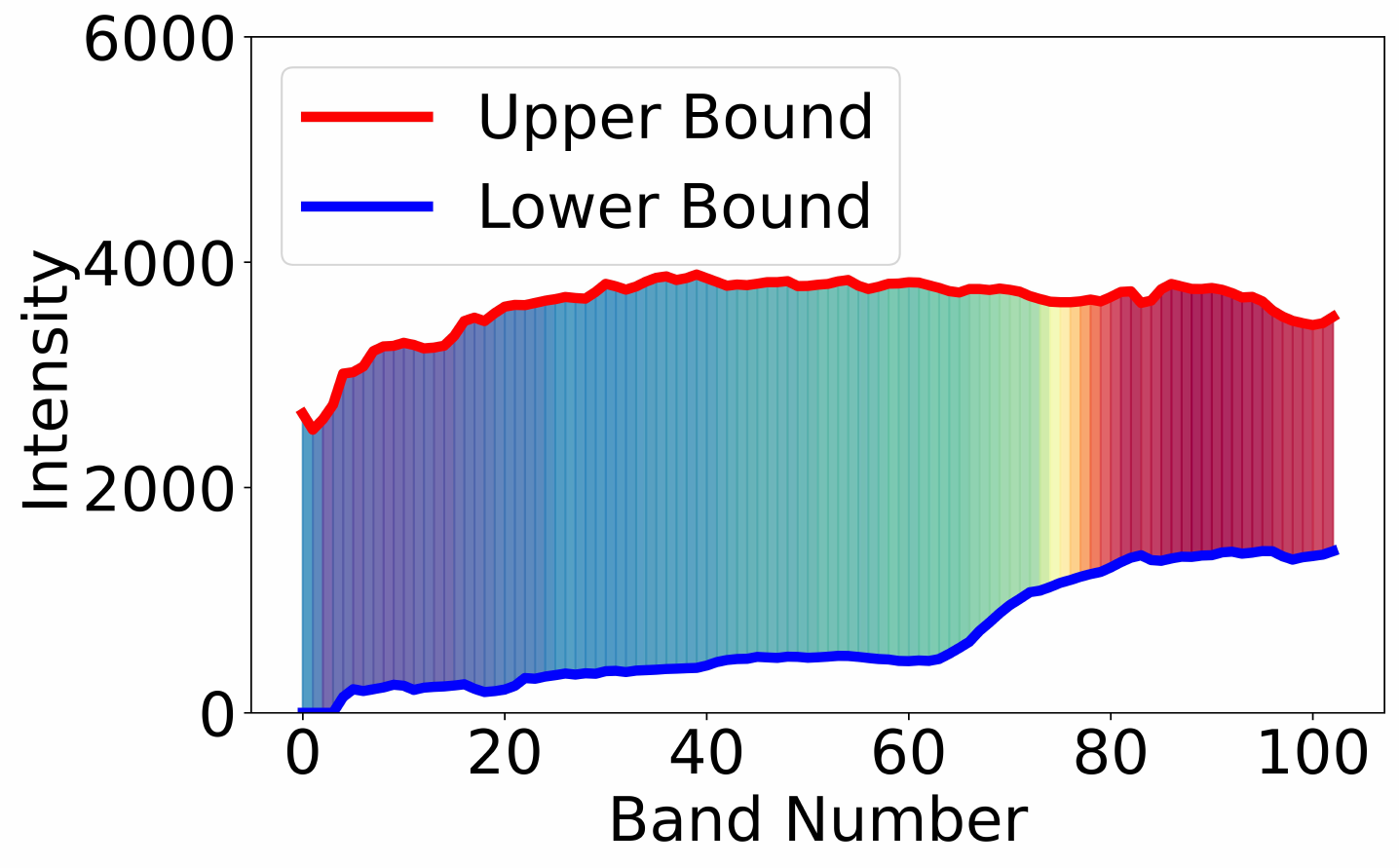}}
    \subfloat[PaviaU–Bare soil (Adv)]{\includegraphics[width=0.25\linewidth]{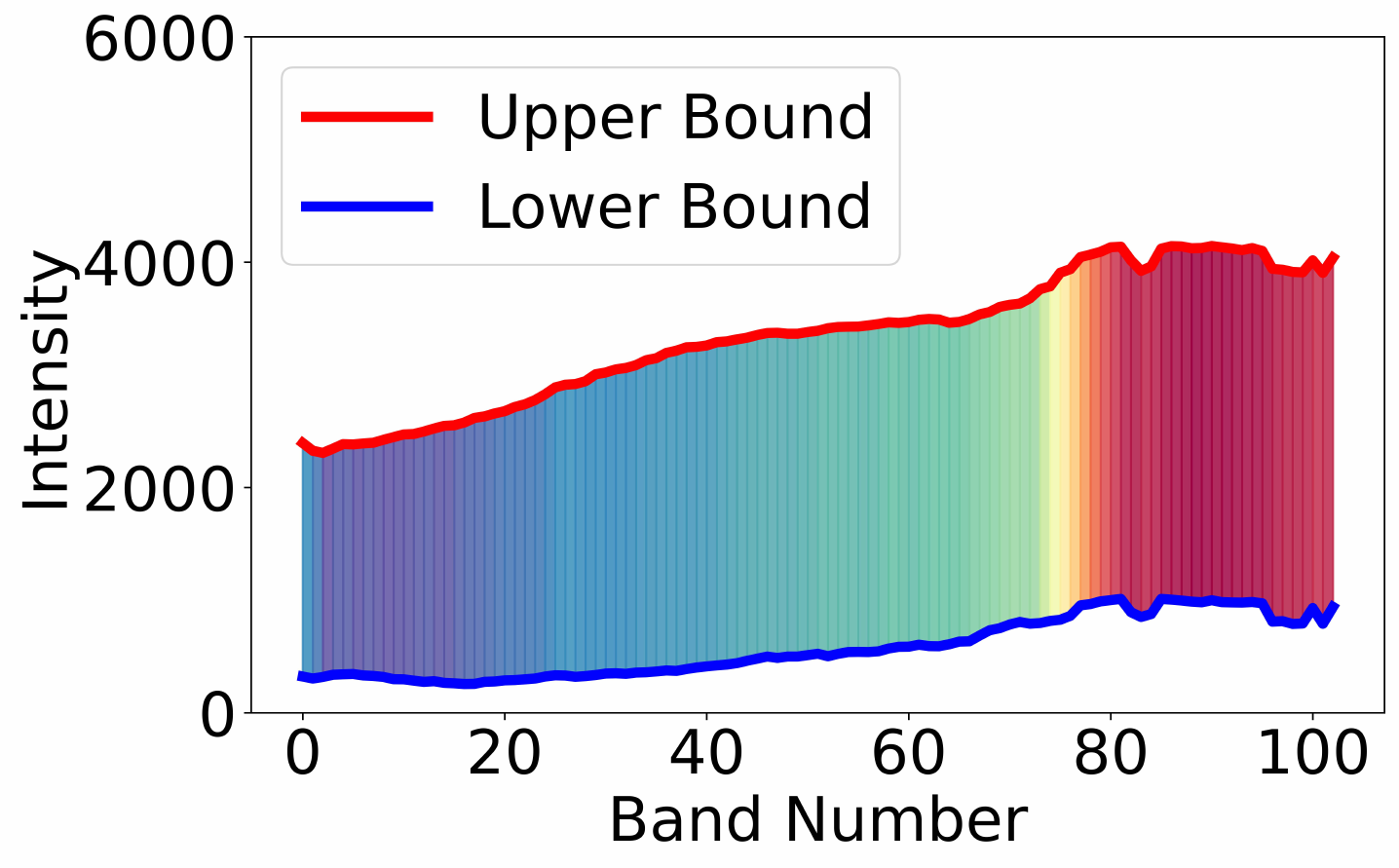}}\\
    \subfloat[Salinasn–Grapes untrained (Benign)]{\includegraphics[width=0.25\linewidth]{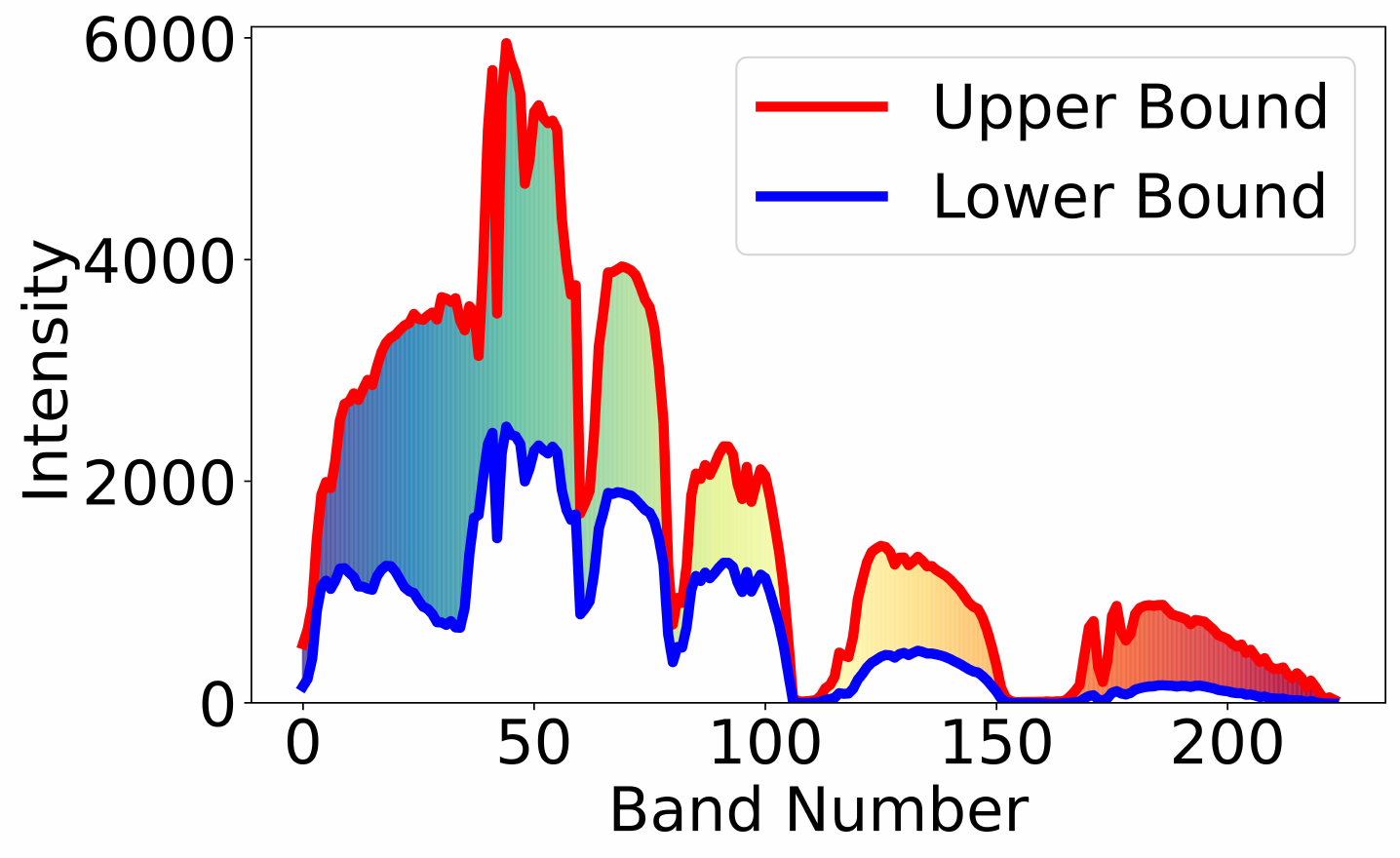}}
    \subfloat[Salinas–Grapes untrained (Adv)]{\includegraphics[width=0.25\linewidth]{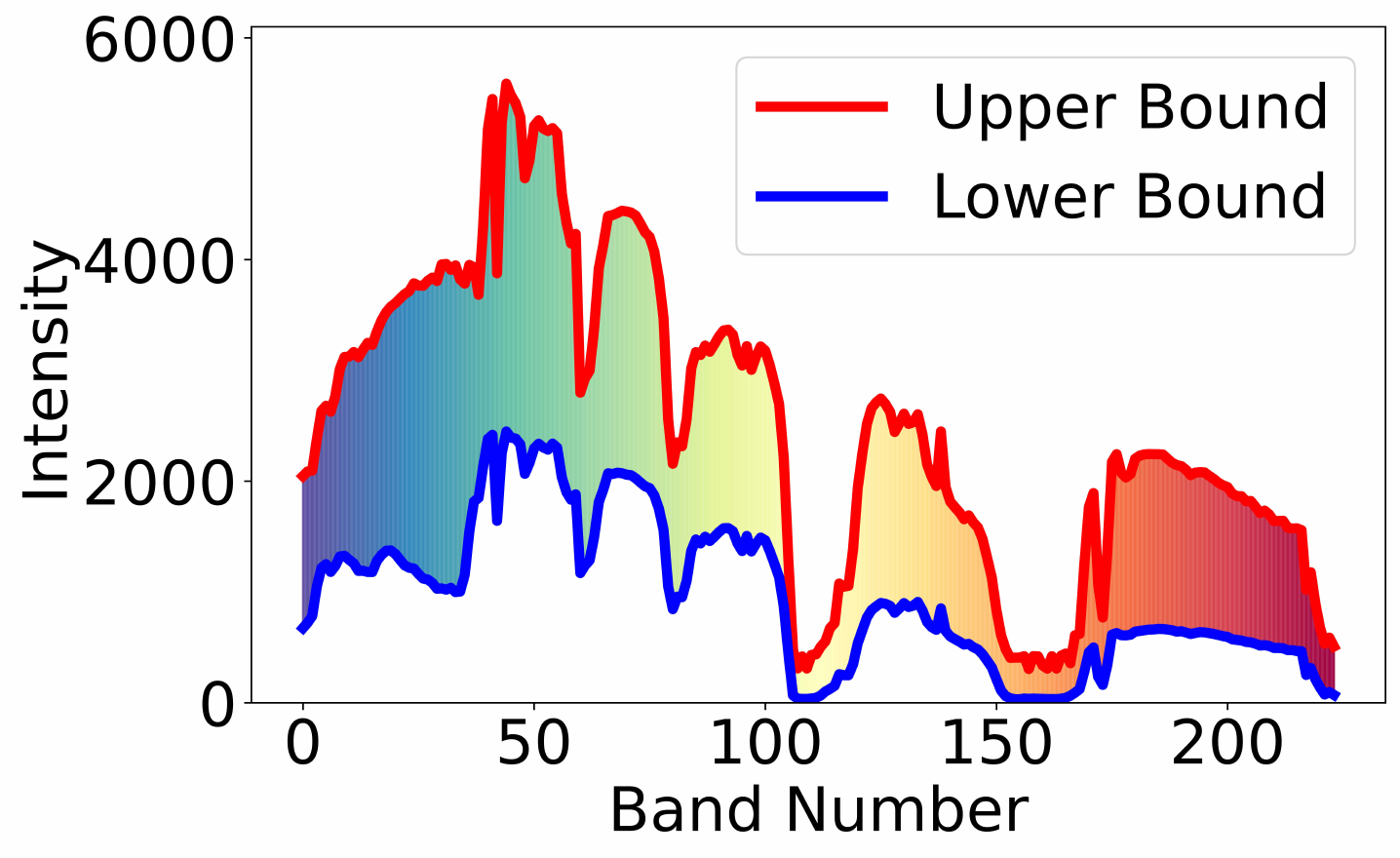}}
    \subfloat[Salinas–Fallow (Benign)]{\includegraphics[width=0.25\linewidth]{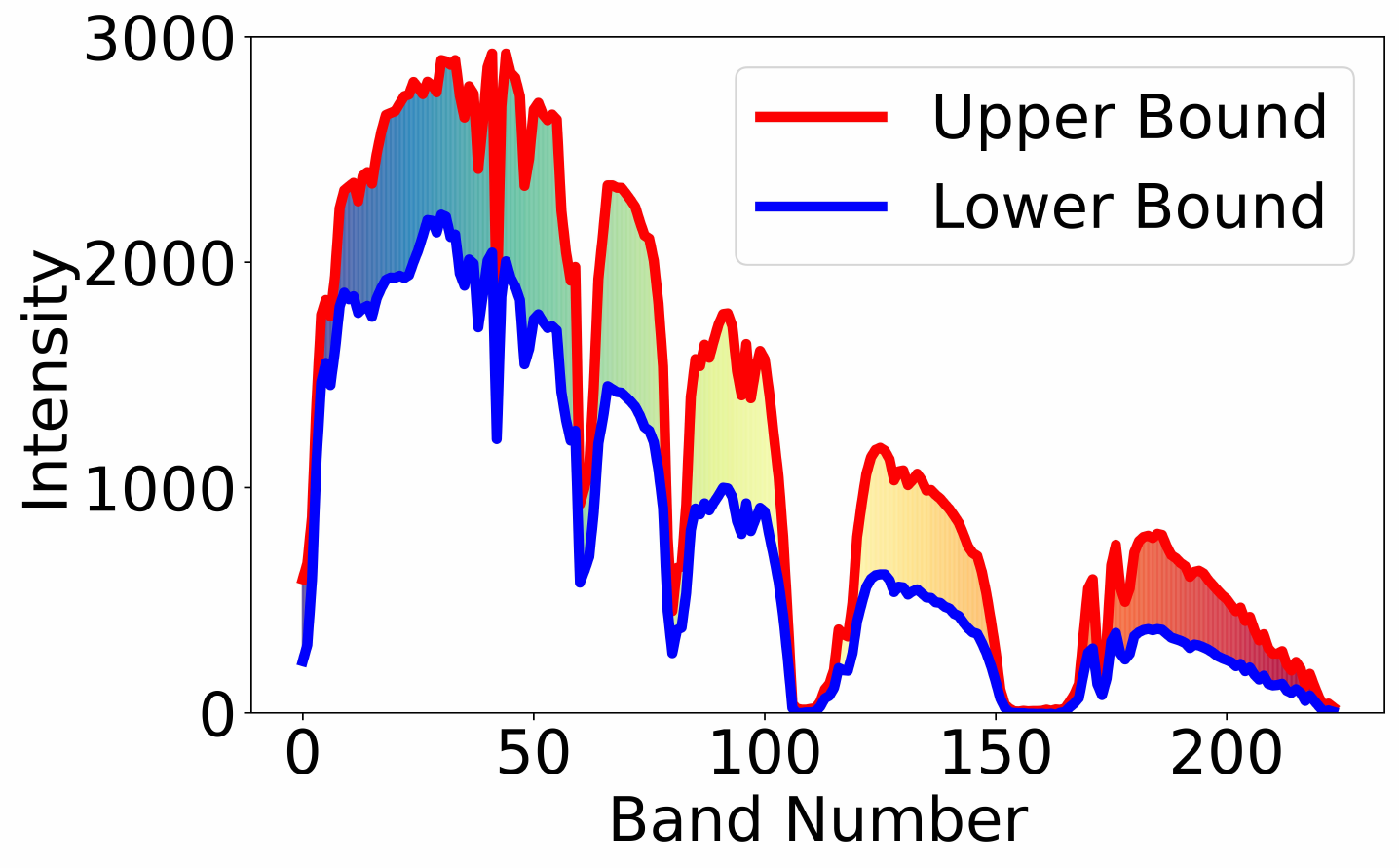}}
    \subfloat[Salinas–Fallow (Adv)]{\includegraphics[width=0.25\linewidth]{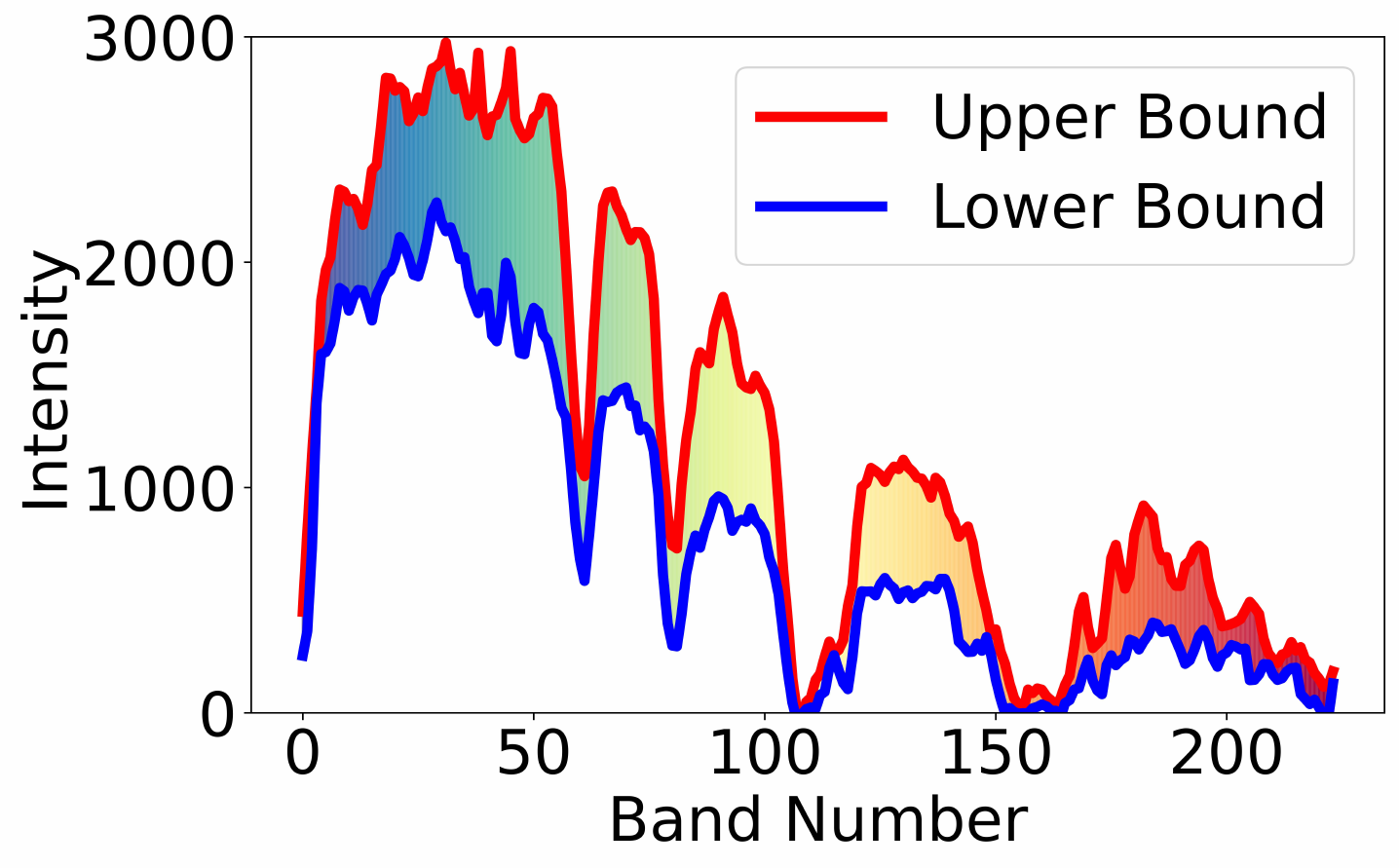}}\\ 
    \subfloat[Houston–Grass stressed (Benign)]{\includegraphics[width=0.25\linewidth]{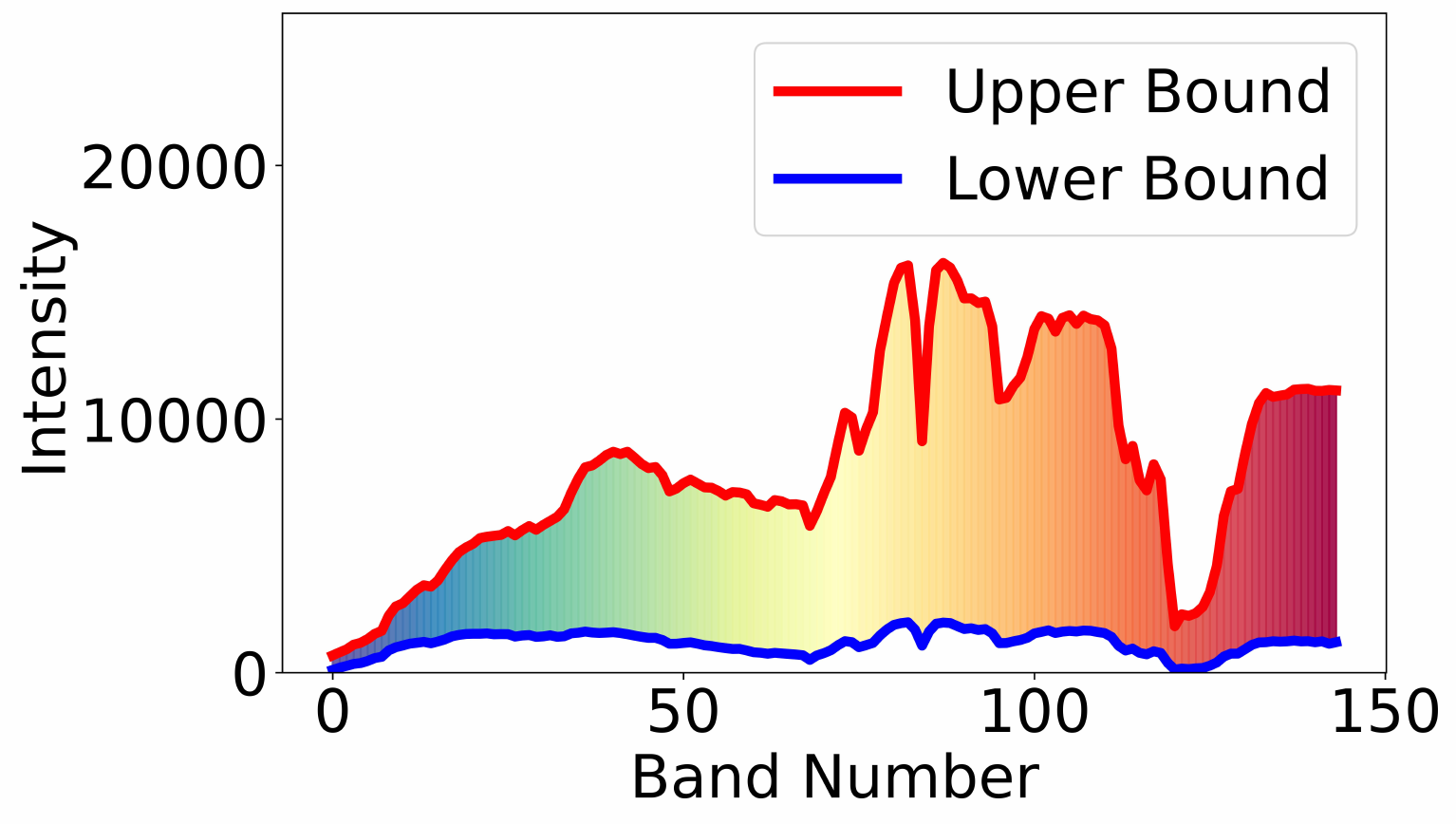}}
    \subfloat[Houston–Grass stressed (Adv)]{\includegraphics[width=0.25\linewidth]{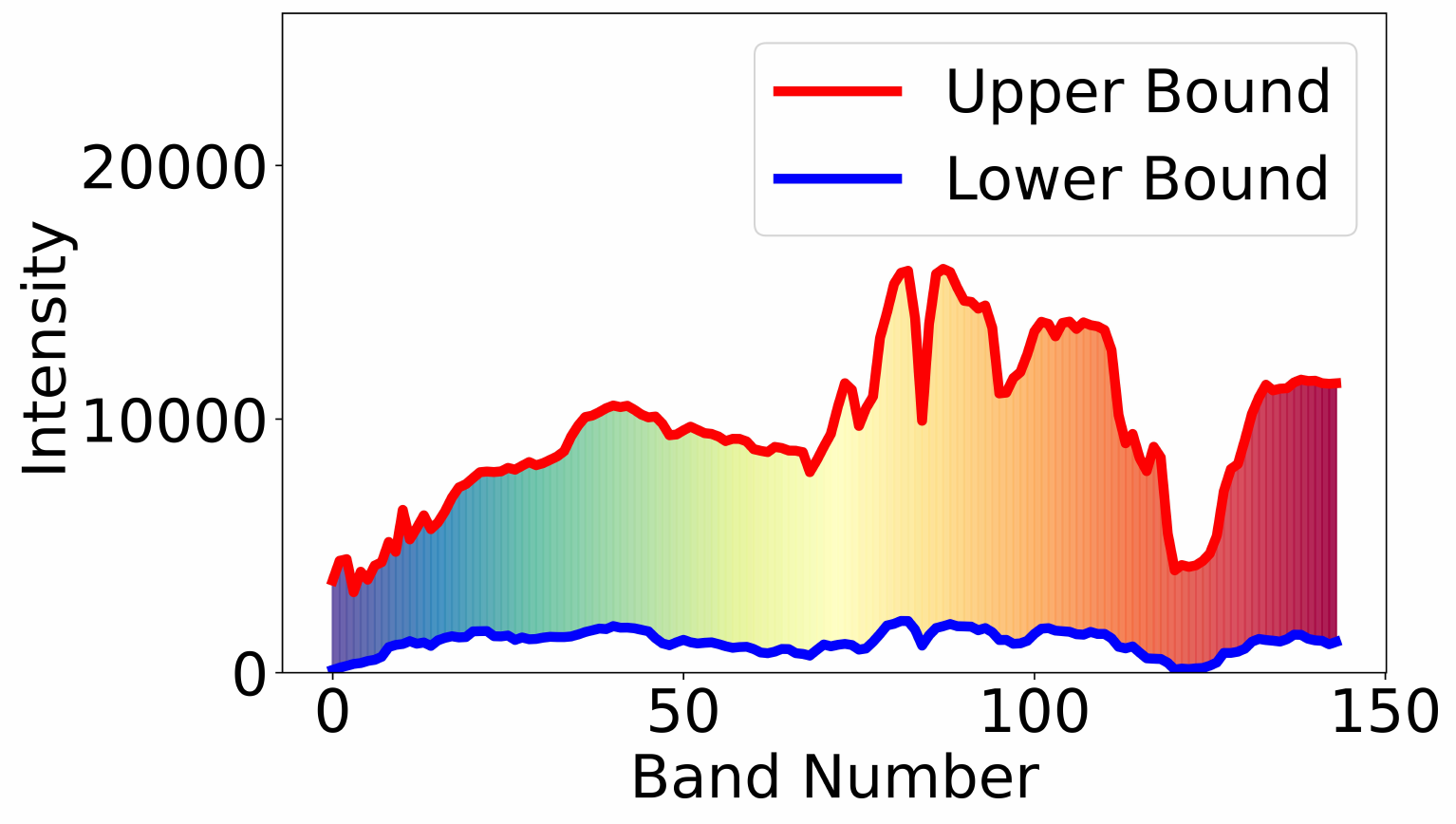}}
    \subfloat[Houston–Highway (Benign)]{\includegraphics[width=0.25\linewidth]{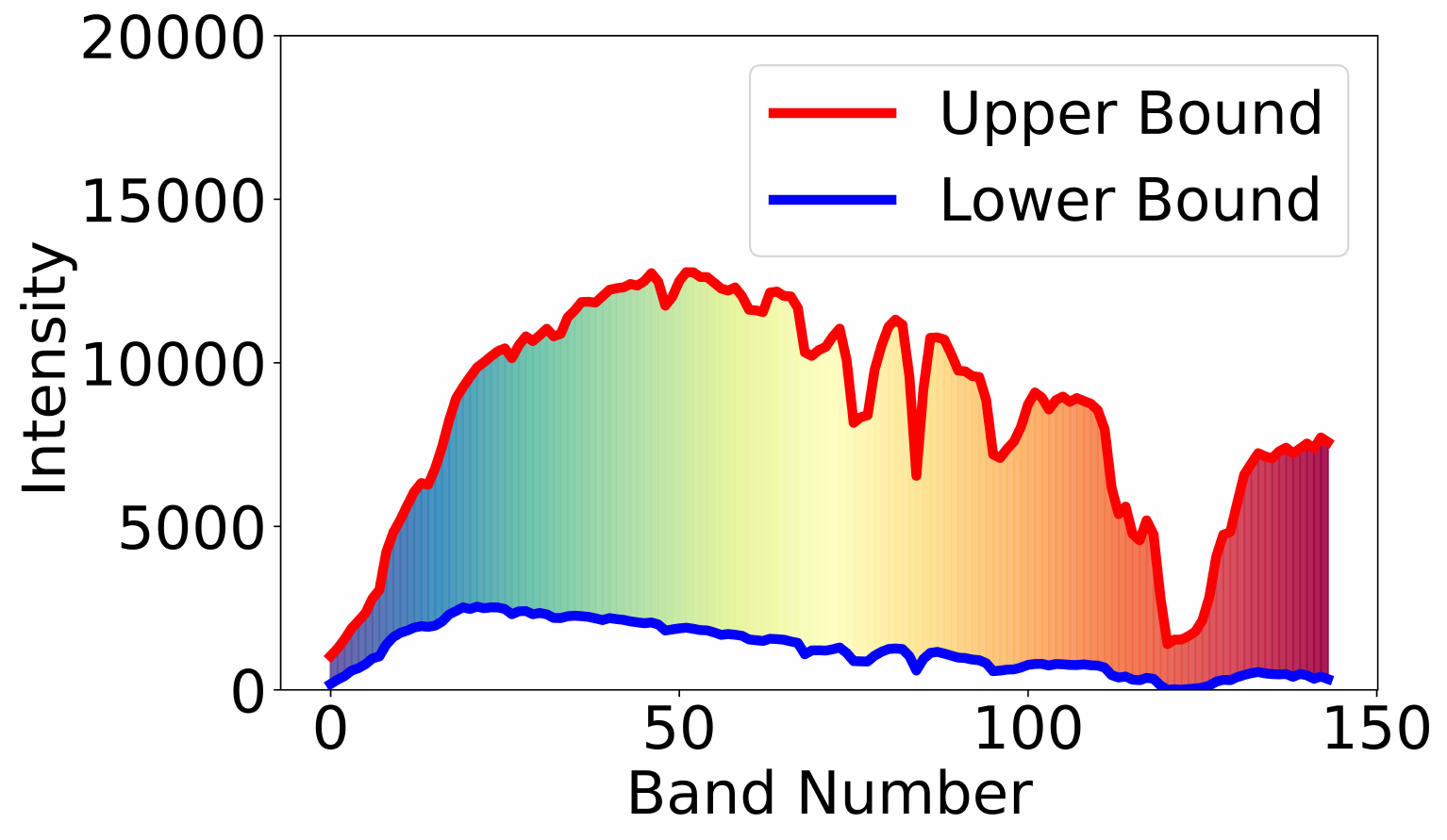}}
    \subfloat[Houston–Highway (Adv)]{\includegraphics[width=0.25\linewidth]{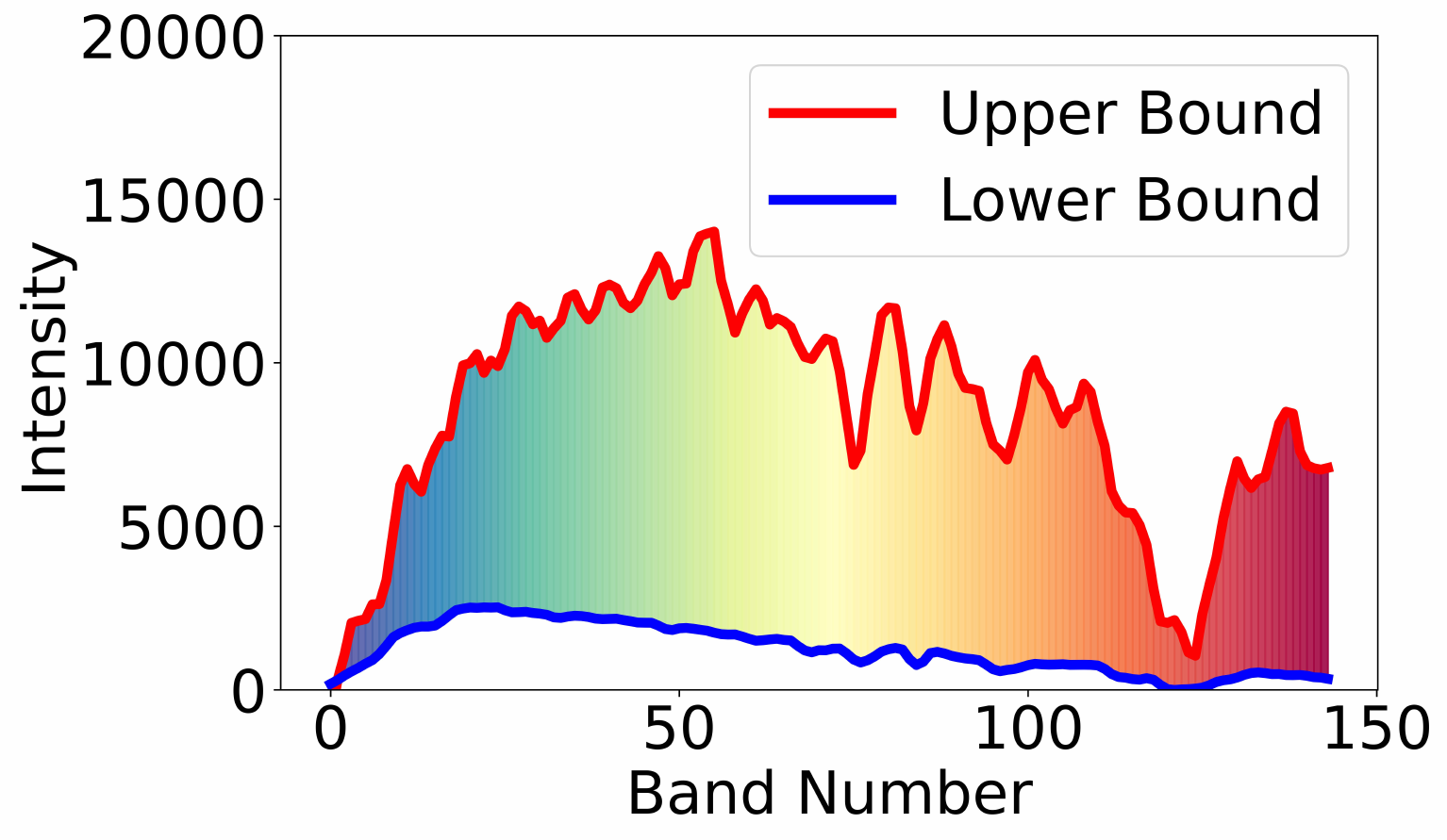}}    
    \caption{Spectral analysis of representative classes from three benchmark hyperspectral datasets: (a)--(d) PaviaU, (e)--(h) Salinas, and (i)--(l) Houston. `Upper Bound' and `Lower Bound' indicate the maximum and minimum spectral intensities, respectively. Compared to benign examples, adversarial examples exhibit elimination and distortion of spectral information, leading to semantic loss and spectral shifts. Such degradation alters decision boundaries and ultimately contributes to misclassification concentration phenomenon.}
    \label{fig1-1}
\end{figure*}

\paragraph{Spectral Analysis of Benign Examples}
We first analyze the spectral characteristics of benign examples from the \textit{Meadows} and \textit{Bare soil} classes. As shown in Fig.~\ref{fig1-1}(a), the benign spectral curve of \textit{Meadows} exhibits relatively low reflectance in the visible bands, followed by a pronounced increase in the near-infrared region, which is consistent with typical vegetation spectral responses. In contrast, the \textit{Bare soil} class in Fig.~\ref{fig1-1}(c) presents a comparatively smooth and stable spectral profile across bands, without prominent peaks or absorption valleys. These distinct spectral patterns provide clear class separability in the spectral domain, enabling accurate classification under benign conditions.

\paragraph{Spectral Analysis of Adversarial Examples}
When adversarial perturbations are introduced, the spectral semantics of both classes are substantially degraded. As shown in Fig.~\ref{fig1-1}(b), adversarial examples of the \textit{Meadows} class exhibit attenuated and shifted spectral responses, where characteristic vegetation-related patterns become weakened and less distinguishable, though not entirely removed. Such deformations indicate that adversarial perturbations distort class-dependent spectral structures, thereby reducing their discriminative power.

In comparison, the \textit{Bare soil} class experiences more severe spectral degradation. As illustrated in Fig.~\ref{fig1-1}(d), its adversarial examples show a pronounced suppression of discriminative responses in the visible bands, causing their spectral profiles to increasingly resemble those of vegetation classes. This effect suggests that adversarial perturbations can eliminate critical spectral semantics for certain classes, leading to a collapse of inter-class spectral separability and a shift of feature distributions toward dominant categories, which in turn increases the likelihood of misclassification. Consistent phenomena are also observed on the Salinas and Houston datasets, as shown in Fig.~\ref{fig1-1}(e)--(l), indicating that adversarial perturbations systematically induce spectral semantic distortion and elimination across different hyperspectral scenes.

\begin{figure}[h]
    \centering
    \subfloat[Salinas-Grapes untrained (Benign)]{\includegraphics[width=0.5\linewidth]{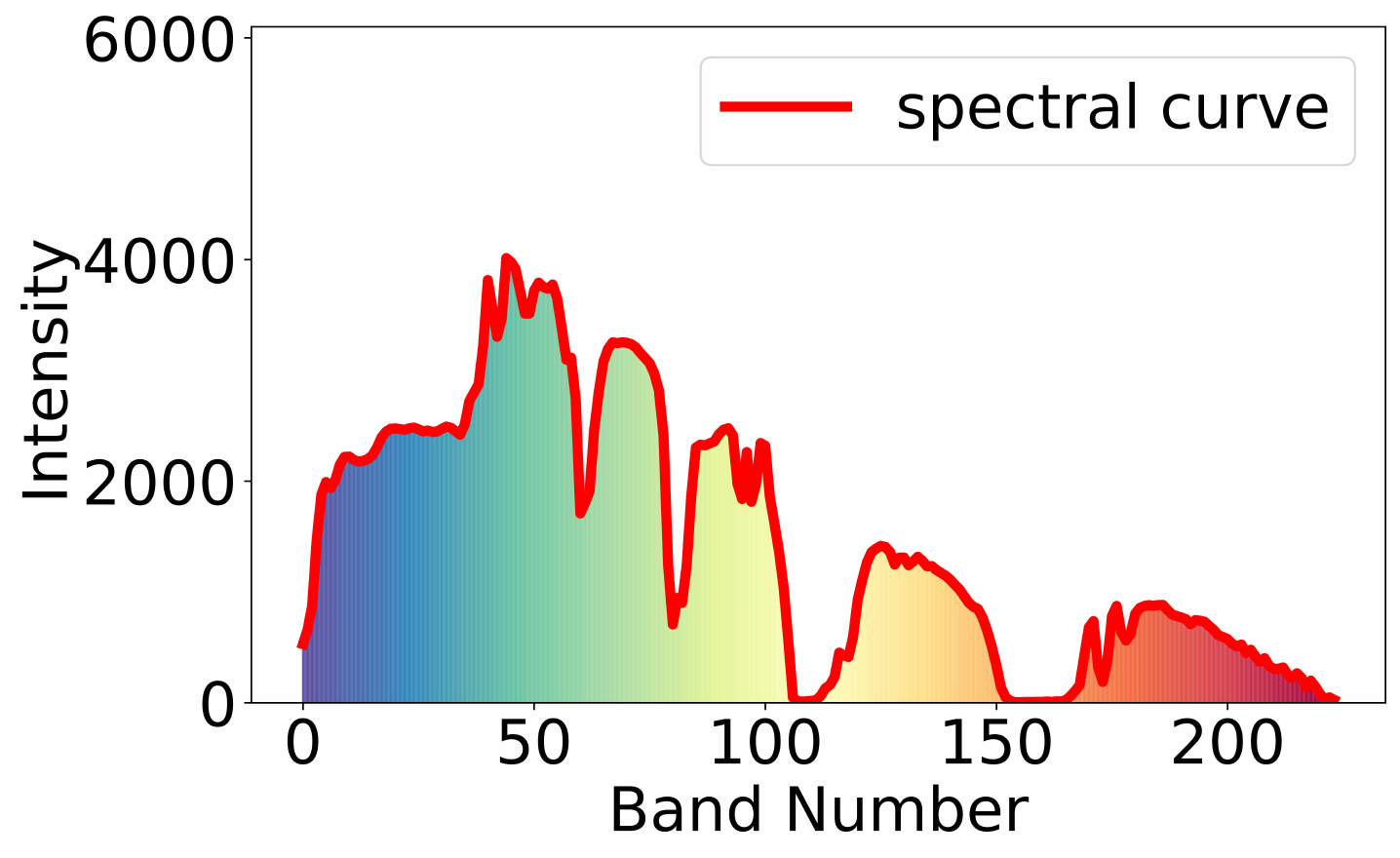}}
    \subfloat[Salinas-Grapes untrained (Adv)]{\includegraphics[width=0.5\linewidth]{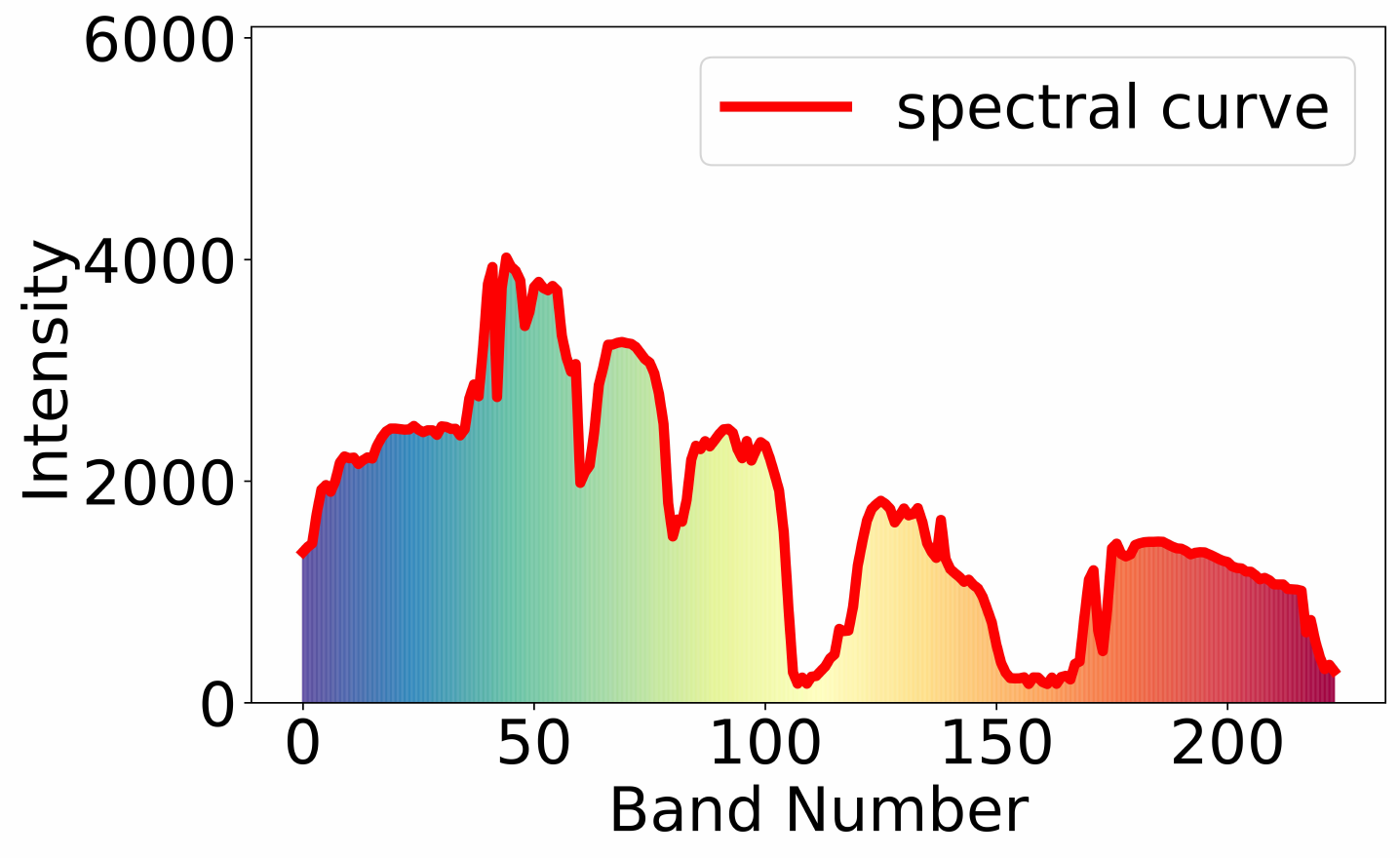}}\\
    \subfloat[Salinas-Fallow (Benign)]{\includegraphics[width=0.5\linewidth]{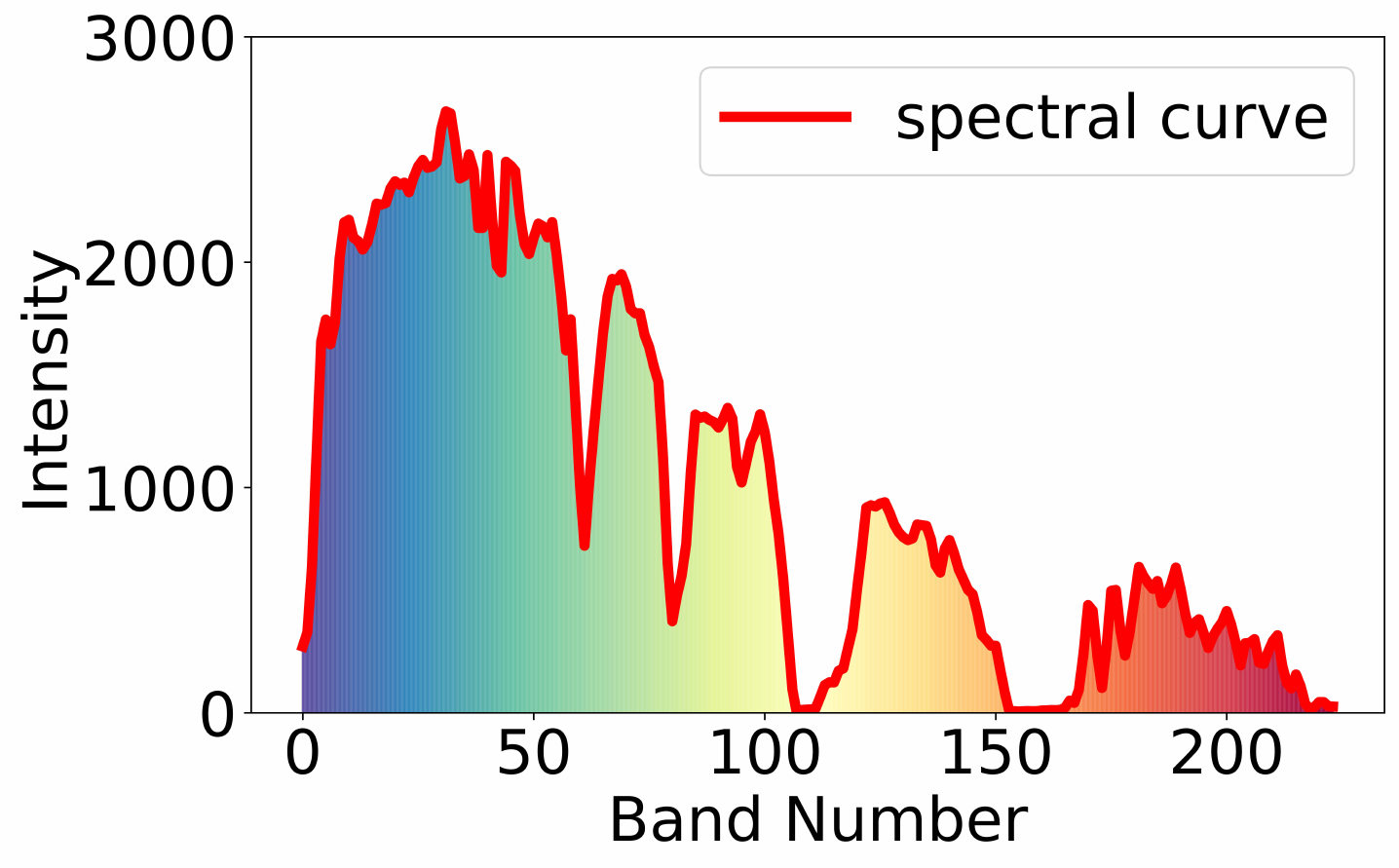}}
    \subfloat[Salinas-Fallow (Adv)]{\includegraphics[width=0.5\linewidth]{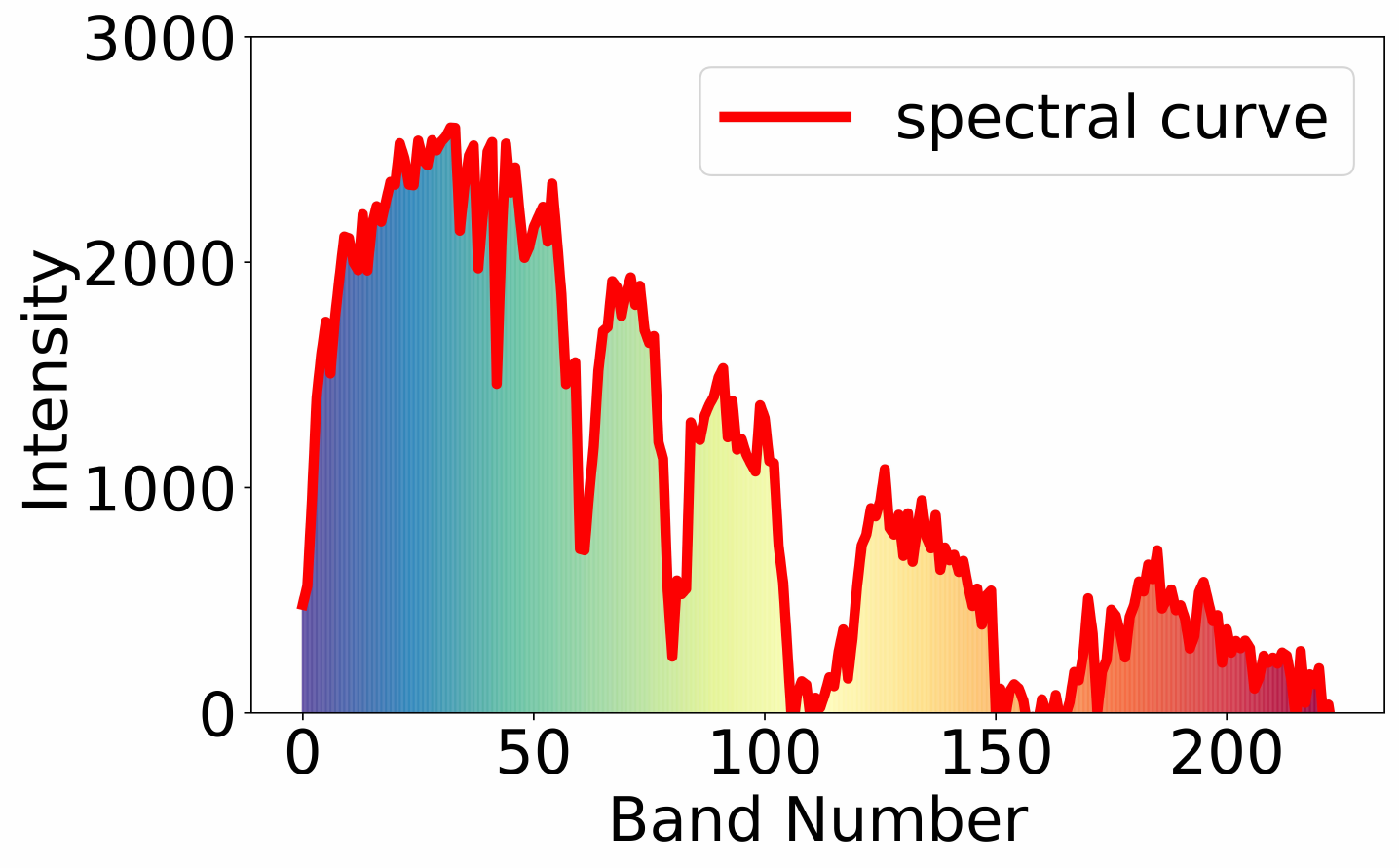}}
    \caption{Sample-level spectral comparison on the Salinas dataset.
    Compared with benign samples, adversarial examples exhibit reduced spectral smoothness and pronounced sawtooth-like fluctuations, indicating disrupted spectral continuity across adjacent bands.}
    \label{fig1-2}
\end{figure}

\paragraph{Sample-Level Spectral Smoothness Analysis}
We further conduct a sample-level spectral analysis. As shown in Fig.~\ref{fig1-2}, adversarial examples exhibit reduced spectral smoothness and pronounced sawtooth-like fluctuations compared to benign examples. Such perturbations disrupt the strong correlations between adjacent spectral bands that are intrinsic to hyperspectral data. The accumulation of such feature variations leads to the elimination and distortion of spectral semantic information. Since convolutional neural networks primarily exploit local spectral–spatial continuity for feature extraction, this disruption biases the learned representations, thereby contributing to the occurrence of the misclassification concentration phenomenon. Additional qualitative results on other datasets exhibit consistent observations and are provided in \textbf{Fig.~1 of the Appendix}.

\section{Proposed Method}\label{Methods}
\subsection{Spectral Angle Analysis and Motivation}
To further investigate the underlying cause of the misclassification concentration phenomenon, we conduct a spectral angle analysis to quantify the semantic consistency and separability among different classes under both benign and adversarial conditions. Specifically, given a hyperspectral image $\mathbf{X} \in \mathbb{R}^{C \times H \times W}$ and its corresponding ground-truth label map $\mathbf{Y} \in \{1, 2, \dots, K\}^{H \times W}$, we first compute the average spectral signature for each class:
\begin{equation}
    \boldsymbol{\mu}_k = \frac{1}{N_k} \sum_{(i,j) \in \mathcal{I}_k} \mathbf{X}_{:,i,j},
\end{equation}
where $\mathcal{I}_k = \{(i,j) \mid \mathbf{Y}_{i,j} = k\}$ denotes the set of pixel indices belonging to class $k$, and $N_k = |\mathcal{I}_k|$ is the total number of pixels in that class. For each class, we then calculate its average spectral angle with respect to all other classes to assess inter-class spectral similarity:
\begin{equation}
    \theta_k = \frac{1}{K - 1} \sum_{\substack{l = 1 \\ l \neq k}}^{K} \cos^{-1} \left( \frac{\langle \boldsymbol{\mu}_k, \boldsymbol{\mu}_l \rangle}{\|\boldsymbol{\mu}_k\|_2 \cdot \|\boldsymbol{\mu}_l\|_2} \right),
\end{equation}
where $\langle \cdot, \cdot \rangle$ denotes the dot product, and $\| \cdot \|_2$ denotes the $\ell_2$ norm. This formulation, known as the Spectral Angle Distance (SAD), provides a rotation-invariant metric to evaluate spectral distinction across classes and is widely used in hyperspectral analysis due to its robustness against illumination differences.

We apply the above procedure separately to the benign and adversarial examples to quantify the spectral degradation introduced by adversarial perturbations. The resulting spectral angle vectors reflect the average angular separation between each class and all other classes in the spectral domain. To facilitate a clearer understanding of the perturbation impact, we visualize these spectral angles for all nine classes under both benign and adversarial conditions. 

\begin{figure}[tbp]
    \centering
    \subfloat[Benign examples]{\includegraphics[width=0.5\linewidth]{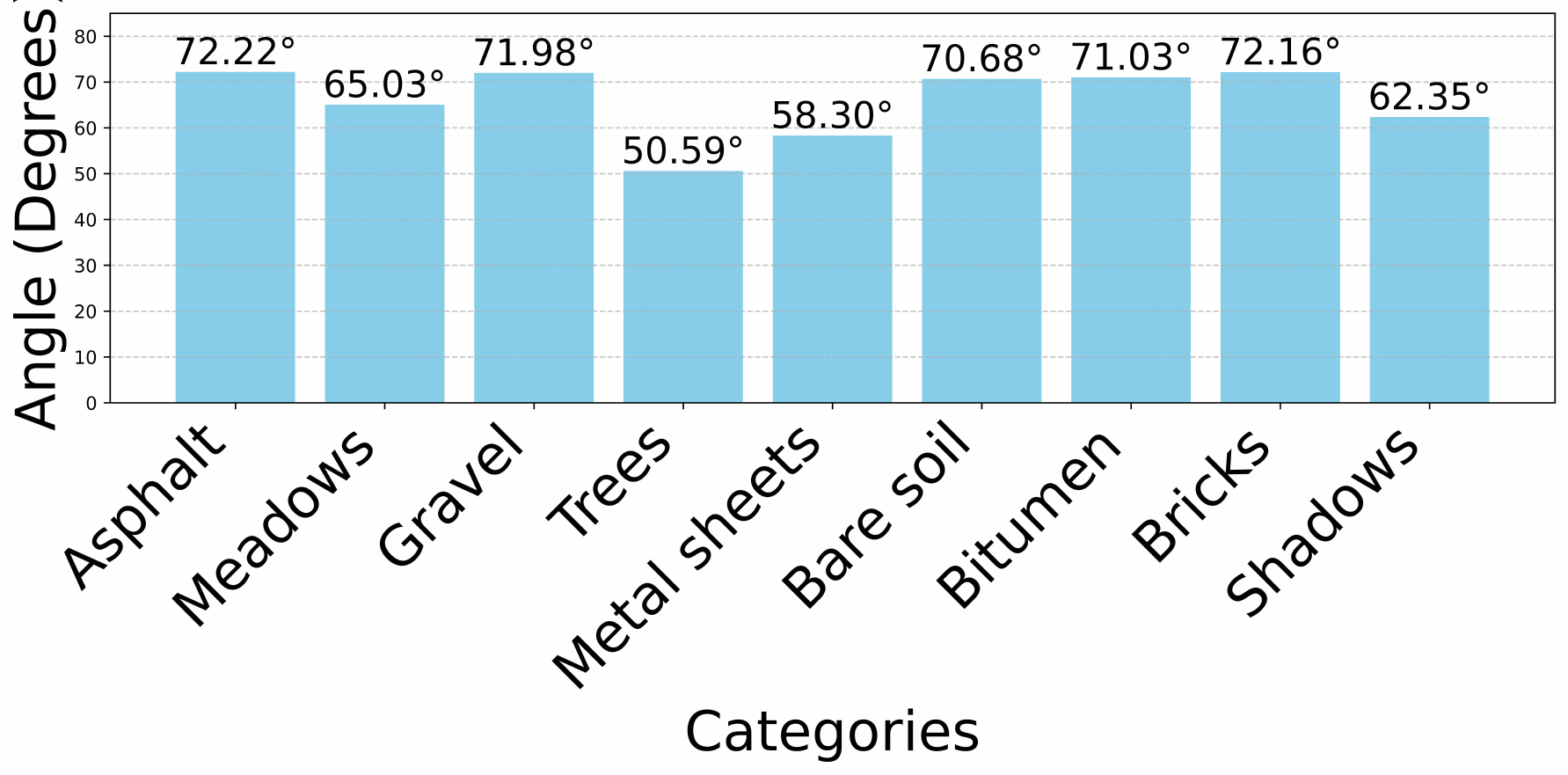}}
    \subfloat[Adversarial examples]
    {\includegraphics[width=0.5\linewidth]{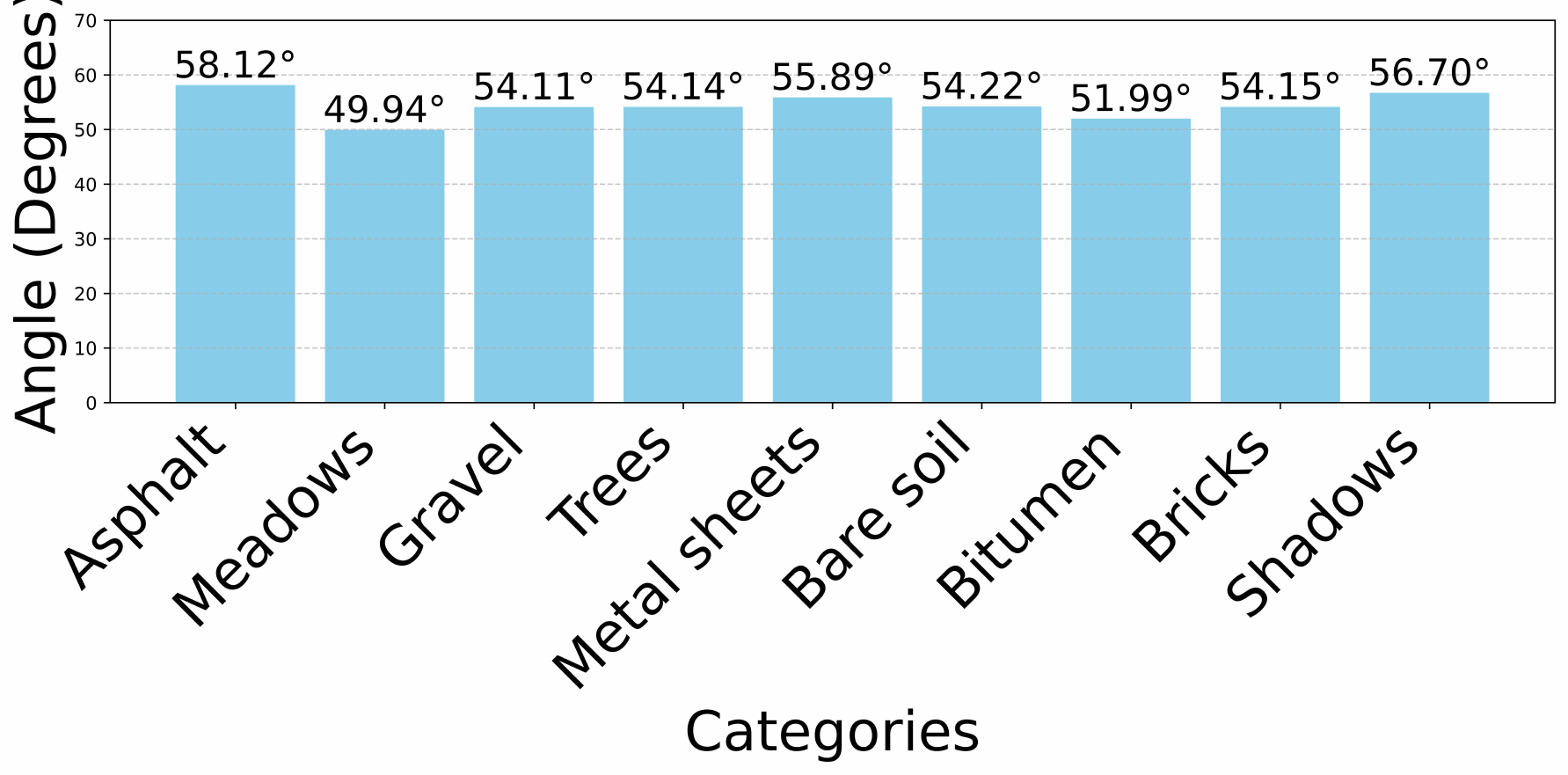}}\\
    \subfloat[Benign examples (radar)]
    {\includegraphics[width=0.5\linewidth]{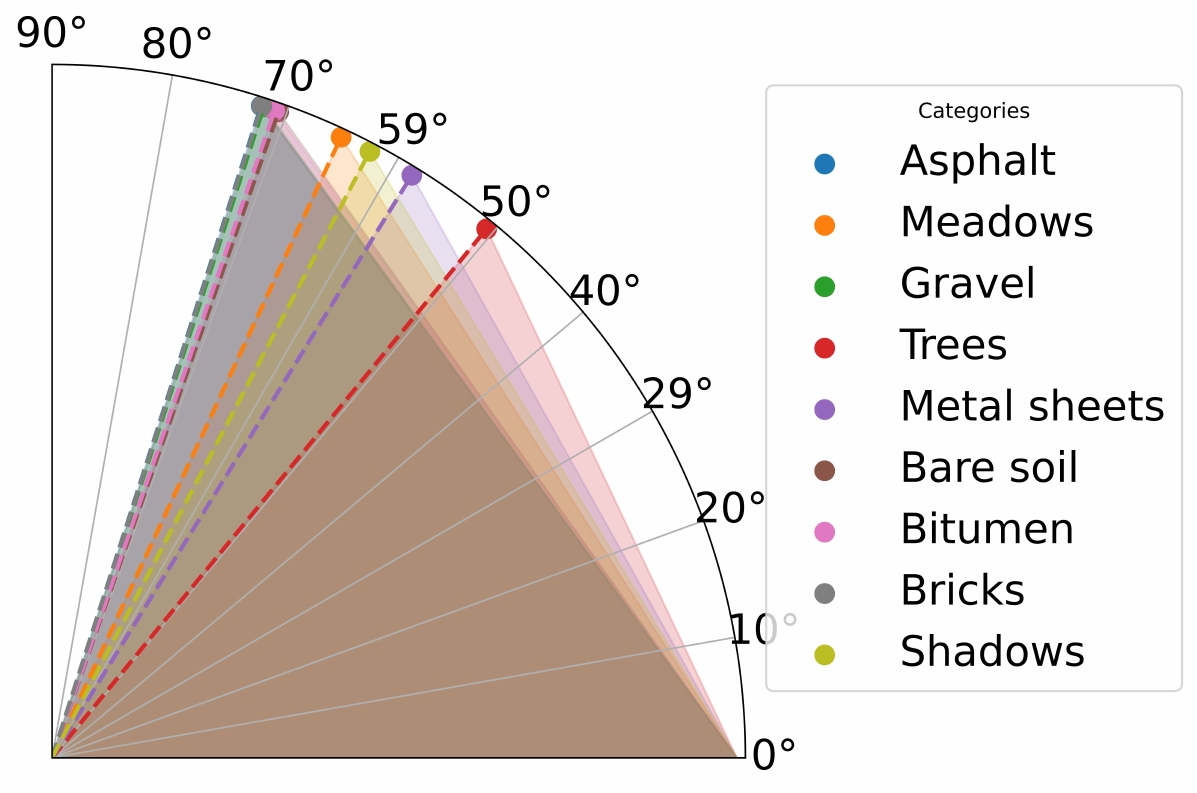}}
    \subfloat[Adversarial examples (radar)]{\includegraphics[width=0.5\linewidth]{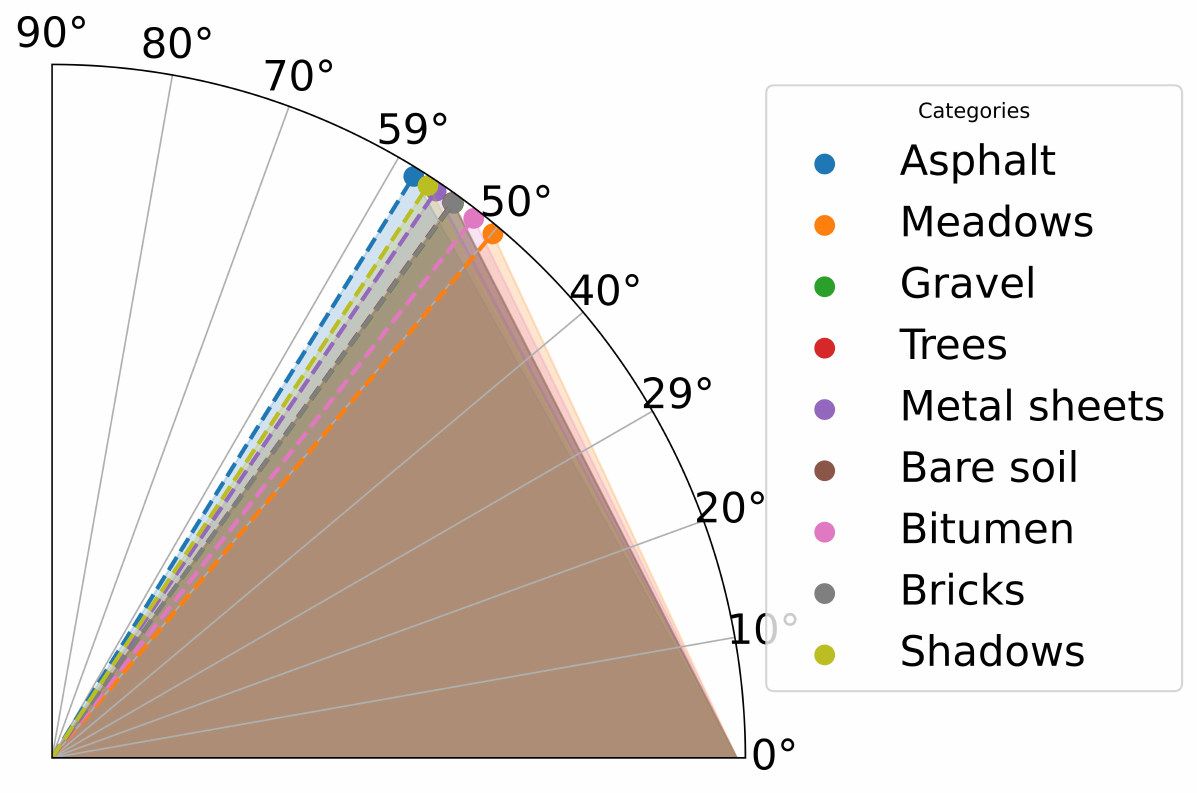}}
    \caption{Class-wise average spectral angles computed from benign and adversarial examples. (a) and (b) illustrate bar charts, while (c) and (d) show the corresponding polar representations. Adversarial attacks compress inter-class spectral angle differences, distort the original spectral semantics, and blur decision boundaries, leading to misclassification concentration phenomenon.}
    \label{fig2-1}
\end{figure}

As illustrated in Fig.\ref{fig2-1}, adversarial attacks tend to narrow the spectral angle differences among classes. For instance, the angles for \textit{Meadows} (class 2) and \textit{Bare soil} (class 6) shift from $65.03^\circ$ to $49.94^\circ$ and $70.68^\circ$ to $54.22^\circ$, respectively. This narrowing indicates a reduction in inter-class separability and a distortion of the original spectral semantics. Consequently, the decision boundaries between certain classes become less distinguishable, disproportionately reducing the classification accuracy of vulnerable classes and giving rise to a pronounced misclassification phenomenon.

Motivated by these findings, we propose a novel method named Adversarial Training with Hyperspectral-Aware Rare-class Learning (AT-HARL), which explicitly addresses this misclassification concentration phenomenon by enhancing the angular discriminability and adversarial robustness of class-wise spectral representations.

\subsection{AT-HARL}

Inspired by \cite{TAET}, to enhance the robustness of HSI classification under adversarial perturbations and address the misclassification concentration phenomenon, AT-HARL introduces two tailored loss functions: \textit{Spectral-Aware Balanced Cross-class Loss (SABCL)} and \textit{Rare Class Spectral Emphasis Loss (RCSEL)}. These components leverage spectral-domain knowledge and class distribution statistics to adaptively reweight the training objective.

\paragraph{Spectral-Aware Balanced Cross-class Loss (SABCL)}

As observed in the preceding analysis, adversarial attacks compress the spectral angle differences between classes. To address this, SABCL incorporates prior knowledge of the original spectral angles to reweight class-specific loss components. By explicitly compensating for the reduction in inter-class spectral distinctions, SABCL aims to restore the underlying spectral semantics and reinforce the robustness of decision boundaries. The loss is defined as:

\begin{equation}
L_{\text{SABCL}} = \frac{1}{C} \sum_{c=1}^{C} \left( \frac{L_c}{\text{SA}_c} \right),
\end{equation}
where $C$ denotes the total number of classes, $L_c$ represents the individual loss for class $c$, and $\text{SA}_c$ denotes the average spectral angle between class $c$ and all other classes.

The spectral angle $\text{SA}_c$ is computed using mean spectral signatures across training examples, serving as a proxy for spectral distinctiveness. By inversely scaling the loss with $\text{SA}_c$, SABCL increases the gradient contribution from spectrally similar classes, thereby improving intra-class compactness and inter-class separability in the spectral domain.

\paragraph{Rare Class Spectral Emphasis Loss (RCSEL)}

As demonstrated in Section~\ref{sec:spec_analysis}, compared to benign examples, the spectral curves of adversarial examples are less smooth and exhibit pronounced sawtooth patterns. The accumulation of these feature variations can eliminate or distort spectral semantic information, thereby contributing to the misclassification concentration phenomenon. To mitigate this effect, RCSEL incorporates class distribution statistics, emphasizing  misclassified classes by jointly considering class frequency and distribution ratio, thereby improving the robustness of minority class predictions. The loss is defined as:

\begin{equation}
L_{\text{RCSEL}}^{t} = \sum_{c=1}^{C} 
\frac{L_c^{t}}{N_c^{(t-1)}} \cdot \frac{1}{1 - \text{Dist}_c^{(t-1)}},
\end{equation}
where $L_c^{t}$ denotes the cumulative loss of class $c$ in the current epoch, $N_c^{(t-1)}$ is the number of examples of class $c$ in the previous epoch, and $\text{Dist}_c^{(t-1)}$ represents the proportion of misclassified examples predicted as class $c$ in the previous epoch. By amplifying the loss associated with underrepresented classes, RCSEL ensures adequate learning pressure on rare classes and prevents their semantic representations from being overwhelmed by dominant classes during AT.

\paragraph{Adversarial Training with Hyperspectral-Aware Rare-class Loss (AT-HARL)}

The final AT-HARL is a weighted sum of the three components:

\begin{equation}
L_{\text{AT-HARL}} = \beta L_{\text{SABCL}} + \gamma L_{\text{RCSEL}} + {\cal L}_{CE} 
\end{equation}

where $\beta$ and $\gamma$ are hyperparameters balancing the two loss terms, \( {\cal L}_{CE} \) represents the cross entropy loss function. This combined formulation enables AT-HARL to adaptively balance spectral discrimination and rare-class enhancement under AT.

\subsection{AT-RA}

Since the cause of the misclassification concentration phenomenon lies in the elimination or distortion of spectral semantic information in HSI, we consider introducing data augmentation methods to regularize AT. The core motivation is to strengthen the correlations between consecutive spectral bands and enrich spectral--spatial variability, thereby preserving the class-wise data distribution in the feature space and mitigating the misclassification concentration phenomenon.

Several studies \cite{47,50,48,49} have attempted to alleviate similar issues through data augmentation on optical datasets. However, whether such strategies remain effective for HSI---characterized by high-dimensional, continuous, and strongly correlated spectral signatures---remains unclear. We initially explored state-of-the-art augmentation techniques, including MixUp \cite{37}, Cutout \cite{38}, CutMix \cite{39}, and AugMix \cite{40}, but these methods did not yield satisfactory results. Although they may enhance spectral diversity, they often disrupt spatial continuity, leading to unsmooth spatial structures and the loss of spatial relationship features \cite{18}, which ultimately degrades classification performance.

\paragraph{RandAugment for Hyperspectral Images}
To address the above limitations, we adopt RandAugment \cite{42} and adapt it to the hyperspectral domain. Let
\begin{equation}
\mathcal{T} = \{T_1, T_2, \dots, T_K\}
\end{equation}
denote a set of spectral--spatial augmentation operators, where each operator $T_k$ is parameterized by an augmentation magnitude $m$. Given an input hyperspectral sample $\mathbf{x}$, RandAugment applies a randomly selected composition of transformations as
\begin{equation}
\tilde{\mathbf{x}} = T_{i_1}^{(m)} \circ T_{i_2}^{(m)} (\mathbf{x}),
\quad T_{i_1}, T_{i_2} \sim \mathcal{T}.
\end{equation}

Due to the unique properties of HSI, augmentation operations such as Posterize, Solarize, and Equalize are not applicable. We therefore retain 11 operators, including ShearX, ShearY, TranslateX, TranslateY, Rotate, Brightness, Color, Contrast, Sharpness, and AutoContrast. These transformations preserve spectral smoothness while introducing controlled spectral--spatial variability.

\paragraph{Adversarial Training With RandAugment}
Under the AT-RA framework, RandAugment is applied prior to adversarial perturbation, and adversarial examples are generated based on the augmented samples. Specifically, adversarial examples are obtained by solving
\begin{equation}
\mathbf{x}_{\text{adv}} = 
\arg\max_{\|\boldsymbol{\delta}\|\le \epsilon}
\mathcal{L}\big(f_\theta(\tilde{\mathbf{x}} + \boldsymbol{\delta}), y\big),
\end{equation}
where $f_\theta(\cdot)$ denotes the classifier parameterized by $\theta$, $y$ is the ground-truth label, and $\mathcal{L}(\cdot)$ represents the cross-entropy loss. The overall optimization objective of AT-RA is formulated as
\begin{equation}
\min_{\theta}\;
\mathbb{E}_{(\mathbf{x},y)\sim \mathcal{D}}
\Big[
\mathcal{L}\big(f_\theta(\mathbf{x}), y\big)
+
\mathcal{L}\big(f_\theta(\mathbf{x}_{\text{adv}}), y\big)
\Big].
\end{equation}

By enriching the spectral--spatial distribution of training samples, AT-RA effectively enlarges the support of each class in the feature space. This prevents adversarial examples from collapsing into a few dominant classes, thereby alleviating the misclassification concentration phenomenon while maintaining spectral continuity and spatial smoothness.

\begin{figure}[htbp]
    \centering
    \subfloat[]{\includegraphics[width=0.5\linewidth]{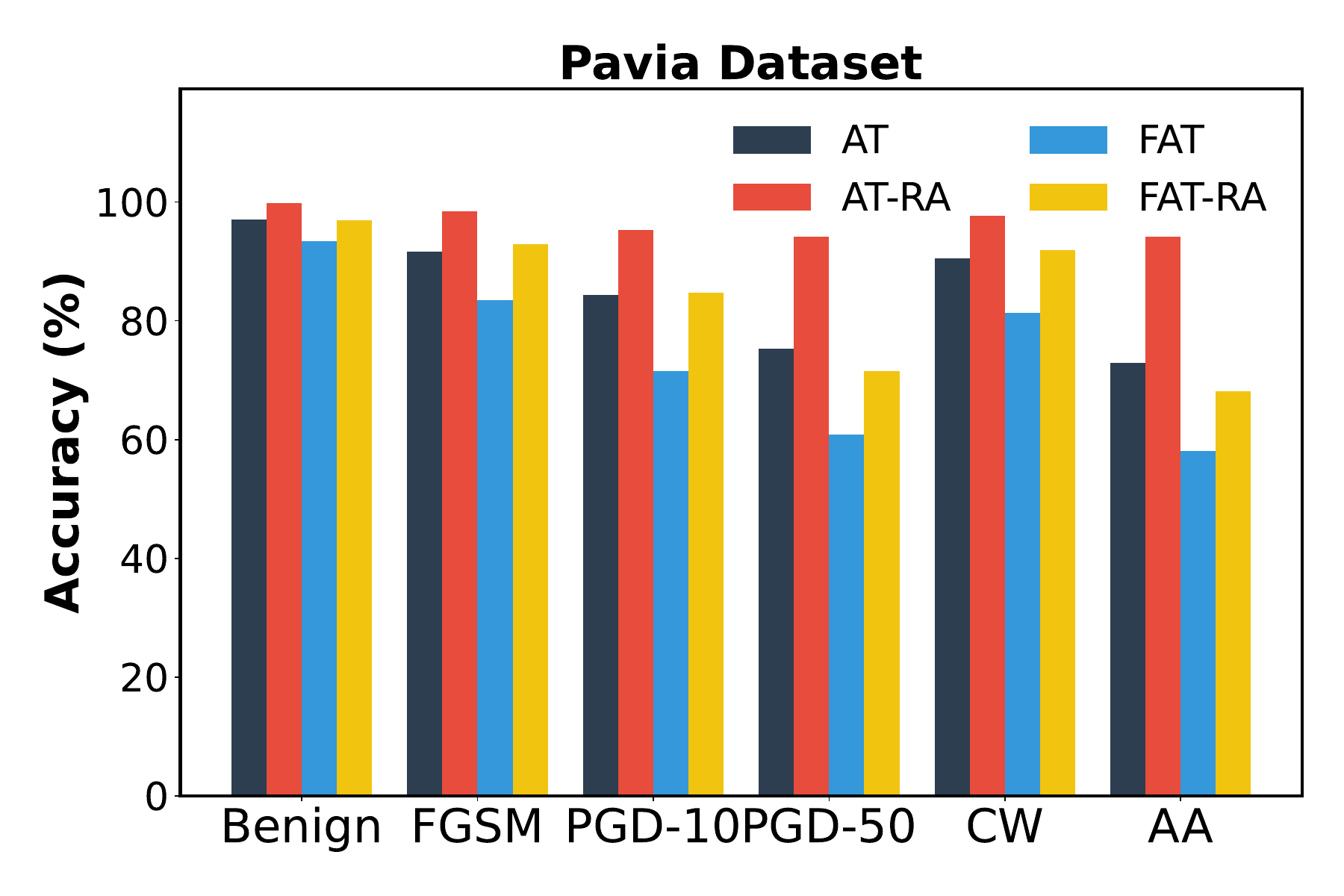}}
    \subfloat[]{\includegraphics[width=0.5\linewidth]{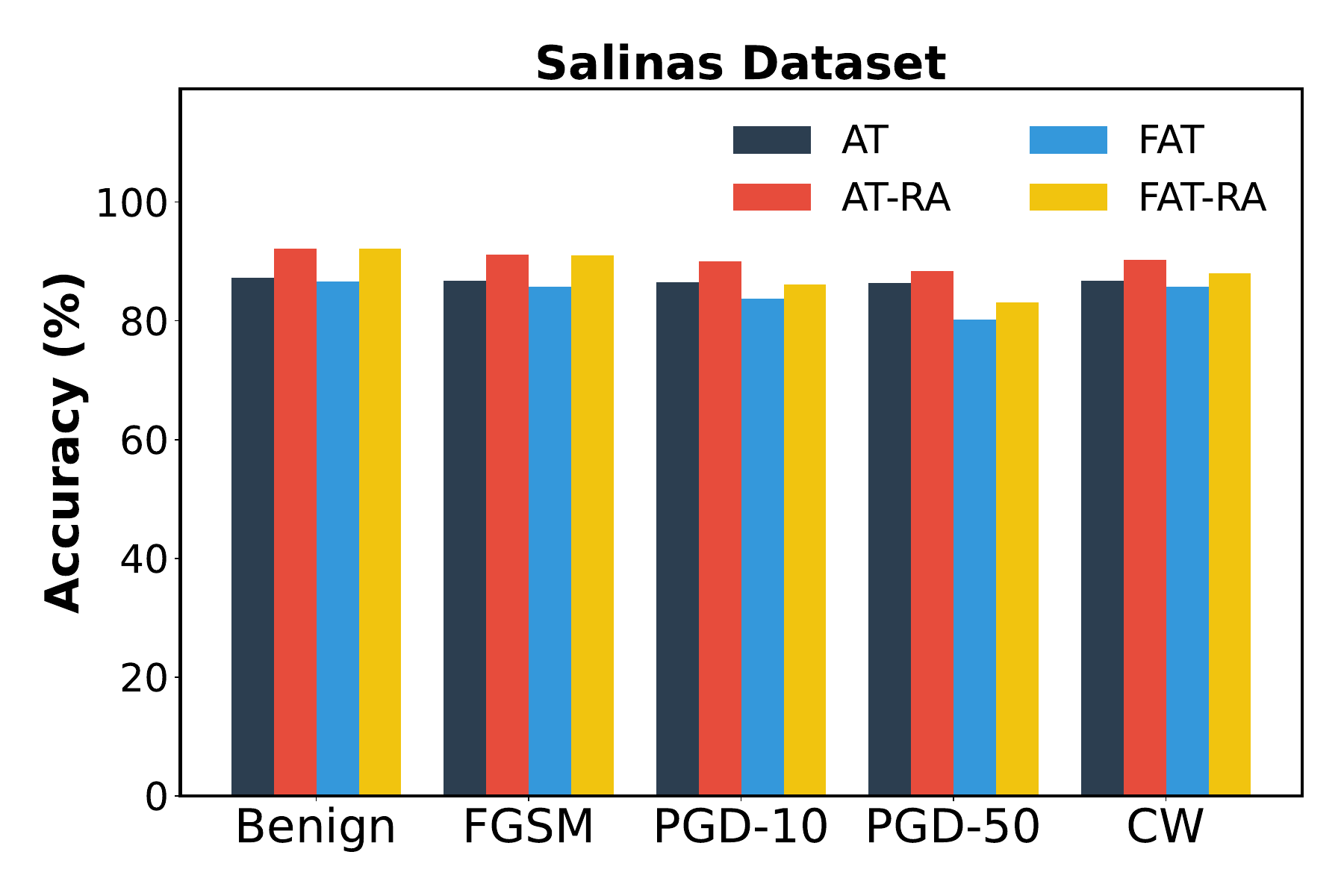}}\\
    \subfloat[]{\includegraphics[width=0.5\linewidth]{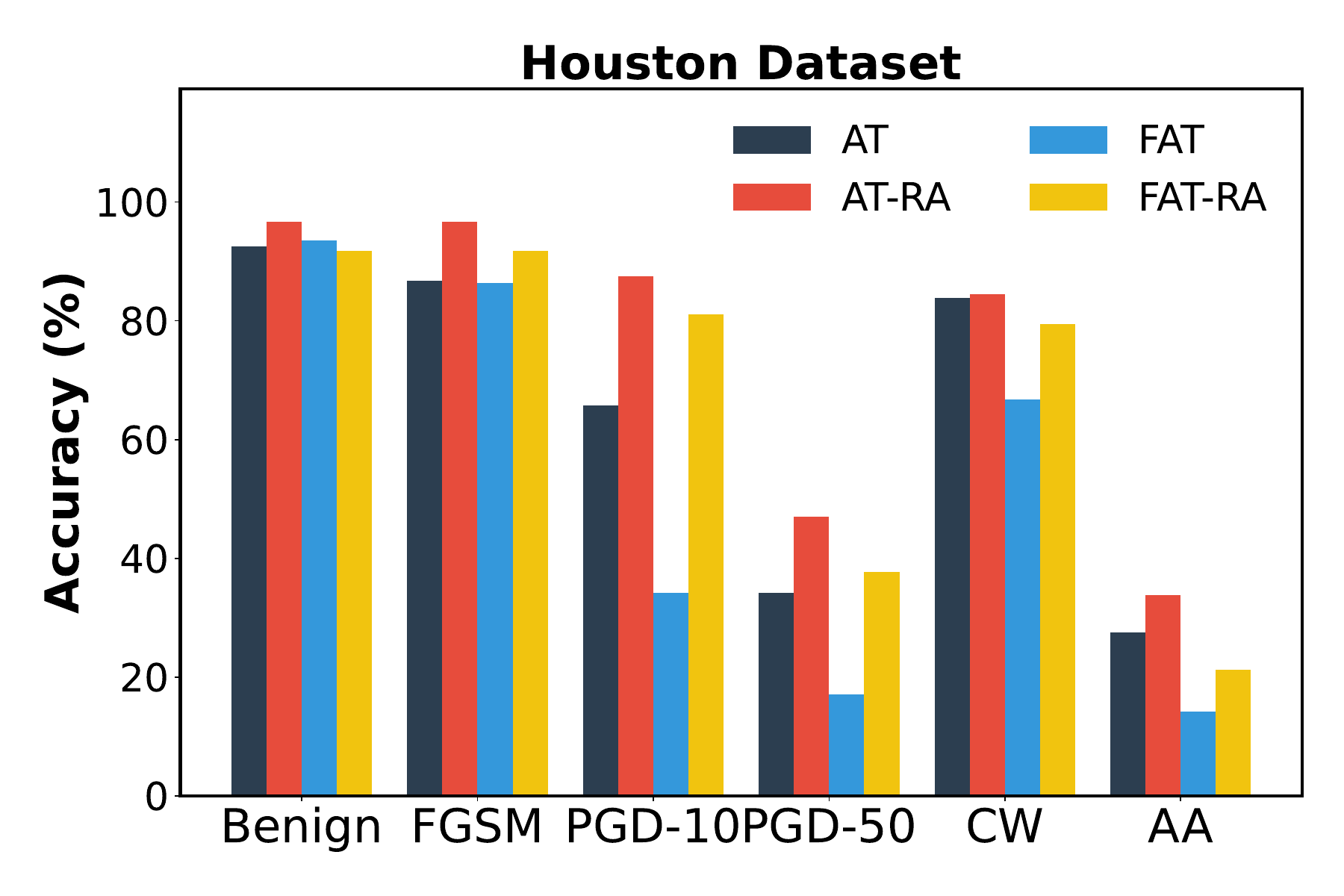}}
    \subfloat[]{\includegraphics[width=0.5\linewidth]{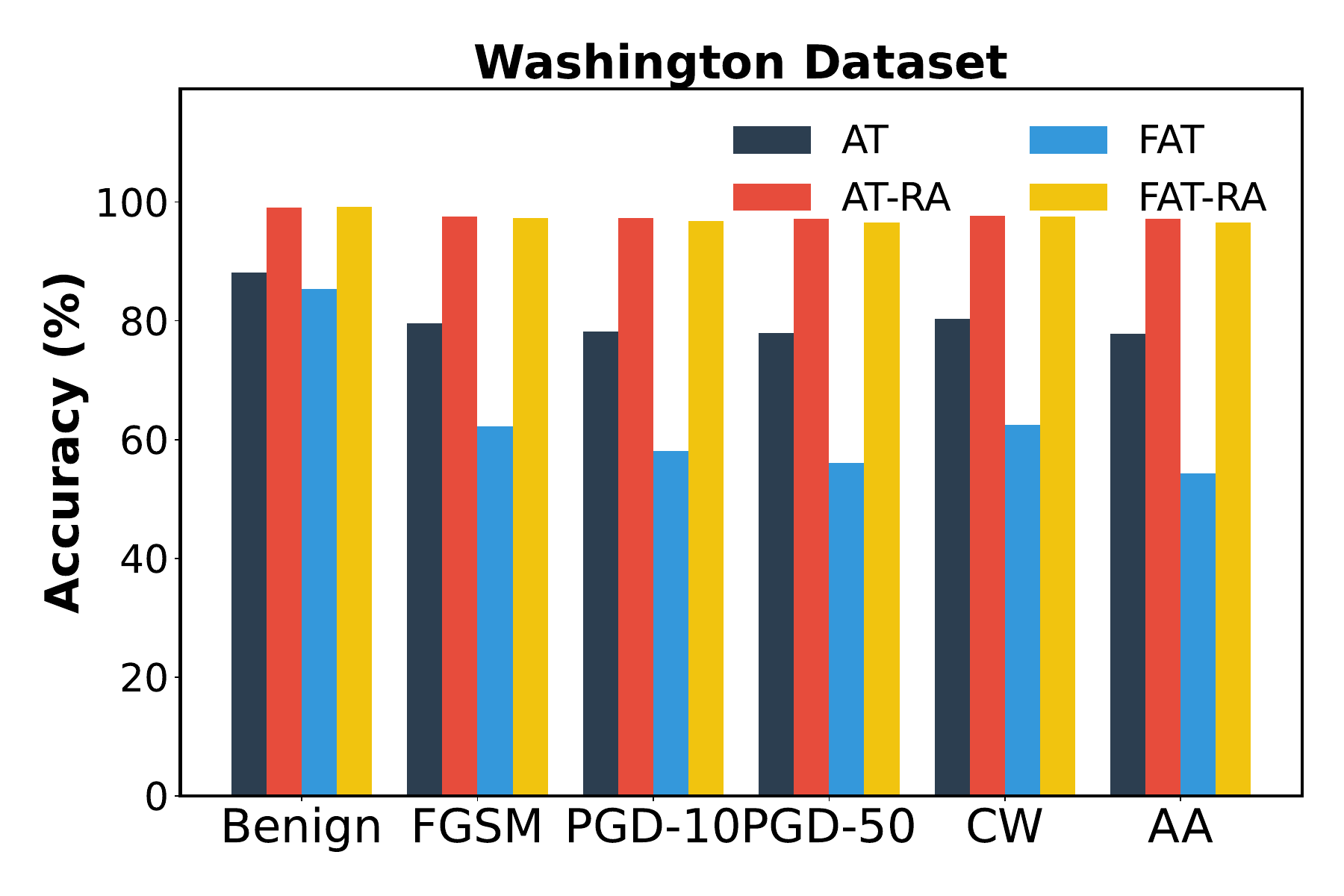}}
    \caption{Performance comparison of AT-RA and FAT-RA with their respective baselines on four benchmark datasets under various adversarial attack settings. Results demonstrate that RA consistently enhances both benign accuracy and adversarial robustness.}
    \label{fig:ra_results}
\end{figure}

\subsection{Applicability of AT-HARL and AT-RA}
It is important to clarify that AT-HARL and AT-RA are designed as two independently deployable AT strategies rather than components to be jointly applied within a single training framework. AT-HARL operates at the loss-function level by explicitly incorporating hyperspectral domain priors to rebalance class-wise optimization dynamics, and is particularly suitable for standard AT settings where the misclassification concentration phenomenon arises from biased decision boundaries and spectral semantic distortion, as evidenced by dominant-class error aggregation and class-wise robustness degradation. In contrast, AT-RA functions at the data level by introducing controlled spectral--spatial variability through RandAugment, making it more effective in scenarios where the misclassification concentration phenomenon is driven by insufficient sample diversity or limited effective training distributions under adversarial perturbations, such as small training sets or highly homogeneous spectral classes.

Due to their fundamentally different operating mechanisms, directly combining AT-HARL and AT-RA may introduce competing effects between loss reweighting and data-level augmentation, potentially destabilizing optimization and obscuring the individual contributions of each strategy. Therefore, we intentionally treat AT-HARL and AT-RA as two complementary yet alternative solutions, allowing practitioners to select the appropriate method according to specific data characteristics and training conditions.

\section{Experiments and Discussions}\label{experiments}

\subsection{Settings}

\paragraph{Datasets}We conduct experiments on four widely used HSI datasets, 
including Pavia University, Houston, Washington DC, and Salinas Valley. 
These datasets cover diverse scenes, spatial resolutions, and class distributions, providing a comprehensive evaluation benchmark for adversarial robustness in HSI classification. 

\paragraph{Metrics} In addition to the benign accuracy, we also evaluate the model's adversarial robustness under an \(\epsilon = 8/255\) bounded perturbation in the \(l_\infty\) norm. The attacks used include the single-step FGSM \cite{23} and multi-step iterative attacks such as PGD-10 \cite{24}, PGD-50 \cite{24}, and CW \cite{43}, all with a step size of 2/255. Furthermore, we also utilize AutoAttack (AA) \cite{44}, which is considered one of the strongest attacks. It should be noted that due to the limitation of GPU memory, the AA on the Salinas dataset cannot be conducted.

\paragraph{Details} We use the Stochastic Gradient Descent (SGD) optimizer to train the model, with an initial learning rate of 0.1, momentum of 0.9, and weight decay of 5e-4. The batch size is set to 128. The training is conducted for 100 epochs, with the learning rate reduced to one-tenth of the original value at the 90th and 95th epochs. For adversarial example generation, we set up PGD attacks to generate adversarial examples with a maximum perturbation of 8/255 and a step size of 2/255, using 5 iterations for internal maximization. Additionally, we also conduct tests on FAT, where adversarial examples are generated using FGSM attacks with a maximum perturbation of 8/255 and a step size of 8/255. In the hyperparameter selection process, we set $\beta$ to 0.005 and $\gamma$ to 1e-5.

\begin{table}[htbp]
\caption{Benign accuracy and adversarial robustness of AT-HARL on four datasets using ResNet-18~\cite{36} when Hyperspectral-Aware Rare-class Loss is applied to adversarial training. The best results are highlighted in bold.}
\label{tab11}
\centering
\setlength{\tabcolsep}{2.3pt}
\begin{tabular}{c|c|c c|c c c c}
\toprule[2pt]
\multirow{2}{*}{Dataset} & \multirow{2}{*}{Method} & \multicolumn{2}{c|}{Components} & \multicolumn{4}{c}{Accuracy(\%)} \\
\cmidrule(r){3-4} \cmidrule(r){5-8}
& & SABCL & HARL & Benign & PGD-50 & CW & AA \\
\toprule[1pt]
\multirow{4}{*}{Pavia}
& AT        &           &           & 97.09 & 75.33 & 90.53 & 72.85 \\
& AT-SABCL  & \checkmark&           & 97.13 & 80.21 & 93.03 & 75.48 \\
& AT-RCSEL  &           & \checkmark& 96.67 & 80.11 & 91.54 & 76.52 \\
& AT-HARL   & \checkmark& \checkmark& \textbf{97.94} & \textbf{81.92} & \textbf{94.86} & \textbf{76.91} \\
\toprule[1pt]
\multirow{4}{*}{Salinas}
& AT        &           &           & 87.19 & 86.40 & 86.73 & -- \\
& AT-SABCL  & \checkmark&           & 87.18 & 86.47 & 86.82 & -- \\
& AT-RCSEL  &           & \checkmark& 87.98 & 86.57 & 86.89 & -- \\
& AT-HARL   & \checkmark& \checkmark& \textbf{89.35} & \textbf{87.73} & \textbf{88.05} & -- \\
\toprule[1pt]
\multirow{4}{*}{Houston}
& AT        &           &           & 92.50 & 34.20 & 83.85 & 27.52 \\
& AT-SABCL  & \checkmark&           & 93.98 & 43.98 & 84.21 & 30.44 \\
& AT-RCSEL  &           & \checkmark& 93.57 & 50.99 & 86.83 & 29.21 \\
& AT-HARL   & \checkmark& \checkmark& \textbf{94.90} & \textbf{52.01} & \textbf{87.03} & \textbf{31.56} \\
\toprule[1pt]
\multirow{4}{*}{Washington}
& AT        &           &           & 88.14 & 77.92 & 80.32 & 77.77 \\
& AT-SABCL  & \checkmark&           & 97.13 & 89.55 & 91.65 & 82.74 \\
& AT-RCSEL  &           & \checkmark& 97.71 & 89.26 & 91.63 & 81.96 \\
& AT-HARL   & \checkmark& \checkmark& \textbf{97.98} & \textbf{89.91} & \textbf{92.21} & \textbf{89.52} \\
\toprule[2pt]
\end{tabular}
\end{table}

\subsection{Main Results}
\paragraph{Effectiveness of Each Component in AT-HARL}

Table~\ref{tab11} presents the benign accuracy and adversarial robustness of AT-HARL across four datasets using ResNet-18. Compared with the baseline AT, introducing either SABCL or RCSEL individually leads to consistent improvements in both benign accuracy and robustness under strong multi-step attacks. For example, AT-SABCL improves the PGD-50 accuracy on the Pavia dataset from 75.33\% to 80.21\%, while AT-RCSEL enhances PGD-50 robustness on the Houston dataset from 34.20\% to 50.99\%. When both components are jointly incorporated, AT-HARL achieves the best overall performance across all datasets and attack settings. In particular, AT-HARL increases the benign accuracy on Washington from 88.14\% to 97.98\% and further improves PGD-50 robustness on Salinas from 86.40\% to 87.73\%. These results indicate that SABCL and RCSEL are complementary, and their combination yields additional gains over either component alone.

\begin{table}[htbp]
\caption{Benign accuracy and adversarial robustness of FAT-HARL on four datasets using ResNet-18~\cite{36} when Hyperspectral-Aware Rare-class Loss is applied to fast adversarial training. The best results are highlighted in bold.}
\label{tab12}
\centering
\setlength{\tabcolsep}{2.3pt}
\begin{tabular}{c|c|c c|c c c c}
\toprule[2pt]
\multirow{2}{*}{Dataset} & \multirow{2}{*}{Method} & \multicolumn{2}{c|}{Components} & \multicolumn{4}{c}{Accuracy(\%)} \\
\cmidrule(r){3-4} \cmidrule(r){5-8}
& & SABCL & HARL & Benign & PGD-50 & CW & AA \\
\toprule[1pt]
\multirow{4}{*}{Pavia}
& FAT        &           &           & 93.36 & 60.88 & 81.38 & 58.12 \\
& FAT-SABCL  & \checkmark&           & 93.05 & 62.79 & 82.36 & 58.44 \\
& FAT-RCSEL  &           & \checkmark& 93.51 & 61.24 & 82.90 & 59.41 \\
& FAT-HARL   & \checkmark& \checkmark& \textbf{93.94} & \textbf{63.79} & \textbf{83.16} & \textbf{60.30} \\
\toprule[1pt]
\multirow{4}{*}{Salinas}
& FAT        &           &           & 86.60 & 80.20 & 85.75 & -- \\
& FAT-SABCL  & \checkmark&           & 87.57 & 84.58 & 85.83 & -- \\
& FAT-RCSEL  &           & \checkmark& 87.41 & 81.30 & 86.44 & -- \\
& FAT-HARL   & \checkmark& \checkmark& \textbf{88.95} & \textbf{85.05} & \textbf{86.74} & -- \\
\toprule[1pt]
\multirow{4}{*}{Houston}
& FAT        &           &           & 93.53 & 17.11 & 66.71 & 14.22 \\
& FAT-SABCL  & \checkmark&           & 94.33 & 20.82 & 68.63 & 18.70 \\
& FAT-RCSEL  &           & \checkmark& 94.45 & 20.13 & 70.87 & 18.35 \\
& FAT-HARL   & \checkmark& \checkmark& \textbf{95.72} & \textbf{27.12} & \textbf{72.18} & \textbf{19.41} \\
\toprule[1pt]
\multirow{4}{*}{Washington}
& FAT        &           &           & 85.29 & 56.04 & 62.46 & 54.28 \\
& FAT-SABCL  & \checkmark&           & 97.47 & 88.33 & 91.84 & 87.60 \\
& FAT-RCSEL  &           & \checkmark& 97.60 & 87.74 & 91.88 & 87.44 \\
& FAT-HARL   & \checkmark& \checkmark& \textbf{97.98} & \textbf{88.86} & \textbf{93.60} & \textbf{88.04} \\
\toprule[2pt]
\end{tabular}
\end{table}

In addition to standard AT, we further evaluate the contribution of each component under the FAT setting, as summarized in Table~\ref{tab12}. Compared with the baseline FAT, incorporating either SABCL or RCSEL consistently improves both benign accuracy and robustness against iterative attacks. For instance, FAT-SABCL improves PGD-50 robustness on the Pavia dataset from 60.88\% to 62.79\%, while FAT-RCSEL enhances CW robustness on the Houston dataset from 66.71\% to 70.87\%. When both components are integrated, FAT-HARL achieves the strongest performance across all datasets. Notably, FAT-HARL boosts PGD-50 accuracy on Houston from 17.11\% to 27.12\% and raises the benign accuracy on Washington from 85.29\% to 97.98\%. These observations further confirm the effectiveness and compatibility of SABCL and RCSEL under FAT. For completeness, the full results under additional attack settings are provided in \textbf{Tables~IV and~V in the Appendix}.

\paragraph{Robustness Enhancement via RandAugment}

The comparison results are illustrated in Fig.~\ref{fig:ra_results}. It can be observed that, across all four datasets, the introduction of RandAugment consistently enhances both benign accuracy and adversarial robustness for AT and FAT. In particular, AT-RA achieves substantial gains against strong iterative attacks such as PGD-50 and AA, while also improving benign accuracy, as evidenced on the Washington dataset. Similarly, FAT-RA yields notable robustness improvements under challenging attack settings, such as raising PGD-10 robustness on the Houston dataset by a large margin. Overall, the bar plots clearly demonstrate that RandAugment serves as an effective regularization strategy, reinforcing model generalization and defense capability under diverse adversarial scenarios. In addition, we employ undirected chord diagrams to qualitatively analyze the misclassification patterns of AT, AT-HARL, and AT-RA, with detailed visualizations provided in \textbf{Fig.~2 of the Appendix}.

\begin{table}[htbp]
\caption{Benign accuracy and adversarial robustness of six state-of-the-art robust deep neural networks on four datasets when AT-HARL is applied. The best results are highlighted in bold.}
\label{tab14}
\centering
\setlength{\tabcolsep}{1pt}
\begin{tabular}{c|c|cc|cc|cc}
\toprule[2pt]
\multirow{2}{*}{Dataset} & \multirow{2}{*}{Model} 
& \multicolumn{2}{c|}{Benign} 
& \multicolumn{2}{c|}{PGD-50} 
& \multicolumn{2}{c}{CW} \\
\cmidrule(r){3-4} \cmidrule(r){5-6} \cmidrule(r){7-8}
& & AT & AT-HARL & AT & AT-HARL & AT & AT-HARL \\
\toprule[1pt]

\multirow{6}{*}{Pavia}
& DilatedFCN      & 95.92 & \textbf{96.05} & 69.59 & \textbf{72.08} & 73.17 & \textbf{78.64} \\
& SSFCN           & 94.57 & \textbf{96.19} & 23.38 & \textbf{45.04} & 35.93 & \textbf{56.26} \\
& SpaFCN          & 94.96 & \textbf{96.20} & 52.94 & \textbf{60.42} & 64.76 & \textbf{66.20} \\
& SACNet          & \textbf{97.14} & 94.79 & 71.48 & \textbf{90.09} & 79.62 & \textbf{89.33} \\
& RCCA            & 95.21 & \textbf{96.08} & 73.98 & \textbf{88.45} & 88.32 & \textbf{92.91} \\
& S\textsuperscript{3}ANet & 97.49 & \textbf{99.01} & 86.37 & \textbf{95.36} & 89.12 & \textbf{97.23} \\
\toprule[1pt]

\multirow{6}{*}{Salinas}
& DilatedFCN      & 96.20 & \textbf{96.52} & 50.72 & \textbf{51.98} & 57.00 & \textbf{69.47} \\
& SSFCN           & 94.62 & \textbf{97.41} & 15.14 & \textbf{26.82} & 31.80 & \textbf{55.03} \\
& SpaFCN          & \textbf{97.54} & 96.46 & 69.29 & \textbf{71.97} & 71.50 & \textbf{74.54} \\
& SACNet          & \textbf{96.91} & 96.11 & 69.82 & \textbf{80.59} & 77.23 & \textbf{86.27} \\
& RCCA            & \textbf{99.52} & 99.10 & 65.01 & \textbf{68.07} & 68.67 & \textbf{73.04} \\
& S\textsuperscript{3}ANet & \textbf{99.92} & 99.71 & 89.16 & \textbf{95.68} & 95.88 & \textbf{98.83} \\
\toprule[1pt]

\multirow{6}{*}{Houston}
& DilatedFCN      & 94.06 & \textbf{95.14} & 47.98 & \textbf{53.82} & 58.65 & \textbf{61.32} \\
& SSFCN           & \textbf{97.30} & 96.12 & 31.68 & \textbf{36.55} & 36.33 & \textbf{43.22} \\
& SpaFCN          & \textbf{97.12} & 94.40 & 34.92 & \textbf{43.34} & 38.45 & \textbf{47.77} \\
& SACNet          & 96.65 & \textbf{97.08} & 82.49 & \textbf{87.29} & 89.88 & \textbf{92.42} \\
& RCCA            & 98.32 & \textbf{98.56} & 65.42 & \textbf{71.55} & 75.18 & \textbf{84.08} \\
& S\textsuperscript{3}ANet & \textbf{99.51} & 98.73 & 79.43 & \textbf{85.50} & 86.75 & \textbf{90.27} \\
\toprule[1pt]

\multirow{6}{*}{Washington}
& DilatedFCN      & 96.18 & \textbf{96.26} & 74.45 & \textbf{85.03} & 78.63 & \textbf{82.83} \\
& SSFCN           & 99.19 & \textbf{99.38} & 75.11 & \textbf{86.67} & 86.78 & \textbf{92.01} \\
& SpaFCN          & 97.52 & \textbf{98.72} & 79.67 & \textbf{86.49} & 83.47 & \textbf{89.36} \\
& SACNet          & 98.15 & \textbf{98.70} & 87.10 & \textbf{93.66} & 89.94 & \textbf{95.21} \\
& RCCA            & 95.39 & \textbf{96.98} & 41.31 & \textbf{49.84} & 49.80 & \textbf{58.09} \\
& S\textsuperscript{3}ANet & 98.48 & \textbf{98.45} & 87.60 & \textbf{94.80} & 96.88 & \textbf{97.86} \\
\toprule[2pt]
\end{tabular}
\end{table}

\subsection{Comparisons With the State-of-the-Art Methods}

\paragraph{Effectiveness of AT-HARL on SOTA Models}

To further validate the effectiveness and generality of the proposed framework, we apply AT-HARL to six state-of-the-art HSI classification networks, including DilatedFCN~\cite{DilatedFCN}, SSFCN~\cite{SSFCN}, SpaFCN~\cite{SSFCN}, SACNet~\cite{20}, RCCA~\cite{27}, and S\textsuperscript{3}ANet~\cite{29}. Our loss formulation is seamlessly integrated into standard AT. The main results under representative strong attacks are reported in Table~\ref{tab14}, while the complete evaluation under additional attacks is provided in \textbf{Tables~VI of the Appendix}.

As shown in Table~\ref{tab14}, under the standard AT setting, AT-HARL consistently improves the performance of all six baseline models across the four datasets. The gains are observed not only in benign accuracy but are particularly pronounced in adversarial robustness under strong iterative attacks. For example, on the Pavia dataset, integrating AT-HARL into SSFCN increases its PGD-50 accuracy from 23.38\% to 45.04\%, and its CW accuracy from 35.93\% to 56.26\%. Similarly, on the Washington dataset, SACNet benefits substantially from AT-HARL, with its PGD-50 robustness improving from 87.10\% to 93.66\%. These results demonstrate that the proposed HARL framework serves as an effective and general enhancement strategy for improving adversarial robustness across diverse HSI classification models.

\paragraph{Effectiveness of RandAugment on SOTA Models}
To further analyze the contribution of RandAugment, we investigate its impact on six state-of-the-art HSI classification models under AT. The results of integrating RandAugment into standard AT are summarized in Table~\ref{tab16}. As can be observed, AT-RA consistently enhances the adversarial robustness of all evaluated models across different datasets. For example, on the Pavia dataset, applying AT-RA to RCCA significantly improves its PGD-50 accuracy from 73.98\% to 97.40\%, demonstrating the effectiveness of RandAugment in strengthening robustness under strong iterative attacks. 

\begin{table}[htbp]
\caption{Benign accuracy and adversarial robustness of six state-of-the-art robust deep neural networks on four datasets when AT-RA is applied. The best results are highlighted in bold.}
\label{tab16}
\centering
\setlength{\tabcolsep}{2pt}
\begin{tabular}{c|c|cc|cc|cc}
\toprule[2pt]
\multirow{2}{*}{Dataset} & \multirow{2}{*}{Model} &
\multicolumn{2}{c|}{Benign} &
\multicolumn{2}{c|}{PGD-50} &
\multicolumn{2}{c}{CW} \\
\cmidrule(r){3-4} \cmidrule(r){5-6} \cmidrule(r){7-8}
 & & AT & AT-RA & AT & AT-RA & AT & AT-RA \\
\toprule[1pt]

\multirow{6}{*}{Pavia}
& DilatedFCN & 95.92 & \textbf{96.52} & 69.59 & \textbf{72.79} & 73.17 & \textbf{86.51} \\
& SSFCN & 94.57 & \textbf{98.62} & 23.38 & \textbf{41.79} & 35.93 & \textbf{43.36} \\
& SpaFCN & 94.96 & \textbf{98.33} & 52.94 & \textbf{57.40} & 64.76 & \textbf{67.86} \\
& SACNet & 97.14 & \textbf{98.12} & 71.48 & \textbf{91.79} & 79.62 & \textbf{93.36} \\
& RCCA & 95.21 & \textbf{98.33} & 73.98 & \textbf{97.40} & 88.32 & \textbf{97.86} \\
& S\textsuperscript{3}ANet & 97.49 & \textbf{97.87} & 86.37 & \textbf{96.11} & 89.12 & \textbf{96.73} \\

\toprule[1pt]
\multirow{6}{*}{Salinas}
& DilatedFCN & 96.20 & \textbf{98.51} & 50.72 & \textbf{72.80} & 57.00 & \textbf{88.30} \\
& SSFCN & 94.62 & \textbf{99.20} & 15.14 & \textbf{31.78} & 31.80 & \textbf{43.42} \\
& SpaFCN & 97.54 & \textbf{99.50} & 69.29 & \textbf{77.35} & 71.50 & \textbf{79.65} \\
& SACNet & 96.91 & \textbf{99.46} & 69.82 & \textbf{72.31} & 77.23 & \textbf{86.62} \\
& RCCA & 99.52 & \textbf{99.68} & 65.01 & \textbf{91.69} & 68.67 & \textbf{97.95} \\
& S\textsuperscript{3}ANet & 99.92 & 99.89 & 89.16 & \textbf{91.73} & 95.88 & \textbf{96.20} \\

\toprule[1pt]
\multirow{6}{*}{Houston}
& DilatedFCN & 94.06 & \textbf{95.01} & 47.98 & \textbf{58.46} & 58.65 & \textbf{63.34} \\
& SSFCN & 97.30 & \textbf{97.37} & 31.68 & \textbf{34.92} & 36.33 & \textbf{49.25} \\
& SpaFCN & 97.12 & 96.95 & 34.92 & \textbf{46.75} & 38.45 & \textbf{59.19} \\
& SACNet & 96.65 & \textbf{97.18} & 82.49 & \textbf{88.12} & 89.88 & \textbf{94.78} \\
& RCCA & 98.32 & \textbf{98.74} & 65.42 & \textbf{88.05} & 75.18 & \textbf{94.56} \\
& S\textsuperscript{3}ANet & 99.51 & \textbf{99.53} & 79.43 & \textbf{88.56} & 86.75 & \textbf{92.39} \\

\toprule[1pt]
\multirow{6}{*}{Washington}
& DilatedFCN & 96.18 & \textbf{96.98} & 74.45 & \textbf{85.87} & 78.63 & \textbf{88.32} \\
& SSFCN & 99.19 & 98.52 & 75.11 & \textbf{87.48} & 86.78 & \textbf{89.36} \\
& SpaFCN & 97.52 & \textbf{98.99} & 79.67 & \textbf{85.52} & 83.47 & \textbf{86.26} \\
& SACNet & 98.15 & 97.61 & 87.10 & \textbf{95.09} & 89.94 & \textbf{97.06} \\
& RCCA & 95.39 & \textbf{97.71} & 41.31 & \textbf{77.86} & 49.80 & \textbf{87.71} \\
& S\textsuperscript{3}ANet & 98.48 & \textbf{98.56} & 87.60 & \textbf{92.33} & 96.88 & \textbf{97.75} \\

\toprule[2pt]
\end{tabular}
\end{table}

\paragraph{Visual Analysis of the Proposed Methods}

\begin{figure}[htbp]
    \centering
    \includegraphics[width=1\linewidth]{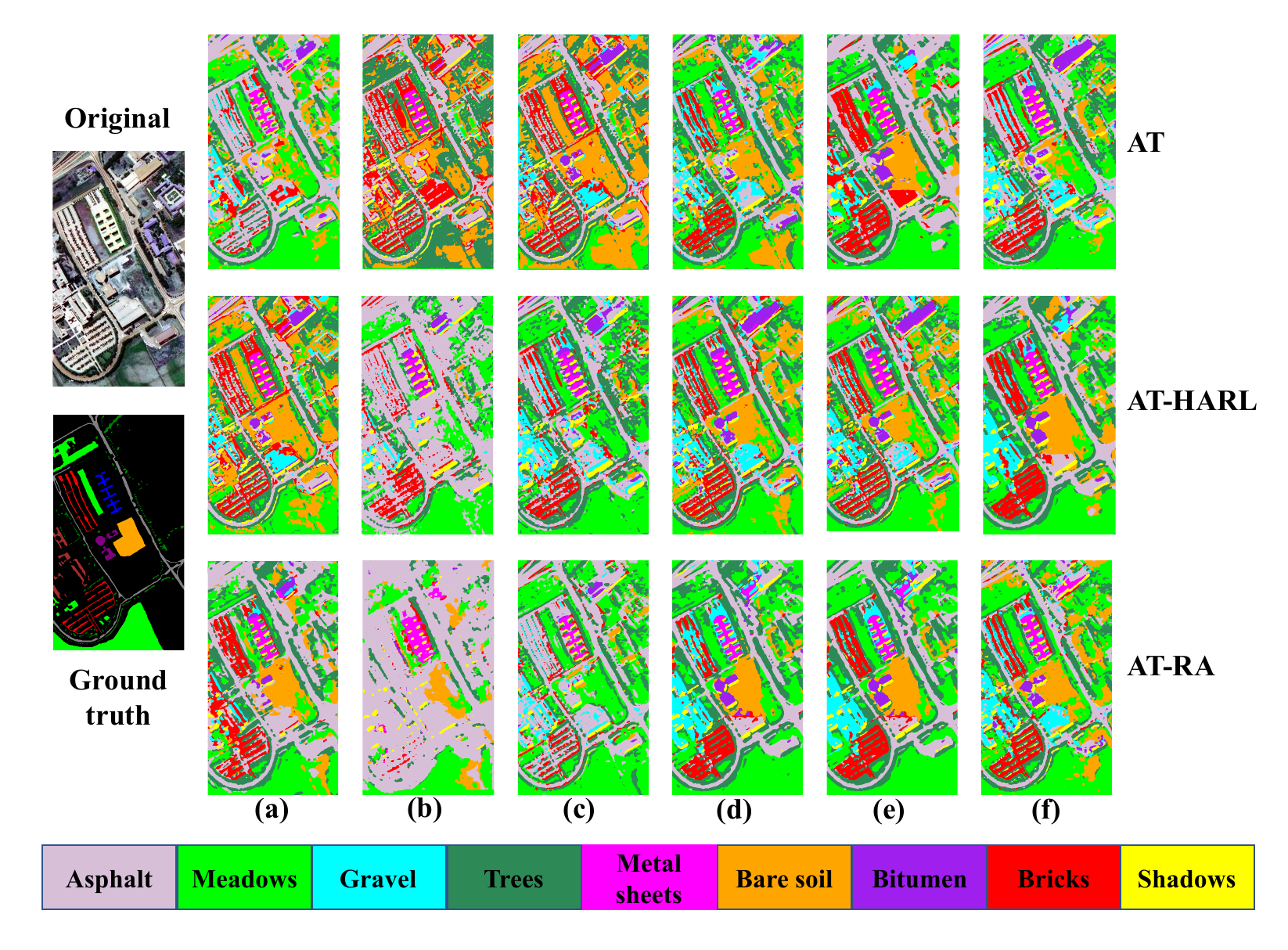}
    \caption{Classification maps for the PaviaU dataset on the adversarial test set. (a) DilatedFCN. (b) SSFCN. (c) SpaFCN. (d) SACNet. (e) RCCA. (f) S\textsuperscript{3}ANet. Compared with standard AT, both AT-HARL and AT-RA effectively alleviate the \textbf{misclassification concentration phenomenon}, particularly reducing the collapse of multiple classes into the \textit{Meadows} class, thereby yielding more balanced classification maps and improved robustness.}
    \label{fig:classification_map_paviaU}
\end{figure}

\begin{figure}[htbp]
    \centering
    \includegraphics[width=1\linewidth]{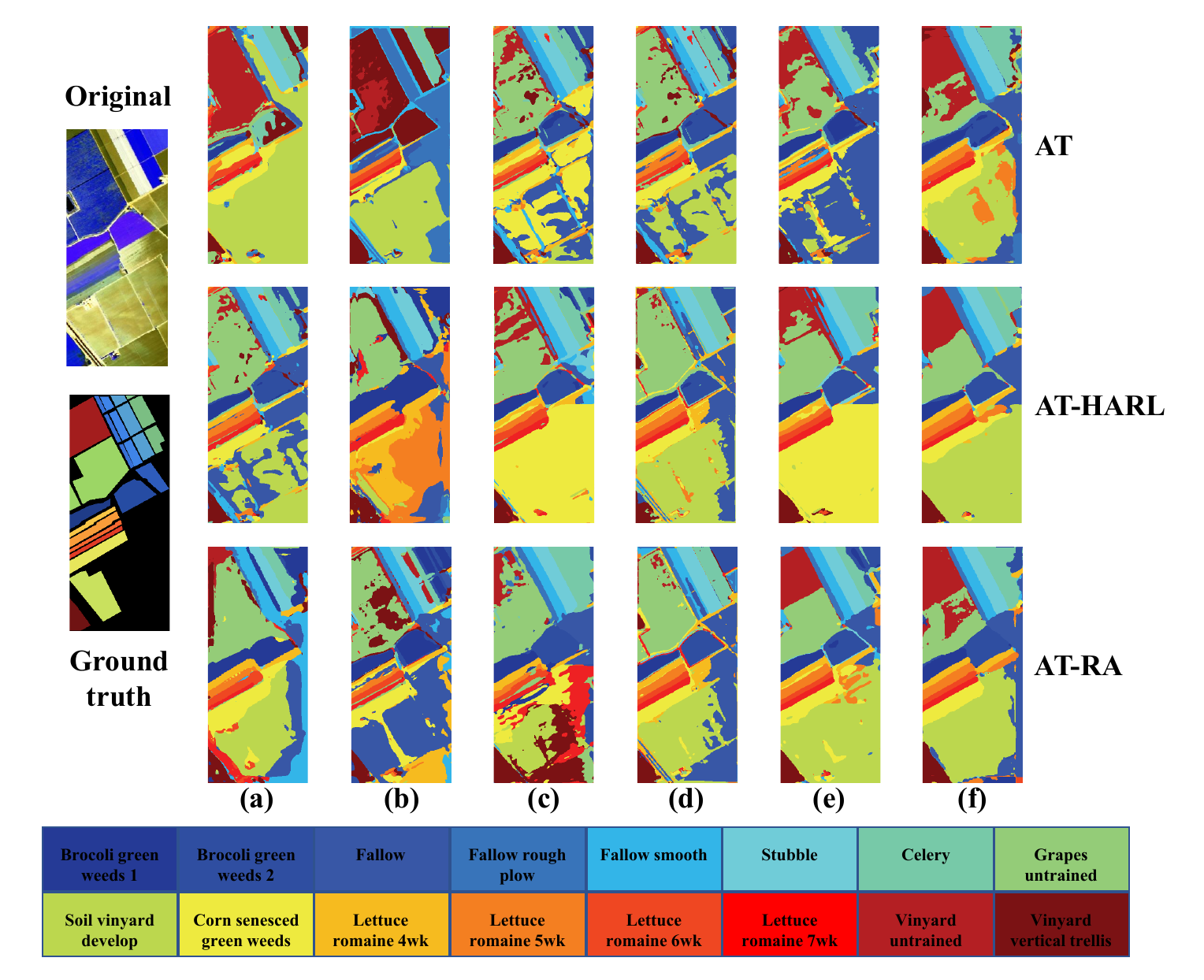}
    \caption{Classification maps for the Salinas dataset on the adversarial test set. (a) DilatedFCN. (b) SSFCN. (c) SpaFCN. (d) SACNet. (e) RCCA. (f) S\textsuperscript{3}ANet. Similar to the PaviaU results in Fig.~\ref{fig:classification_map_paviaU}, AT-HARL and AT-RA significantly reduce the excessive misclassification toward the dominant \textit{Grapes untrained} class, demonstrating their effectiveness in mitigating concentrated misclassification and enhancing adversarial robustness.}
    \label{fig:classification_map_salinas}
\end{figure}

To further validate the effectiveness of our proposed strategies, we visualize the classification maps obtained under different adversarial training schemes. 
Fig.~\ref{fig:classification_map_paviaU} and Fig.~\ref{fig:classification_map_salinas} illustrate the results on the PaviaU and Salinas datasets, respectively. 
It can be observed that compared with standard AT, AT-HARL and AT-RA produce more spatially coherent and balanced classification maps, indicating that both methods effectively alleviate concentrated misclassification and restore class separability under adversarial perturbations.

\begin{table*}[htbp]
\caption{Benign accuracy and adversarial robustness of AT-HARL on four datasets under different hyperparameter settings. The best of these results have been bolded.}
\label{tab18}
\centering
\begin{tabular}{c| c| c c c c c| c c c c c}
\toprule[2pt]
\multirow{2}{*}{Dataset}    & \multirow{2}{*}{Attack Method} & \multicolumn{5}{c|}{value of $\beta$ ($\gamma$=1)}                                                                                                                                            & \multicolumn{5}{c}{value of $\gamma$ ($\beta$=0.005)}                                                                                                                                                     \\ \cline{3-12} 
                            &                                & 0.001                           & 0.003                           & 0.005                           & 0.007                           & 0.009                           & 1                                     & 3                            & 5                            & 7                           & 9                            \\ \toprule[1pt]
\multirow{6}{*}{Pavia}      & Benign                         & 90.71                           & 94.39                           & \textbf{97.94}                  & 93.62                           & 89.09                           & \textbf{97.94}                           & 96.11                           & 94.42                           & 93.12                           & 90.89                           \\
                            & FGSM                           & 85.00                           & 89.73                           & \textbf{94.40}                  & 90.57                           & 87.23                           & \textbf{94.40}                           & 92.06                           & 90.92                           & 87.50                           & 85.45                           \\
                            & PGD-10                         & 81.36                           & 84.49                           & \textbf{87.23}                  & 84.27                           & 81.67                           & \textbf{87.23}                           & 86.12                           & 84.73                           & 83.47                           & 82.27                           \\
                            & PGD-50                         & 78.22                           & 80.13                           & \textbf{81.92}                  & 80.04                           & 77.66                           & \textbf{81.92}                           & 81.43                           & 80.61                           & 80.16                           & 79.43                           \\
                            & CW                             & 84.75                           & 90.34                           & \textbf{94.86}                  & 88.18                           & 83.35                           & \textbf{94.86}                           & 82.76                           & 80.04                           & 87.30                           & 84.99                           \\
                            & AA                             & 75.45                           & 76.26                           & \textbf{76.91}                  & 75.16                           & 73.41                           & \textbf{76.91}                           & 76.69                           & 76.24                           & 75.98                           & 75.55                           \\ \toprule[1pt]
\multirow{6}{*}{Salinas}    & Benign                         & 86.02                           & 87.80                           & \textbf{89.35}                  & 88.43                           & 87.60                           & \textbf{89.35}                           & 88.97                           & 88.62                           & 883.60                          & 88.01                           \\
                            & FGSM                           & 85.29                           & 87.27                           & \textbf{88.93}                  & 88.19                           & 87.43                           & \textbf{88.93}                           & 88.71                           & 88.53                           & 88.20                           & 88.00                           \\
                            & PGD-10                         & 84.99                           & 86.76                           & \textbf{88.80}                  & 87.73                           & 86.74                           & \textbf{88.80}                           & 88.27                           & 87.69                           & 87.32                           & 86.90                           \\
                            & PGD-50                         & 84.93                           & 86.39                           & \textbf{87.73}                  & 87.52                           & 86.65                           & \textbf{87.73}                           & 87.50                           & 87.03                           & 86.86                           & 86.79                           \\
                            & CW                             & 85.26                           & \textbf{88.18}                  & 88.05                           & 87.32                           & 86.96                           & \textbf{88.05}                           & 87.86                           & 87.64                           & 87.26                           & 87.11                           \\
                            & AA                             & \diagbox{}{}
 & \diagbox{}{} & \diagbox{}{} & \diagbox{}{} & \diagbox{}{} & \diagbox{}{} & \diagbox{}{} & \diagbox{}{} & \diagbox{}{} & \diagbox{}{} \\ \toprule[1pt]
\multirow{6}{*}{Houston}    & Benign                         & 93.57                           & \textbf{95.21}                  & 94.90                           & 94.31                           & 93.65                           & \textbf{94.90}                           & 94.73                           & 94.69                           & 94.46                           & 94.18                           \\
                            & FGSM                           & 91.58                           & 92.14                           & \textbf{92.89}                  & 88.82                           & 86.69                           & \textbf{92.89}                           & 92.68                           & 92.47                           & 92.22                           & 91.96                           \\
                            & PGD-10                         & 58.81                           & 63.15                           & \textbf{69.77}                  & 55.85                           & 44.25                           & \textbf{69.77}                           & 66.59                           & 61.73                           & 54.44                           & 53.21                           \\
                            & PGD-50                         & 50.99                           & 51.50                           & \textbf{52.01}                  & 46.47                           & 41.44                           & \textbf{52.01}                           & 46.73                           & 39.69                           & 33.63                           & 29.15                           \\
                            & CW                             & 74.83                           & 81.38                           & \textbf{87.03}                  & 79.54                           & 72.24                           & \textbf{87.03}                           & 83.50                           & 77.73                           & 71.83                           & 67.31                           \\
                            & AA                             & 27.81                           & 29.63                           & \textbf{31.56}                  & 27.21                           & 21.35                           & \textbf{31.56}                           & 29.22                           & 26.22                           & 25.09                           & 22.18                           \\ \toprule[1pt]
\multirow{6}{*}{Washington} & Benign                         & 96.92                           & 97.39                           & 97.98                           & 97.03                           & \textbf{98.24}                  & 97.98                                    & 98.01                           & 98.11                           & 97.92                           & \textbf{98.18}                  \\
                            & FGSM                           & 90.33                           & 90.56                           & \textbf{91.28}                  & 90.80                           & 90.65                           & \textbf{91.28}                           & 88.62                           & 88.71                           & 84.20                           & 82.43                           \\
                            & PGD-10                         & 89.13                           & 89.67                           & \textbf{90.23}                  & 89.35                           & 88.70                           & \textbf{90.23}                           & 87.50                           & 86.58                           & 83.74                           & 81.40                           \\
                            & PGD-50                         & 88.73                           & 89.65                           & \textbf{89.91}                  & 87.74                           & 85.51                           & 89.91                                    & \textbf{90.23}                  & 86.98                           & 85.15                           & 81.20                           \\
                            & CW                             & 81.13                           & 87.00                           & \textbf{92.21}                  & 91.45                           & 90.56                           & \textbf{92.21}                           & 90.98                           & 85.36                           & 83.43                           & 83.23                           \\
                            & AA                             & 87.17                           & 89.33                           & \textbf{89.52}                  & 87.88                           & 86.34                           & \textbf{89.52}                           & 86.74                           & 85.42                           & 83.59                           & 81.19 \\                          
\toprule[2pt]
\end{tabular}
\end{table*}

\subsection{Hyperparameter Sensitivity Analysis}
The performance of the AT-HARL framework is contingent on the proper tuning of two key hyperparameters: $\beta$ and $\gamma$, which modulate the contributions of the SABCL and RCSEL components, respectively. We conducted a detailed sensitivity analysis to evaluate the impact of these parameters on model performance across all four datasets. As shown in Table~\ref{tab18}, the framework's performance exhibits a clear dependency on the value of $\beta$. Fixing $\gamma$ at $1$, we observed that both benign accuracy and adversarial robustness consistently improved as $\beta$ was increased from 0.001 to 0.005. Performance peaked at $\beta=0.005$ before declining with further increases to 0.007 and 0.009. This trend suggests an optimal balance point, where the SABCL component effectively enhances adversarial robustness without disrupting the overall training objective.

Similarly, we analyzed the effect of varying $\gamma$ while holding $\beta$ constant at 0.005. The results indicate that the framework is highly sensitive to this parameter, with the best performance consistently achieved at the smallest tested value, $\gamma=1\mathrm{e}{-5}$. As $\gamma$ was increased, a gradual degradation in both benign and adversarial accuracy was observed across all datasets. This is likely due to an excessively large weight on the RCSEL component, which may cause the model to overfit to the spectral features of rare classes, consequently compromising its generalization and robustness to a broader range of adversarial perturbations. These findings underscore the importance of judicious hyperparameter selection and provide a guideline for configuring the AT-HARL framework for optimal performance.

\subsection{Ablation Study}

\begin{figure}[htbp]
    \centering
    \subfloat[]{\includegraphics[width=0.5\linewidth]{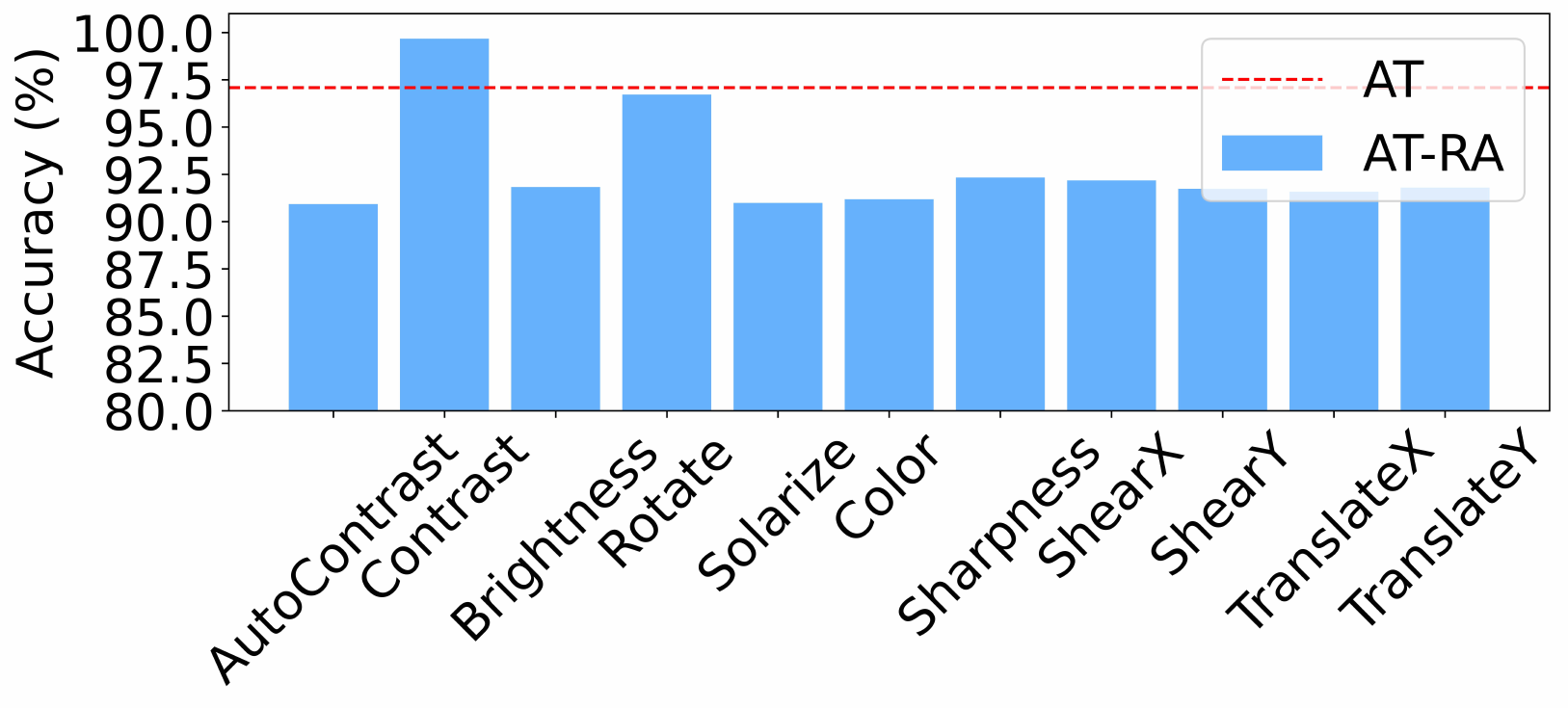}}
    \subfloat[]{\includegraphics[width=0.5\linewidth]
    {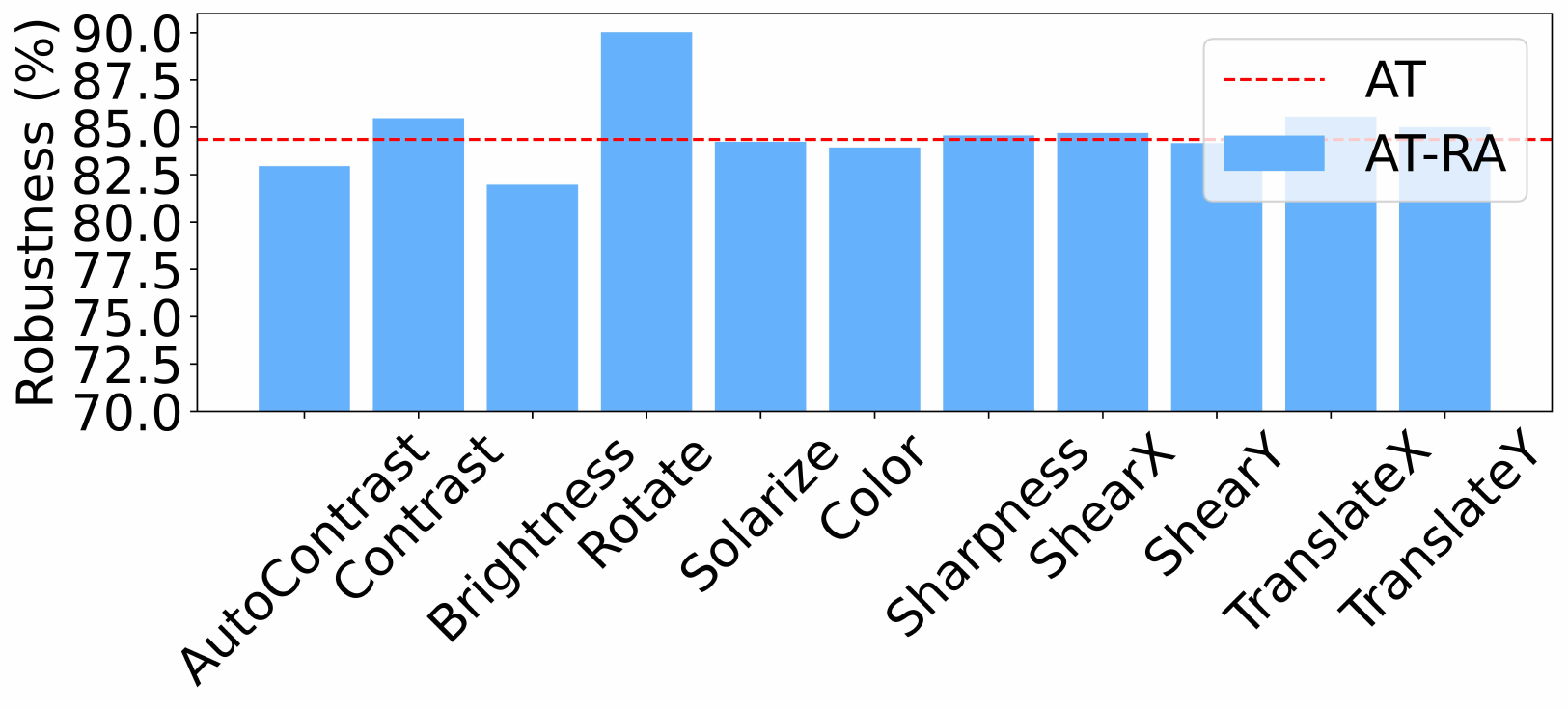}}\\
    \subfloat[]{\includegraphics[width=0.5\linewidth]{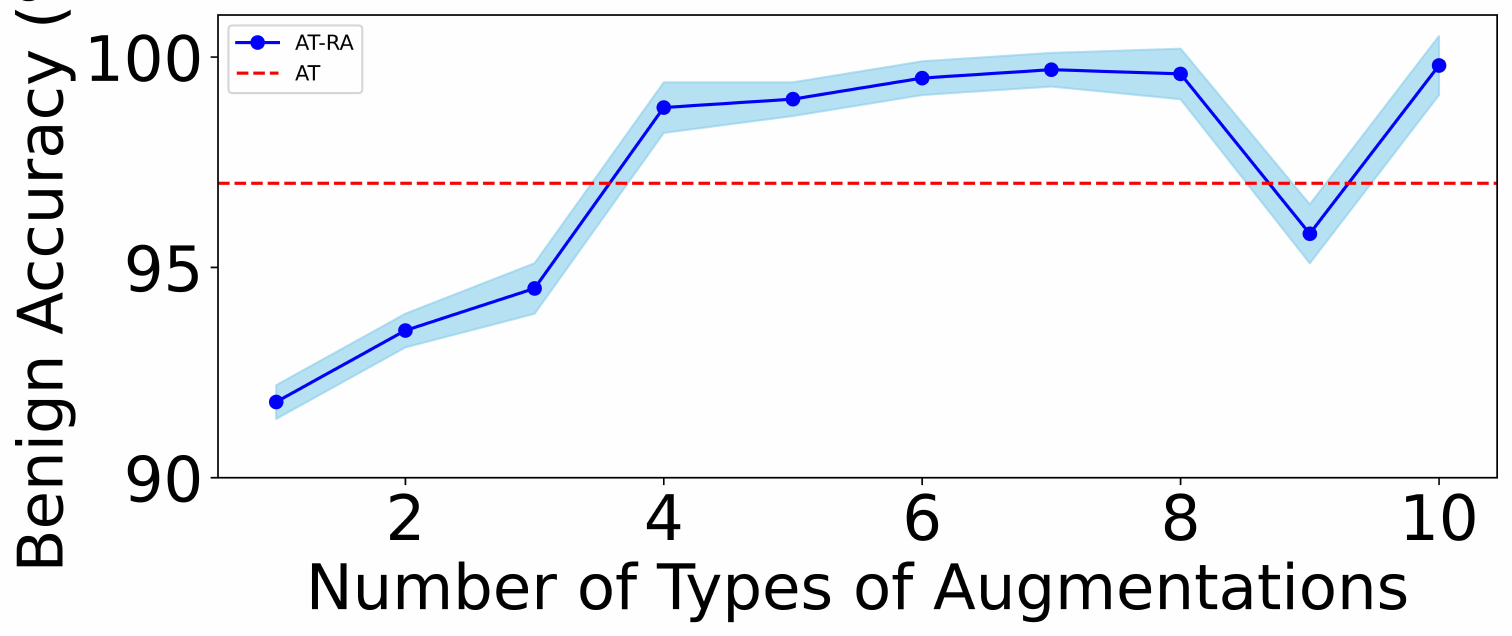}}
    \subfloat[]{\includegraphics[width=0.5\linewidth]
    {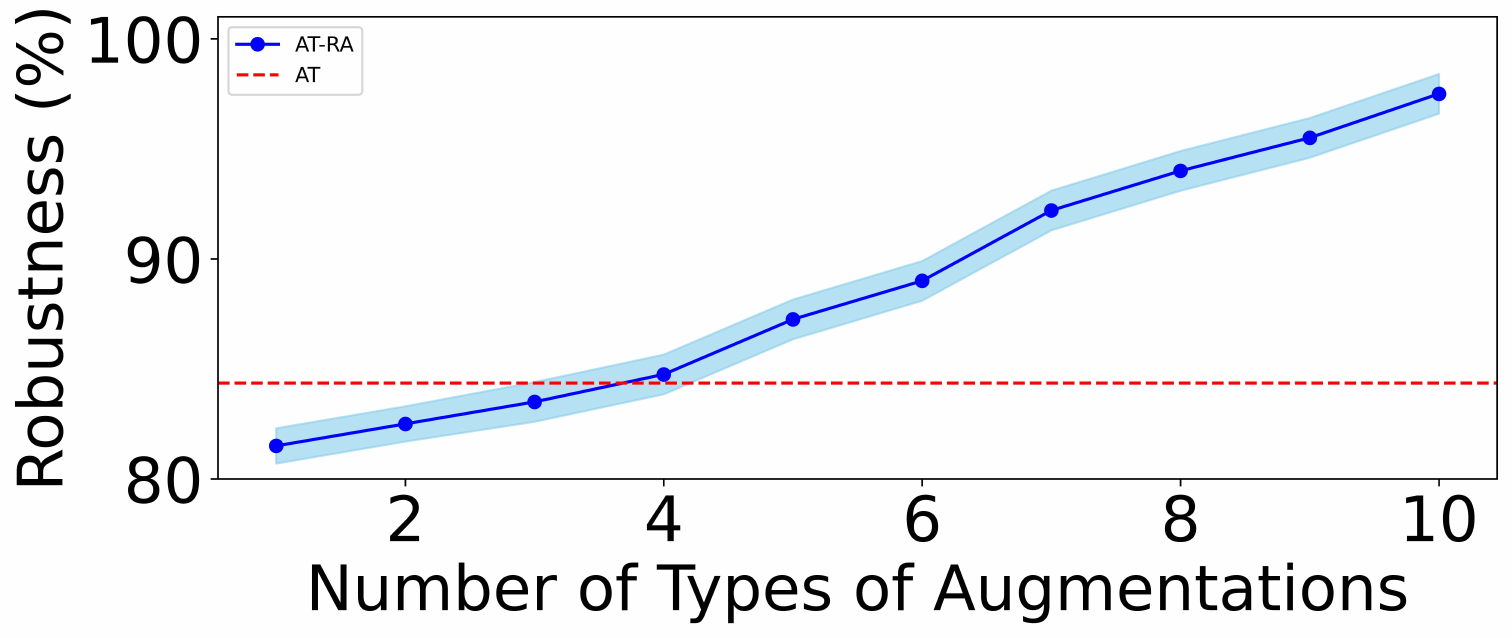}}
    \caption{Ablation study of data augmentation strategies under adversarial training. (a)--(b) Performance of individual augmentation methods. (c)--(d) Effect of increasing augmentation diversity via RandAugment on benign accuracy and robustness.}
    \label{fig2-2}
\end{figure}

\paragraph{Effect of Individual Data Augmentation Methods}
RandAugment applies a sequence of augmentation operations randomly selected from a predefined set of methods. To examine the contribution of each augmentation type, we conduct an ablation study by restricting the augmentation pool to a single operation and enforcing its consistent application across all training samples. The resulting benign accuracies are reported in Fig.~\ref{fig2-2}(a). It can be observed that, with the exception of \textit{Contrast}, no individual augmentation method leads to a noticeable improvement in benign accuracy.

We further evaluate the impact of each individual augmentation on adversarial robustness, as shown in Fig.~\ref{fig2-2}(b). Similar to the benign setting, no single augmentation method consistently improves robustness under adversarial attacks, except for a marginal gain observed with \textit{Rotate}. These results indicate that relying on a single augmentation operation is insufficient to enhance both benign accuracy and adversarial robustness under AT.

\paragraph{Effect of Combining Multiple Data Augmentation Methods}
We next investigate the effect of increasing augmentation diversity by progressively enlarging the RandAugment pool. In each experiment, \( n \) augmentation types (\( 2 \leq n \leq 11 \)) are randomly selected to construct the augmentation set. The corresponding results are illustrated in Fig.~\ref{fig2-2}(c) and Fig.~\ref{fig2-2}(d). As the number of augmentation types increases, benign accuracy exhibits a clear upward trend despite minor fluctuations, suggesting that a more diverse augmentation pool facilitates more robust feature learning. More importantly, adversarial robustness improves steadily with increasing augmentation diversity. This behavior indicates that combining multiple augmentation operations effectively enhances spectral–spatial variability while preserving structural consistency, enabling the model to learn more stable and semantically meaningful representations under adversarial perturbations.

Overall, these results demonstrate that the effectiveness of RandAugment stems not from any single augmentation operation, but from the complementary effects of multiple augmentation types. Such diversity is essential for mitigating spectral semantic degradation and improving both benign accuracy and adversarial robustness during AT.

\section{Conclusion}\label{Conclusion}
Adversarial robustness has become a central concern in HSI due to the growing vulnerability of deep models to imperceptible perturbations. In this work, we systematically examined adversarial training for HSI and identified a critical limitation that standard AT tends to induce misclassification concentration phenomenon, primarily caused by the elimination and distortion of spectral semantic information under adversarial perturbations. To address this challenge, we developed two complementary methods. The proposed AT-HARL introduces hierarchical losses to restore the underlying spectral
semantics and reinforce the robustness of decision boundaries, while AT-RA leverages RandAugment to enrich spectral–spatial diversity and mitigate the misclassification phenomenon. Extensive experiments across four widely used hyperspectral datasets demonstrated that both approaches markedly improve benign accuracy and adversarial robustness, surpassing state-of-the-art robust models. These findings underscore the importance of adversarial training in HSI and provide a foundation for enhancing the robustness of the hyperspectral model.

\bibliographystyle{IEEEtran}
\bibliography{reference}

\end{document}